\documentclass[acmsmall,screen,nonacm]{acmart}
\setcopyright{none}
\makeatletter
\@ifundefined{hyxmp@parse@acmart}{}{\let\hyxmp@parse@acmart\relax}
\makeatother
\usepackage{amsmath,booktabs,array,tabularx,enumitem,microtype,tikz,needspace}
\usetikzlibrary{positioning,arrows.meta,shapes.geometric,fit}
\usepackage{xurl}
\definecolor{vizP1}{HTML}{4C78A8}
\definecolor{vizP2}{HTML}{D58B73}
\definecolor{vizP3}{HTML}{4C9F8A}
\definecolor{vizS4}{HTML}{D4AE64}
\definecolor{vizS5}{HTML}{B78DA4}
\definecolor{vizS6}{HTML}{4C9F8A}
\definecolor{vizS7}{HTML}{8A7BB8}
\definecolor{vizC0}{HTML}{707B86}
\definecolor{vizC1}{HTML}{8A7BB8}
\definecolor{vizC2}{HTML}{439783}
\definecolor{vizSuccess}{HTML}{439783}
\definecolor{vizProcess}{HTML}{D59864}
\definecolor{vizEvidence}{HTML}{8A7BB8}
\definecolor{vizBlue2}{HTML}{C4D8E7}
\definecolor{vizBlue25}{HTML}{A4C1DB}
\definecolor{vizBlue4}{HTML}{4C78A8}
\definecolor{vizBlue45}{HTML}{4C78A8}
\definecolor{vizBlue6}{HTML}{254D6D}
\definecolor{vizRed}{HTML}{C66C6C}
\definecolor{vizRedLight}{HTML}{E5B5B2}
\definecolor{vizMid}{HTML}{F0F1F3}
\definecolor{vizNone}{HTML}{D8DBDF}
\definecolor{vizInk}{HTML}{20262E}
\definecolor{vizInk2}{HTML}{505963}
\definecolor{vizMuted}{HTML}{87909A}
\definecolor{vizGrid}{HTML}{E5E8EB}
\definecolor{vizAxis}{HTML}{B9C1C9}
\definecolor{vizQuery}{HTML}{439783}
\definecolor{vizRetry}{HTML}{C66C6C}
\definecolor{vizNoFollow}{HTML}{707B86}
\definecolor{vizUntriggered}{HTML}{E5E8EB}
\definecolor{vizCommitted}{HTML}{4C78A8}
\definecolor{vizUncommitted}{HTML}{D58B73}
\tikzset{
  vizgrid/.style={draw=vizGrid,line width=0.3pt},
  vizaxis/.style={draw=vizAxis,line width=0.4pt},
  viztick/.style={font=\scriptsize,text=vizInk2,inner sep=1pt},
  vizlab/.style={font=\scriptsize,text=vizInk,inner sep=1pt},
  viztitle/.style={font=\footnotesize,text=vizInk,inner sep=1pt},
  vizgap/.style={draw=white,line width=0.6pt},
}

\newcommand{\CtwoAssignedEpisodes}{360}
\newcommand{\CtwoTriggeredEpisodes}{310}

\newcommand{\CtwoToolQueryEpisodes}{60}
\newcommand{\CtwoVerificationFirstEpisodes}{63}
\newcommand{\CtwoRetryFirstEpisodes}{12}
\newcommand{\CtwoNoFollowUpEpisodes}{235}
\newcommand{\CtwoIndeterminateEpisodes}{0}
\newcommand{\CtwoNotTriggeredEpisodes}{50}

\hypersetup{hidelinks,keeppdfinfo,pdftitle={AGIBenchmark: Evaluating General-Purpose AI Agents in Dynamic Stateful Environments},pdfauthor={Genliang Zhu; Ziyuan Wang; Xiaobo Chen; Chu Wang},pdfcreator={LaTeX}}
\setlist{nosep,leftmargin=*}
\newcolumntype{Y}{>{\raggedright\arraybackslash}X}
\newcommand{\code}[1]{\texttt{#1}}

\title[AGIBenchmark: General-Purpose Agent Evaluation]{AGIBenchmark: Evaluating General-Purpose AI Agents in Dynamic Stateful Environments}
\subtitle{Scenario-based task design, frozen protocol, collection record, and results}
\author{Genliang Zhu}
\affiliation{\institution{Accentrust}\city{Vancouver}\country{Canada}}
\affiliation{\institution{Georgia Institute of Technology}\department{School of Computer Science}\city{Atlanta}\state{Georgia}\country{USA}}
\email{sebby@gatech.edu}
\author{Ziyuan Wang}
\affiliation{\institution{Duke University}\city{Durham}\state{North Carolina}\country{USA}}
\email{ziyuan.wang@ieee.org}
\author{Xiaobo Chen}
\affiliation{\institution{University of Massachusetts Amherst}\city{Amherst}\state{Massachusetts}\country{USA}}
\email{xchen@umass.edu}
\author{Chu Wang}
\affiliation{\institution{Accentrust}\city{Vancouver}\country{Canada}}
\affiliation{\institution{University of Illinois Urbana-Champaign}\city{Urbana}\state{Illinois}\country{USA}}
\email{wallywang@ieee.org}

\date{}
\begin{document}
\begin{abstract}
Stateful AI agents must complete tasks despite revised information and uncertain operation outcomes. AGIBenchmark evaluates these capabilities on twenty-four original synthetic task families in eight scenario groups spanning daily, enterprise, and technical requests. Every configuration receives the same tasks and tools. C0 supplies final information from the start; C1 delivers an explicit correction while preserving the same final obligations. In twelve state-changing families, C2 returns an ambiguous operation status in two hidden worlds: the operation either took effect or did not. Task-specific scorers evaluate final-state goal completion, applicable process constraints, and reported evidence. Fifteen open-weight models run through the official OpenHands Software Agent SDK with Ollama inference and vendor-default reasoning under a frozen resource contract, yielding 72 episodes per configuration and 1,080 overall. A separately implemented scorer reconstructed every goal outcome; task-card adjudication revised three evidence outcomes. Under C0, two configurations completed 22 of the 24 tasks, while three completed none. Across 360 configuration--task pairs, C1 produced 52 success-to-failure and 20 failure-to-success transitions. An exploratory C2 breakdown found 67 completions in 180 committed-world episodes and 14 in 180 uncommitted-world episodes, compared with 70 in 180 matched C0 episodes. Post-hoc operation-matched analysis found neither a successful target-covering state query nor a successful related retry in \CtwoNoFollowUpEpisodes{} of \CtwoTriggeredEpisodes{} triggered episodes. The benchmark separates ordinary completion, correction handling, and outcome verification, with exact task-level outcomes and documented configuration differences.
\end{abstract}

\maketitle
\hypersetup{keeppdfinfo,pdfauthor={Genliang Zhu; Ziyuan Wang; Xiaobo Chen; Chu Wang},pdfcreator={LaTeX}}
\clearpage
\setcounter{tocdepth}{1}
\tableofcontents
\clearpage
\section{Introduction}\label{sec:introduction}
People increasingly ask language-model agents to carry out requests rather than answer questions: rebook a disrupted train connection, process a return under a store's policy, match invoices against purchase orders before posting them, revise a report against a fact sheet, or fix a reported bug. Each request requires the agent to gather information through tools, apply rules that may contain exceptions, and often change the state of a system. What matters is the resulting state and whether the agent's report about it is true, not how convincing its final message sounds. A booking summary can describe a ticket that was never issued. A refund can be submitted twice. A code change can pass the visible tests while breaking a protected module.

Several benchmarks evaluate agents in interactive environments. AgentBench covers multiple environments for reasoning and decision making, and GAIA poses general-assistant questions requiring browsing and tool use~\citep{liu2024agentbench,mialon2024gaia}. $\tau$-bench and its successor evaluate tool-using agents in customer-service domains with policies and simulated users~\citep{yao2025taubench,barres2025tau2}. AppWorld and WorkArena cover application and knowledge-work interactions~\citep{trivedi2024appworld,drouin2024workarena}, and SWE-bench and OSWorld check software changes and computer operations by execution~\citep{jimenez2024swebench,xie2024osworld}. These resources motivate an evaluation that combines breadth across kinds of requests with controlled changes to the information and observations under which the same task is attempted.

Three distinctions are central to this study. \emph{Breadth} concerns the range of requests: everyday errands, enterprise processes, and technical work. More models evaluated on one application workflow do not supply it. \emph{Adaptation} concerns behavior after an explicit correction to the information of the same task. It should not be inferred from comparing two unrelated tasks of different difficulty. \emph{Reliable execution} concerns what an agent actually changes and how it establishes the result when the system's reply is inconclusive. It should not be imposed on a task that only requires an answer.

AGIBenchmark specifies such an evaluation across eight scenario groups in three categories. The daily category covers travel and scheduling, personal affairs and purchasing, and everyday inquiry. The enterprise category covers customer service and orders, finance and procurement, and office documents. The technical category covers software engineering and data analysis. Each group contributes three formal task families with different objectives and checks. A rail trip with a disrupted return, a reimbursement review under an expense policy, and a refactoring under behavioral constraints differ in substance, not only in their entities. All tasks are original synthetic constructions with fictional people and organizations and no dependence on any company's product.

The model-comparison contract is uniform. Every configuration receives the same twenty-four tasks, the same applicable conditions, the same tools, the same resource limits, and the same scoring. Fifteen open-weight models are evaluated through the official OpenHands Software Agent SDK~\citep{wang2025openhands,openhandsSDK}, with task-specific tools, Ollama as the inference server, and each model's default reasoning behavior. Three operators are each responsible for five configurations. Execution is divided by model, never by task; the actual platforms and collection deviations are documented in Section~\ref{sec:collection}.

Within that matrix, condition C0 supplies the final task information from the start. C1 starts from an earlier version of one piece of information and delivers an explicit correction when the agent first uses it, while keeping the final problem identical to C0. C2 applies only to the twelve families whose completion requires a state-changing operation. The first call of that operation returns an ambiguous status, and in one of two hidden worlds the operation took effect while in the other it did not. One instance per family, one run each for C0 and C1, and one run in each C2 world give 72 episodes per configuration and 1,080 in total. These are 1,080 episodes on twenty-four base tasks, not 1,080 independent problems.

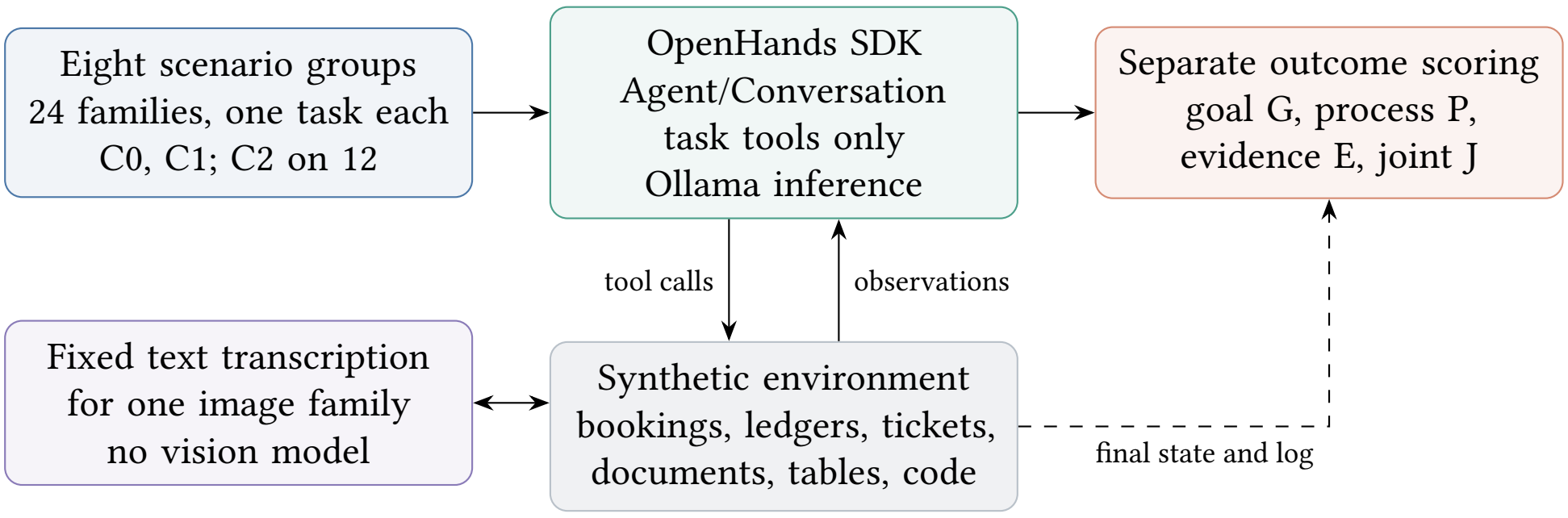
\begin{figure}[t]
\centering
\begin{tikzpicture}[font=\small,>=Stealth,
 box/.style={draw=vizAxis,rounded corners,align=center,text width=.25\linewidth,minimum height=1.3cm,inner sep=5pt}]
\node[box,draw=vizP1,fill=vizP1!8] (task) {Eight scenario groups\\24 families, one task each\\C0, C1; C2 on 12};
\node[box,draw=vizP3,fill=vizP3!8,right=.045\linewidth of task] (agent) {OpenHands SDK\\Agent/Conversation\\task tools only\\Ollama inference};
\node[box,draw=vizP2,fill=vizP2!10,right=.045\linewidth of agent] (check) {Separate outcome scoring\\goal G, process P,\\evidence E, joint J};
\node[box,fill=vizMid,below=1cm of agent] (env) {Synthetic environment\\bookings, ledgers, tickets,\\documents, tables, code};
\node[box,draw=vizS7,fill=vizS7!8,below=1cm of task] (vision) {Fixed text transcription\\for one image family\\no vision model};
\draw[->] (task.east)--(agent.west);
\draw[->] (agent.east)--(check.west);
\draw[->] ([xshift=-.45cm]agent.south)--node[left,font=\scriptsize]{tool calls} ([xshift=-.45cm]env.north);
\draw[->] ([xshift=.45cm]env.north)--node[right,font=\scriptsize]{observations} ([xshift=.45cm]agent.south);
\draw[<->] (vision.east)--(vision.east-|env.west);
\draw[->,dashed] (env.east)-|node[pos=.3,below,font=\scriptsize]{final state and log} (check.south);
\end{tikzpicture}
\Description{Twenty-four families in eight scenario groups are attempted by the official OpenHands Agent/Conversation loop with task-specific tools and Ollama inference. Tool calls and observations pass through an isolated synthetic environment; one family uses a fixed text transcription of images. A separate outcome scorer reads the final state and event log to compute goal, process, evidence, and joint outcomes. The diagram shows the evaluation structure, not results.}
\caption{Evaluation boundary. All configurations receive the same tasks and tools. The agent acts only through task tools in an isolated synthetic environment; one family receives image content as fixed text. Scoring reads the final state and event log, separately from the agent loop, to compute G, P, E, and J. The figure depicts the evaluation design.}\label{fig:overview}
\end{figure}

Goal completion G, the task-specific check of the final state or submission, is the primary endpoint. Applicable process constraints P, evidence checks E, and their conjunction J are reported alongside it. They separate, for example, a correct final state reached through a forbidden operation from one reported with an unsupported claim. Figure~\ref{fig:overview} summarizes the evaluation boundary.

The research questions are:
\begin{enumerate}
\item \textbf{Completion across scenario groups.} Under a shared task and resource contract, how do configurations perform across the eight groups and twenty-four families?
\item \textbf{Adaptation to corrections.} When final obligations are matched, how does an explicit correction change task completion and resource use?
\item \textbf{Execution under ambiguous replies.} On tasks with genuine state changes, how do ambiguous operation replies affect final state, duplicated effects, and supported completion?
\item \textbf{Configured-system diagnostics.} Under the common framework, which failures are associated with recorded properties of the configuration, such as terminal reasons, tool-call formats, and resource use? Causal comparisons of frameworks, reasoning settings, or hardware are outside this design.
\end{enumerate}

This manuscript contributes the task inventory and its specifications, the condition design with matched final obligations and two-world ambiguity, a frozen and fingerprinted execution protocol, a documented collection record, and the results of all 1,080 episodes. Under C0, completion ranged from 22 of the 24 tasks for the two strongest configurations to none for three configurations, and four configurations reached at least 17. Under the explicit correction, 52 of the 360 configuration--task pairs changed from success to failure and 20 from failure to success. In the exploratory breakdown of the twelve state-changing tasks, completion was 67 of 180 episodes when the operation had taken effect and 14 of 180 when it had not, compared with 70 of 180 matched C0 episodes. A post-hoc operation-matched classification further distinguishes state-query-first, related-retry-first, and neither-successful-follow-up paths on the recorded tool events (Section~\ref{sec:c2-recovery}). Section~\ref{sec:results} reports the results and Section~\ref{sec:analysis} distinguishes the original summaries from exploratory reanalyses. The separate scoring implementation and task-card adjudication are documented in Section~\ref{sec:verification-status}, including their development provenance and information access.

\section{Related Work and Measurement Scope}\label{sec:related}
\subsection{Broad capabilities and interactive agents}
AgentBench organizes evaluation across distinct interactive environments and identifies reasoning, decision making, and instruction following as relevant dimensions of agent behavior~\citep{liu2024agentbench}. GAIA combines everyday questions with multiple abilities, including reasoning, browsing, and tool use~\citep{mialon2024gaia}. Their relevance here is methodological: a broad agent evaluation needs different problem settings and must inspect the work needed to answer or act. A collection confined to one product's operations cannot acquire the same breadth merely through a broad title.

AGIBenchmark adopts an inventory of request types rather than product modules. Its design adds a particular comparison structure. All configurations receive the same versioned task keys, corrections are matched to the same final obligations, and ambiguous-effect comparisons are restricted to tasks with actual state changes. These choices make the relationship between an ordinary task and its changed interaction conditions explicit. The study's value rests on its tasks and the outcomes measured under that structure, not on broad evaluation as a new idea.

HELM provides a precedent for presenting multiple evaluation dimensions instead of a single performance number~\citep{liang2023helm}. In the present study, a profile over eight scenario groups and twenty-four families is reported alongside any summary, and quality, resource use, and execution diagnostics remain separately interpretable. An equally weighted group average summarizes this finite inventory. It does not establish a universal scale on which all forms of intelligence or all future tasks have been measured.

\subsection{Reasoning-centric evaluations and the scope of this inventory}
MATH evaluates mathematical problem solving, and ScienceWorld contributes interactive scientific tasks in a text environment~\citep{hendrycks2021math,wang2022scienceworld}. An earlier version of this benchmark had dedicated mathematics and science groups. The current inventory instead organizes tasks by the kinds of requests people make of agents, so arithmetic, statistics, and rule application appear inside travel budgets, expense limits, invoice matching, and data analysis. They do not appear as standalone mathematical or scientific problems, and the study makes no claim about mathematical or scientific reasoning as such. Tool access is declared, so where a calculator or query tool is available the tasks evaluate formulation, tool selection, and verification as well as the generated answer.

\subsection{Software changes and computer interaction}
SWE-bench evaluates repository modifications for real software issues and emphasizes changes that can span multiple functions or files~\citep{jimenez2024swebench}. This supports patch validity and regression behavior as native software endpoints. The software families here cover defect repair, interface migration, and constrained refactoring of small synthetic packages, with hidden tests and protected files. A solution is judged by its behavior and its preservation obligations, not by textual similarity to a reference patch. The packages are small, so the families do not measure work on large real repositories.

OpenHands provides an open platform for developer-like agents, sandboxed execution, and benchmark integration~\citep{wang2025openhands}. The main configuration uses the Agent/Conversation loop of its official Software Agent SDK with benchmark-specific tools, and Ollama supplies model inference rather than a competing agent loop~\citep{openhandsSDK,openhandsCustomTools,openhandsLocalLLMs}. This is a configured OpenHands evaluation, not a claim about the unchanged default product. Holding the framework and task tools fixed lets model configurations be compared without a framework difference. It does not identify the effect of the framework itself.

WebArena provides an executable web environment for agent tasks, while OSWorld evaluates open-ended computer operations with execution-based checking~\citep{zhou2024webarena,xie2024osworld}. These resources show that a browser, graphical interface, or operating system is itself part of the evaluated setting. The present interface instead provides bounded, recorded task tools. This improves control of the shared assignment but changes what is measured: explicit booking or file tools do not establish graphical grounding ability in an unrestricted desktop.

ALFWorld aligns abstract text environments with embodied tasks~\citep{shridhar2021alfworld}. It illustrates that representing a world and executing within it are separable experimental choices. The itinerary and scheduling tasks here use explicit simulated systems. Their outcomes describe performance in those environments, not physical action or real-world logistics.

\subsection{Images and fixed transcriptions}
MMMU evaluates multidisciplinary multimodal understanding, and ChartQA concerns chart questions that require visual and logical reasoning~\citep{yue2024mmmu,masry2022chartqa}. Both show why giving an agent a symbolic description of an image changes the assessment: information that originally requires perception has already been supplied in another representation. The single image-based family in this study does exactly that. Its tool returns a fixed text transcription of photos and screenshots, identical for every model. The family therefore measures the use of transcribed image content, not perception. The study makes no claim about visual or multimodal capability, and a native multimodal configuration would be a separate design.

\subsection{Documents, sources, and knowledge work}
XTREME demonstrates the importance of evaluating linguistic transfer across languages rather than assuming that English performance characterizes multilingual behavior~\citep{hu2020xtreme}. All tasks here are in English. The document families check specific relations, such as values matching a fact sheet, protected paragraphs left untouched, or a revision log matching the changes, and do not assess writing quality in general.

The inquiry families distinguish a correct answer from a supported answer. They require reading a frozen collection of guides, notices, and sources, applying an explicit priority rule when sources conflict, and citing the section or source that supports each answer. The collection and its corrections are controlled parts of the environment. This gives the correction comparison a stable target without assuming that a live web search returns the same material on every run.

Professional and enterprise benchmarks provide specialized examples. AppWorld evaluates application interactions, WorkArena and WorkArena++ cover knowledge-work and compositional tasks, TheAgentCompany provides a simulated workplace, and GDPval concerns professional work products~\citep{trivedi2024appworld,drouin2024workarena,boisvert2024workarenapp,xu2025agentcompany,patwardhan2025gdpval}. In the present catalogue, enterprise processes are three of eight groups, written as generic synthetic processes. No company's feature list defines which requests count.

\subsection{Changing information and ambiguous execution}
ToolSandbox, the $\tau$-bench family, and Gaia2 study stateful interaction, policy-dependent behavior, repeated reliability, or dynamic environments~\citep{lu2025toolsandbox,yao2025taubench,barres2025tau2,froger2026gaia2}. BizAgentBench is a direct neighbor for versioned business policies and long-horizon process constraints~\citep{zhao2026bizagentbench}. The C1 condition here is a narrower matched comparison: one episode begins with the final information, while the other receives an explicit correction to the same final information and obligations. The interpretation follows that intervention rather than a broad claim of continual learning.

Ambiguous tool effects have a direct neighbor in Limbo, and AgentChaos studies fault injection at the model interface~\citep{li2026limbo,tan2026agentchaos}. These are different failure locations. A well-formed action can produce an effect whose confirmation is missing, while a malformed model response can fail before any tool action occurs. C2 addresses the former on the twelve state-changing families. Formatting and transport failures are identified separately by terminal reason, so a score change is not attributed to recovery reasoning when its cause was a failed request.

\subsection{Completion, constraints, and evidence}
ST-WebAgentBench combines completion with policy adherence, and Agent-Diff uses state-difference evaluation for API tasks~\citep{levy2026stweb,pysklo2026agentdiff}. SafeActBench and AgentEval motivate careful treatment of the evidence available at action and reporting boundaries~\citep{lin2026safeactbench,gao2026agenteval}. The present specification keeps task-specific correctness and declares separately which process and evidence predicates apply.

AgentDojo and ToolEmu address risk and adversarial behavior in tool-using settings~\citep{debenedetti2024agentdojo,ruan2024toolemu}. Their objectives complement this study. A clean completion record under the present conditions is not an evaluation against an adversarial attack distribution, and a process constraint such as not acting on an already-reimbursed report is a task rule, not a security benchmark.

\subsection{Comparison quality, assets, and resource accounting}
SkillsBench provides a contemporary reference for diverse task assets, paired task conditions, and distinctions between model and harness configurations~\citep{li2026skillsbench}. AGIBenchmark uses it as a construction and presentation reference. The current design has no factor for the presence of Skills, and its task counts are reported in their own terms: twenty-four base tasks, not 1,080.

AI Agents That Matter and HAL motivate transparent system and cost comparisons~\citep{kapoor2025agentsmatter,kapoor2026hal}. tinyBenchmarks studies efficient evaluation with fewer examples~\citep{polo2024tinybenchmarks}. Here every included configuration receives the full task set. A model-specific subset would answer a different comparison. Resource tables report requests, tool calls, output tokens, and time, including those of failed episodes, and compare time only within a platform.

Rabanser et al. separate consistency, robustness, predictability, and safety in a reliability profile~\citep{rabanser2026reliability}. The present design does not measure consistency across reruns: no task is repeated under the same condition and hidden world. Its condition contrasts address only the two perturbations declared here. A high C0 rate therefore remains distinct from stability across reruns.

Datasheets for Datasets supplies a documentation precedent for sources, intended use, collection, and distribution~\citep{gebru2021datasheets}. The task package accordingly documents that all tasks are original synthetic constructions and records the origin of the six families carried over from an earlier version. The earlier AGIBench is a separate work~\citep{tang2023agibench}, and the present title does not imply affiliation with it. Lifecycle research by the authors is a related study~\citep{zhu2026lifecycle}; its traces and enterprise tasks are not relabelled as observations of this benchmark.

\section{Measurement Objectives and Research Contribution}\label{sec:objectives}
A benchmark is useful when its measurements answer a question that survives changes of model, task content, and deployment. For AGIBenchmark, the question is how well a configured agent completes different kinds of requests, and how its completion responds to two specified changes: an explicit correction of information, and an inconclusive reply from a state-changing operation. The organizing object is a task with a declared objective and a verification procedure, not an application feature. This section connects the inventory to these claims and explains why a common task matrix is needed.

\subsection{Coverage as a design property}
The eight scenario groups contribute complementary demands. Travel tasks combine search, timing constraints, and bookings. Personal-affairs tasks combine documents, rules, and purchases. Inquiry tasks combine conflicting or versioned sources with citation. Customer-service tasks apply store policies to orders and tickets, and finance tasks apply matching rules and limits to ledgers and purchase orders. Office tasks transform notes, fact sheets, and tables into documents and tasks. Software tasks change code under behavioral contracts, and data tasks compute results from tables under explicit rules. A task can draw on several of these demands, but it belongs to one group before outcomes are observed.

The catalogue records the operation needed to solve each task, which is more informative than a topic label. A travel task that only asks for the cheapest fare differs from one that requires checking transfer times, a fare policy, and an existing booking. Each family's card therefore states the objective, the information dependencies, the permitted tools, and the scorer together, so that the task actually exercises the distinction used to describe it.

Content coverage is not measured competence. The study covers the twenty-four declared families. It does not measure audio or speech interaction, long-video understanding, physical action, real social interaction, long-term learning, standalone mathematics or science, or native visual perception, and no such coverage is filled into the profile.

\subsection{A common task population for model comparison}
The common assignment contract prevents task selection from becoming an unrecorded explanation of model differences. Every configuration receives the same twenty-four families, the same instance, and the same applicable conditions and C2 worlds. A model run on only part of the inventory has incomplete coverage and is not given a comparable complete-set score. No model was selected for additional tasks after its first results.

Uniformity applies to the task environment, not to every internal computation. Models tokenize the same text differently, use different chat templates, and by default reason to different extents. These differences belong to the configured system. The comparison holds constant the user-visible task, the authorized information, the tools, the resource limits, and the scoring, and records the remaining deployment differences.

Execution responsibility is separate from task content. Each operator runs the models assigned to a device on the full inventory. Task authorship and review are recorded separately from execution, so an apparent model advantage cannot be explained by one operator running a different version of the task: fingerprints bind the code each episode used.

\subsection{Generalization across task families}
Generalization is operationalized by separating development and formal families by content and solution structure. Changing a name or a constant creates a variation, not a new family. The eight development families were used only to exercise the runner and tools, and no prompt or budget was tuned on formal families. With one instance per family, the generalization claim rests on the variety of families and groups. It describes how a configured agent applies its tools and instructions to these unseen tasks. It does not imply that similar material was absent from pretraining.

\subsection{Adaptation with matched final obligations}
C1 compares two ways of supplying information for the same final problem. C0 makes the final information available from the start, while C1 supplies an earlier version followed by an explicit correction at a declared event. The final target, the permitted operations, and the scorer are matched. Without this match, a score difference would reflect a different task rather than a response to the correction. The realized work can differ, because rereading a source or redoing a calculation costs resources, and that cost is part of the measured response.

\subsection{Reliable execution where actions change a state}
C2 is relevant only where an operation changes the state of a simulated system. The agent's reply can be inconclusive while the change did or did not happen. This separates the action, its actual effect, and what the agent observes. On these tasks the scorer can ask whether an agent finishes correctly, duplicates an effect, and reports what is actually true. Answer-only tasks receive no artificial operation, and the C2 contrast uses the same twelve families and their own C0 episodes for every configuration.

\subsection{Distinguishing the agent from its tools}
The evaluated object is a configuration: a model, its agent loop, and its declared tools. A calculator or query tool computes, but the agent must formulate the operation and use the result appropriately. A test tool reports public test results, while hidden tests remain unavailable. The fixed transcription in the image-based family supplies the same text to every model, so that family measures what the agent does with the transcription and not how it perceives images. Explanations of failures, such as a misread rule, a skipped verification, or an unparsed tool call, require trace evidence and fixed labels and are not inferred from a final binary score.

\section{Task Formulation and Verification Criteria}\label{sec:task-formulation}
A task in AGIBenchmark is a declared objective over a submitted answer, an artifact, or the final state of a simulated system. What makes it an agent task is the permitted sequence of tool calls and observations through which the system investigates and completes the objective. Some tasks only require reading and combining information, such as answering questions from a fixed set of documents. Others change state, such as booking travel, posting invoices, or submitting a code change. The formulation makes these cases comparable at the level of task identity while keeping each task's own correctness criteria.

\subsection{Scenario groups, families, and configurations}
Let $\mathcal G$ be the eight scenario groups and $\mathcal F_g$ the three formal families of group $g$. A family fixes an objective, an information and tool structure, and a verification procedure. In this study each family has exactly one instance, a complete initial problem with its data, documents, repository state, and constants, so family and base task coincide. A condition defines a controlled variation in how information is delivered or how an effect is observed. A C2 repeat resets the agent session and environment and changes only the hidden effect world.

A configuration identifies the model and its exact weights and quantization, the inference server and its version, the chat template, the agent framework and its tools, the resource contract, and the device. Two quantizations of the same base weights would be different configurations but not independent model families. The fifteen configurations of this study are listed in Table~\ref{tab:models}.

The task key consists of family, instance, repeat, and condition. Model and device identifiers belong to the assignment record, not to the task key. Every configuration receives the same keys, which allows matched comparisons and makes omissions detectable without confusing a new model with a new question. The key does not expose the hidden world, answer data, or reference solutions.

\subsection{Applicability as part of the task contract}
Let $a_{fc}\in\{0,1\}$ indicate whether condition $c$ applies to family $f$. The design is
\begin{equation}
 a_{f0}=a_{f1}=1,\qquad
 a_{f2}=\mathbf{1}\{f\in\mathcal{F}_{\mathrm{C2}}\},
 \quad |\mathcal{F}_{\mathrm{C2}}|=12 .
 \label{eq:condition-applicability}
\end{equation}
$\mathcal F_{\mathrm{C2}}$ contains the families of the travel, service, finance, and software groups. In each of these families, completing the task requires an operation that changes the state of the simulated system: a booking, an invitation, a refund, a reshipment, an escalation, a ledger posting, an expense decision, a purchase order, or a code change request. The mask is the same for all models. With one instance per family, one run each for C0 and C1, and two runs for C2, the number of episodes per configuration is
\begin{equation}
 \sum_{f\in\mathcal{F}}\bigl(a_{f0}+a_{f1}+2\,a_{f2}\bigr)
 = 24+24+2\cdot12=72 .
 \label{eq:uniform-task-count}
\end{equation}
An inapplicable cell is a design property, not an unobserved outcome. A missing run, a service failure, a model failure, and an inapplicable condition have separate representations. Treating them all as zero would change both the estimate and the meaning of the task matrix.

The task card also declares which diagnostics apply. A process constraint can be meaningful for an answer task, for example a prohibition on a tool that is not part of the allowed route. An execution-duplication diagnostic requires an actual effect. Evidence checks can concern a cited document section or a reported total rather than an operation receipt. The analysis reports the eligible denominator of each diagnostic and does not invent a universal receipt requirement to make its columns uniform.

\subsection{Interaction state and observations}
An initial specification $\theta_{f}$ binds the instance's public materials, private checking data, tools, final target, and versioned condition semantics. During a run, the environment has state $s_t$ and returns an observation $o_t$ for each tool call. The agent chooses action $u_t$ from its visible history and its configuration. We write the trajectory and transition as
\begin{equation}
 \tau=(o_0,u_1,o_1,\ldots,u_T,o_T),\qquad
 s_{t+1}=F_{\theta_{f},c}(s_t,u_{t+1},w).
 \label{eq:agent-trajectory}
\end{equation}
The symbol $w$ is the hidden C2 world, which decides whether the designated operation took effect. It is not a model-specific modification of the problem. The agent sees only what the tools return. Hidden test cases, answer data, and the world assignment do not become observable because the environment stores them.

The state is native to each task. In a travel task it holds bookings, cancellations, and the requests made to the booking system. In a finance task it holds the ledger, journal entries, and request results. In a software task it holds the repository files, test runs, and submitted change requests. In a document task it holds the current draft and its revision history. Every state change and every observation is logged in order, which allows common logging and task identity without forcing all tasks into one action model.

The tasks use no real-time events. Corrections and ambiguous replies are tied to the agent's own tool calls, such as the first read of a document or the first submission of an operation. Elapsed time is consumed only by actual inference and tool execution.

\subsection{Primary task completion}
The goal predicate is a family-specific function
\begin{equation}
 G_f(\tau)=
 \operatorname{checkGoal}_f\bigl(\theta_{f}^{\mathrm{final}},
             s_T,y_T\bigr)\in\{0,1\},
 \label{eq:goal-completion}
\end{equation}
where $y_T$ is the submitted answer or artifact and the superscript indicates the final specification. The scorer reads only the final state and the event log, never the agent's own description of what it did. The same final predicate is applied to C0, to C1, and to both C2 worlds.

The goal is native to each family. In a travel booking, the final state must hold exactly one valid outbound ticket, one valid return ticket, and one hotel stay meeting the constraints, within budget. In invoice matching, the set of invoices actually posted must equal the set that matches purchase orders and goods receipts. In a code-repair task, the last change request must pass public and hidden tests without modifying protected files. In a data-cleaning task, applying the submitted change set to the original table must reproduce the expected table cell by cell. Appendix~\ref{app:task-family-catalog} gives the goal of every family.

A scorer does not impose a single reference route. Where a task has several valid answers, such as several admissible itineraries or several correct refactorings, every valid answer is accepted. The reference solution only shows that the task is feasible under the declared tools and budget. Similarity to the reference route or a preferred tool order is never used in place of task correctness.

G is the primary endpoint because it has a declared task meaning in all twenty-four families. A group score averages the three families of a group, and the overall summary weights the eight groups equally. Section~\ref{sec:analysis} specifies these weights; they do not depend on which configuration has the most complete or most favorable records.

\subsection{Process constraints and evidence}
For an applicable task, $P_f(\tau)$ checks constraints on the route as well as the final state. Examples include booking a first-class ticket against a travel policy, holding two tickets for the same leg at any point, approving an expense report that was already reimbursed, writing to a protected file, or submitting a second change request. An ordinary invalid tool call that is rejected without effect is not a violation. A later repair does not erase a violation that the task forbids.

Evidence is defined by what the agent actually received and reported, not by stated confidence. Let $Q_f(y_T)$ be the task-relevant assertions in the agent's final report or submission. Let $O_f^{\mathrm{valid}}(\tau)$ be the observations actually delivered to the agent that pass the declared version and content checks. A generic coverage requirement is
\begin{equation}
 Q_f(y_T)\subseteq
 \bigcup_{e\in R_f(y_T)\cap O_f^{\mathrm{valid}}(\tau)}
       \operatorname{supports}_f(e),
 \label{eq:task-evidence-coverage}
\end{equation}
where $R_f(y_T)$ identifies the cited sources or records. Each family specifies its concrete form. In the itinerary task, the itinerary sent to the traveler must match the final bookings, and the reported spending must match the spending computed from the rules. In the document-question tasks, each answer must cite the section that supports it, in the version that is current. In code tasks, a passing test run must exist on the submitted workspace. An empty report does not pass because the empty set is covered.

Evidence quality and goal completion can overlap by construction and are not assumed independent. If a task's goal requires a correct cited value, a wrong citation can affect G as well as E. A supported partial result can also remain incomplete. These differences are recorded so that a reader understands each predicate rather than reading every E as the same kind of proof.

\subsection{Joint outcome and its scope}
Where G, P, and E all apply, the joint outcome is
\begin{equation}
 J_f(\tau)=G_f(\tau)P_f(\tau)E_f(\tau).
 \label{eq:joint-diagnostic}
\end{equation}
Where P does not apply, $J_f=G_fE_f$. J is reported with its components and does not replace the C0 group profile. On the same fully scored records, the identity
\begin{equation}
 G-J=G(1-P)+GP(1-E)
 \label{eq:completion-gap}
\end{equation}
separates a reached goal with a process violation from a reached goal with conforming process but insufficient evidence. The two terms are mutually exclusive. Averaging preserves the identity only over the same records and weights.

Table~\ref{tab:outcome-examples} illustrates these distinctions without implying observed frequencies. It shows what the scorer observes: a correct state reached by a forbidden route, a correct state with an unsupported report, or a fully supported completion. No example is a model outcome.
\begin{table}[t]
\centering\small
\begin{tabularx}{\linewidth}{Ycccc}
\toprule
Contract example & G & P & E & J\\\midrule
Code repair: the last change request passes public and hidden tests, no protected file was touched, and a passing test run exists on the submitted workspace. & 1 & 1 & 1 & 1\\
Code repair: the change request passes all tests and no protected file was touched, but the tests were never run on the submitted workspace. & 1 & 1 & 0 & 0\\
Code repair: the change request passes all tests and was tested, but the agent earlier tried to write a protected file. & 1 & 0 & 1 & 0\\
Document revision: one value still disagrees with the fact sheet; no protected paragraph was edited, and the revision log matches the changes actually made. & 0 & 1 & 1 & 0\\\bottomrule
\end{tabularx}
\caption{Contract examples from two formal families, not model outcomes. Each row follows the family's task card (Appendix~\ref{app:task-family-catalog}). The last row shows a supported but incomplete result. These diagnostics do not replace the C0 goal-completion endpoint.}\label{tab:outcome-examples}
\end{table}

\section{Matched Information Updates and Execution Observations}\label{sec:conditions}
The three conditions answer different questions on the same task population. C0 measures ordinary task completion. C1 measures completion after an explicit correction to the information of the same final problem. C2 measures completion when a genuine state-changing operation returns a reply that does not reveal whether it took effect. Conditions are properties of families, not privileges of particular models: every configuration receives the same applicability mask, final objectives, and task keys.

\begin{figure}[t]
\centering\begin{tikzpicture}[x=\linewidth,y=1cm,font=\small,>=Stealth,
 box/.style={draw,rounded corners,text width=.17\linewidth,align=center,minimum height=1.2cm,inner sep=3pt},
 stable/.style={box,draw=vizC0,fill=vizC0!7},
 revised/.style={box,draw=vizC1,fill=vizC1!8},
 uncertain/.style={box,draw=vizC2,fill=vizC2!8}]
\node[font=\footnotesize,anchor=east] at (.035,0) {C0};
\node[stable] (a0) at (.15,0) {Final information\\from the start};
\node[stable] (b0) at (.395,0) {Inspect and\\solve task};
\node[stable] (c0) at (.64,0) {Answer or\\required artifact};
\node[stable] (d0) at (.885,0) {Check final\\task obligations};
\node[font=\footnotesize,anchor=east] at (.035,-1.6) {C1};
\node[revised] (a1) at (.15,-1.6) {Earlier\\information};
\node[revised] (b1) at (.395,-1.6) {Correction attached\\to a tool result};
\node[revised] (c1) at (.64,-1.6) {Revise and\\complete task};
\node[revised] (d1) at (.885,-1.6) {Same final\\task obligations};
\node[font=\footnotesize,anchor=east] at (.035,-3.2) {C2};
\node[uncertain] (a2) at (.15,-3.2) {Final information\\state-changing\\families only};
\node[uncertain] (b2) at (.395,-3.2) {Designated operation\\(first call)};
\node[uncertain] (c2) at (.64,-3.2) {Ambiguous status:\\effect unknown};
\node[uncertain] (d2) at (.885,-3.2) {Verify the state,\\then act};
\foreach \r in {0,1,2}{\draw[->] (a\r.east)--(b\r.west);\draw[->] (b\r.east)--(c\r.west);\draw[->] (c\r.east)--(d\r.west);}
\node[align=center,font=\footnotesize,text width=.92\linewidth] at (.52,-4.25) {C0/C1: all 24 families. C2: the same 12 applicable families for every model.\\Logical order; illustrative paths, not mandatory tool sequences or model results.};
\end{tikzpicture}
\Description{C0 begins with final information. C1 receives an explicit correction and reaches the same final task obligations. Both apply to twenty-four task families. C2 makes the first call of a designated operation return an ambiguous status, only in twelve state-changing families. All models receive the same applicability and assignments.}
\caption{Matched-condition design. C0 begins with the final information. C1 begins with an earlier version and delivers an explicit correction attached to the result of a designated tool call; the final obligations equal those of C0. C2 applies to the twelve state-changing families and makes the first call of a designated operation return an ambiguous status. The sequences describe available information and events, not a required solution route or a result.}\label{fig:conditions}
\end{figure}
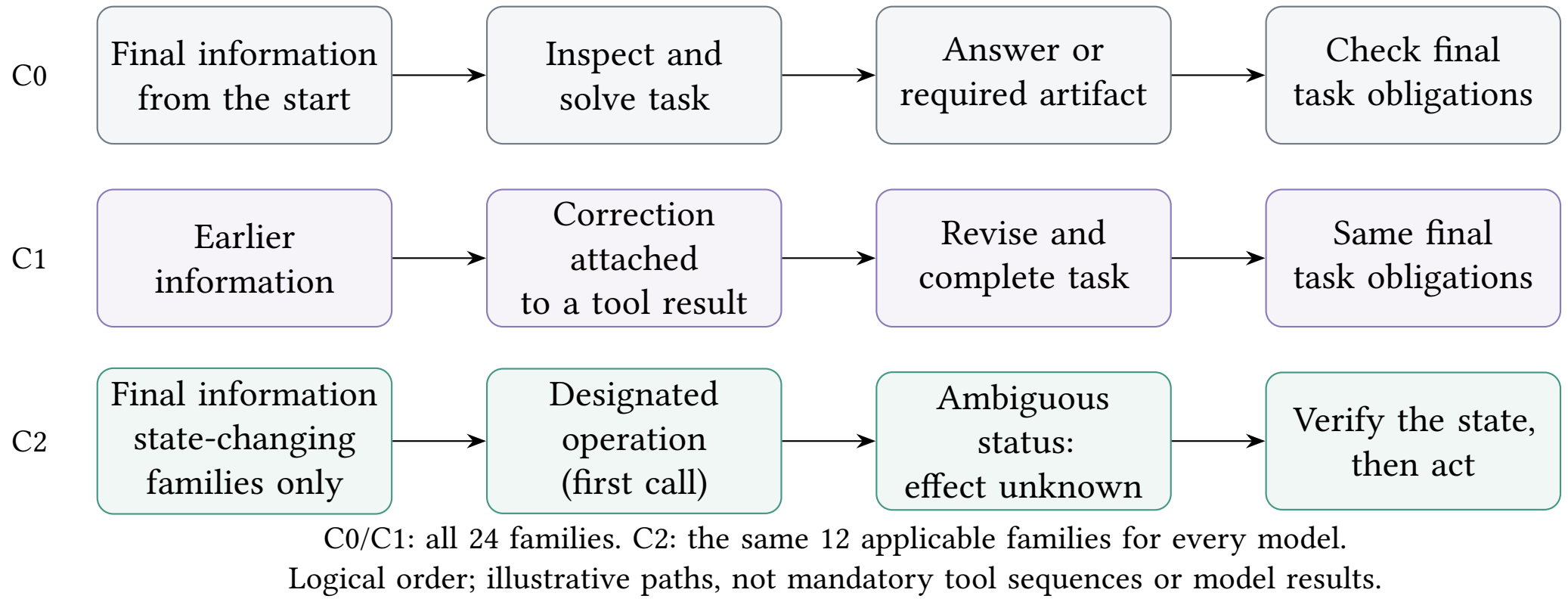

Figure~\ref{fig:conditions} shows the event sequences. Its horizontal axis is logical order, not elapsed time. The two-world construction of C2 is shown in more detail in Figure~\ref{fig:uncertain-worlds}.

\subsection{C0: the shared baseline}
A C0 episode supplies the final request, materials, rules, and tools from the start. It contains no correction and no ambiguous reply. The agent can inspect the materials it needs, use the allowed tools, submit its result, and stop within the resource contract. A normal C0 environment is not an oracle. A search can return irrelevant offers, a query can be rejected for a malformed argument, a test run can fail, and the task can contain deliberate distractors such as a cheap but non-compliant option. These belong to the task. What C0 excludes is only the specific intervention of C1 or C2.

The C0 population contains all twenty-four families, and its group profile is the primary view of performance. A configuration that fails an applicable task is scored by the task contract. An infrastructure failure is recorded as such and rerun once (Section~\ref{sec:protocol}), so that a server fault does not become an unexplained task failure.

\subsection{C1: a correction to the same final problem}
C1 begins with an earlier version of one piece of task information. A correction is delivered at a fixed point, and the final information and requested outcome are identical to those of C0. The agent does not have to guess that a correction exists: an explicit notice states what changed and where the current version can be read.

The trigger is the agent's own first use of the affected information. Depending on the family, it is the first read of a document, the first query of a table, the first call of a search or booking tool, or the first successful write of a source file. The notice is attached to the result of that tool call, and from then on \code{read\_task} shows the current rules and the notice. In most families the triggering call itself still returns the old content together with the notice, so the agent must read again to obtain the current version. A correction is delivered once per episode.

The corrections are native to each task. A conference moves forward by a day, a colleague's calendar entry was wrong, or a weekend budget is reduced. A return-policy clause, a stock level, or a customer's ticket is updated. A goods receipt records rejected units, an expense limit changes, or the requested quantity grows. A subscription's price change takes effect a month later, a charger's wattage is corrected, or an insurance section is superseded. A public guide is revised, a published source is corrected, or a screenshot turns out to show the wrong device. Meeting notes, a fact sheet, or an order table is corrected. In the data and software groups, a rule in the task specification is revised. Appendix~\ref{app:task-family-catalog} lists each correction with its trigger.

Every correction is constructed so that the solution under the old information fails the final scorer. Each family has a negative control that applies the superseded information and must fail on G. Otherwise C1 could be passed without using the correction at all. The final scorer is identical to C0's.

The match of final obligations is the central control. C1 never adds a deliverable, a tool, or a requirement absent from C0. The realized work can still differ: rereading a source or redoing a calculation costs requests and tokens, and that cost is part of the response being measured, so it is not removed afterwards.

The trigger point determines what C1 measures. In most families the correction arrives when the agent first touches the affected information, usually before it has committed to an action. C1 then measures whether the agent takes up an announced correction and redoes the affected reasoning, not whether it can undo actions already taken. In the software families the correction arrives after the first successful write of an editable file, so the agent must revise work it has already started. C1 does not test parameter learning, memory across sessions, or detection of an unannounced change.

\subsection{C2: effect and reply are different events}
C2 applies to the twelve families of the travel, service, finance, and software groups. In each family one operation is designated: booking a train journey, sending a meeting invitation, reserving a timed entry, submitting a refund, creating a reshipment, escalating a ticket, posting invoices, recording an expense decision, creating a purchase order, or submitting a code change request. Its first call returns an ambiguous status, \code{submitted}, \code{pending}, or \code{processing}, with a message saying that the result is not yet confirmed. In one hidden world the operation took effect; in the other it did not. Later calls of the same operation behave normally.

The operation is a necessary part of completing the task, not an artificial step added to make the condition applicable. Families whose work consists of reading and answering have no such operation, so C2 is not applied to them. Their cells are marked inapplicable rather than given an invented transaction.

Each C2 family offers an authorized way to establish the true state. Examples are listing the traveler's bookings and request results, reading the refunds on an order, reading a ticket's escalations, reading the ledger status of invoices, or listing submitted change requests. Two simple strategies each fail in one world. Retrying blindly in the world where the first call took effect produces a duplicate: two tickets, two refunds, two escalations, or two change requests. This violates P and in some families also G. Finishing without checking in the world where it did not take effect leaves the required effect missing, so G fails. Checking the state before deciding succeeds in both worlds. The benchmark scores the resulting state, the process constraints, and the report. It does not reward any particular action label.

\subsection{Hidden worlds and equal information}
The two worlds share the complete visible history up to and including the ambiguous reply. Its text and any request identifier it contains are identical in both worlds. The world cannot be inferred from the task introduction, the seed, a tool description, or the timing of the reply. The offline tests check this equality for every C2 family by comparing the serialized observations of both worlds (Section~\ref{sec:software-validation}). Each C2 family is run once in each world, and the assignment of worlds to the two runs is the same for every configuration. One configuration therefore cannot receive mostly committed outcomes and another mostly uncommitted ones.

An agent can take a different route and never call the designated operation, for example by stopping early. The ambiguous reply is then never produced, and none is fabricated. The episode stays in the C2 denominator, and whether the reply occurred is reported as a diagnostic (Section~\ref{sec:analysis}).

\subsection{Comparisons, order, and resets}
C1 is compared with C0 on all twenty-four families and C2 with C0 on the same twelve families. The study neither subtracts a twelve-family C2 average from a twenty-four-family C0 average nor includes a combined C1-plus-C2 condition. Every episode starts with a fresh agent session and a freshly built environment. No result, file, or message from one episode reaches another, and the condition order within each family block is fixed in the episode manifest and identical for every configuration.

\section{Benchmark Construction}\label{sec:construction}

\subsection{Scenario groups as the construction contract}
The inventory is organized around kinds of real requests rather than academic disciplines or the modules of one application. It has three categories and eight scenario groups. The daily category contains travel and scheduling, personal affairs and purchasing, and everyday inquiry. The enterprise category contains customer service and orders, finance and procurement, and office documents and collaboration. The technical category contains software engineering and data analysis. Each group contributes three formal families with different objectives, information structures, and checks. Within customer service, for instance, the families are a return-and-refund decision under a store policy, an order modification with a reshipment, and ticket triage with escalation. They share a setting but not a solution structure. Figure~\ref{fig:coverage} shows the groups and their condition applicability, and Appendix~\ref{app:task-family-catalog} specifies every family.

\begin{figure}[t]
\centering\begin{tikzpicture}[x=\linewidth,y=1cm,font=\footnotesize,
 box/.style={draw=vizAxis,rounded corners,align=center,text width=.19\linewidth,minimum height=1.45cm,inner sep=4pt}]
\node[font=\footnotesize\bfseries,anchor=west] at (-.01,.95) {Daily};
\node[box,draw=vizC2,fill=vizC2!8] at (.115,0) {Travel and scheduling\\three families\\C0, C1, C2};
\node[box,draw=vizP1,fill=vizP1!7] at (.365,0) {Personal affairs\\three families\\C0, C1};
\node[box,draw=vizP1,fill=vizP1!7] at (.615,0) {Everyday inquiry\\three families\\C0, C1};
\node[font=\footnotesize\bfseries,anchor=west] at (-.01,-1.05) {Enterprise};
\node[box,draw=vizC2,fill=vizC2!8] at (.115,-2) {Customer service\\three families\\C0, C1, C2};
\node[box,draw=vizC2,fill=vizC2!8] at (.365,-2) {Finance and procurement\\three families\\C0, C1, C2};
\node[box,draw=vizP1,fill=vizP1!7] at (.615,-2) {Office documents\\three families\\C0, C1};
\node[font=\footnotesize\bfseries,anchor=west] at (-.01,-3.05) {Technical};
\node[box,draw=vizC2,fill=vizC2!8] at (.115,-4) {Software engineering\\three families\\C0, C1, C2};
\node[box,draw=vizP1,fill=vizP1!7] at (.365,-4) {Data analysis\\three families\\C0, C1};
\node[align=left,text width=.21\linewidth,anchor=north west] at (.735,.7) {Teal: C2 applies (12 families).\\[3pt]Blue: C0 and C1 only (12 families).\\[3pt]One instance per family.\\[3pt]72 episodes per configuration.};
\node[align=center,text width=.96\linewidth] at (.49,-5.15) {Design coverage, not a measured capability profile. Every configuration receives the same applicable tasks.};
\end{tikzpicture}
\Description{Eight scenario groups in three categories each contain three formal families. Daily: travel and scheduling, personal affairs, everyday inquiry. Enterprise: customer service, finance and procurement, office documents. Technical: software engineering, data analysis. All families have C0 and C1; the travel, customer service, finance, and software groups also have C2. Colors show condition applicability, not performance.}
\caption{Design coverage. Eight scenario groups in three categories contribute three formal families each. The four groups with state-changing operations also receive C2. Applicability is identical for every model. The figure shows the design, not measured capability.}
\label{fig:coverage}
\end{figure}

This organization replaced an earlier academic-domain catalogue on 2026-11-12. The earlier catalogue had separate mathematics, natural-science, and multimodal groups. Under the new organization, document verification, conflicting-source checking, and itinerary planning were folded into the everyday and enterprise groups. Standalone mathematics, science, and visual-chart groups are no longer part of the inventory, and the study makes no claim about them. Only one family needs image content, and it receives that content as fixed text (Section~\ref{sec:construction-perception}).

\subsection{Original synthetic tasks}
All twenty-four formal families are original synthetic constructions. Their people, organizations, places, products, policies, and documents are fictional. The enterprise scenarios are generic processes, such as three-way invoice matching or ticket triage. They are not derived from any company's product, customer data, internal procedures, or roadmap. The insurance and medication-leaflet family tests only reading comprehension of the given documents. It neither asks for nor scores any personal legal or medical conclusion, and the leaflet states that it is a fictional document written for a reading test. All task materials, tool descriptions, and correction notices are in English.

The three software and three data families were carried over from an earlier internal version of the benchmark. Their data and corrections are regenerated from the same seed strings, and their interfaces follow the common tool conventions. The other eighteen families were written for this version. None of the tasks was public before the formal collection. Similarity to public tasks or to material seen in pretraining cannot be ruled out, however, so the results describe measured performance on these tasks rather than a proven transfer to unseen problems.

\subsection{Development and formal separation}
Eight development families, one per scenario group, exercise the runner and tools. They were used for the chain check and model registration and never enter a formal denominator. Each formal family must differ from its group's development family in entities, tools, numbers, and solution structure, and its task card states the difference. For example, the travel development family schedules three home services on one day with ordering dependencies. Its formal families instead book a multi-leg rail trip with a hotel and a disrupted return connection, schedule a meeting across time zones with room and invitation constraints, and plan a two-day itinerary under opening hours, travel times, and a budget.

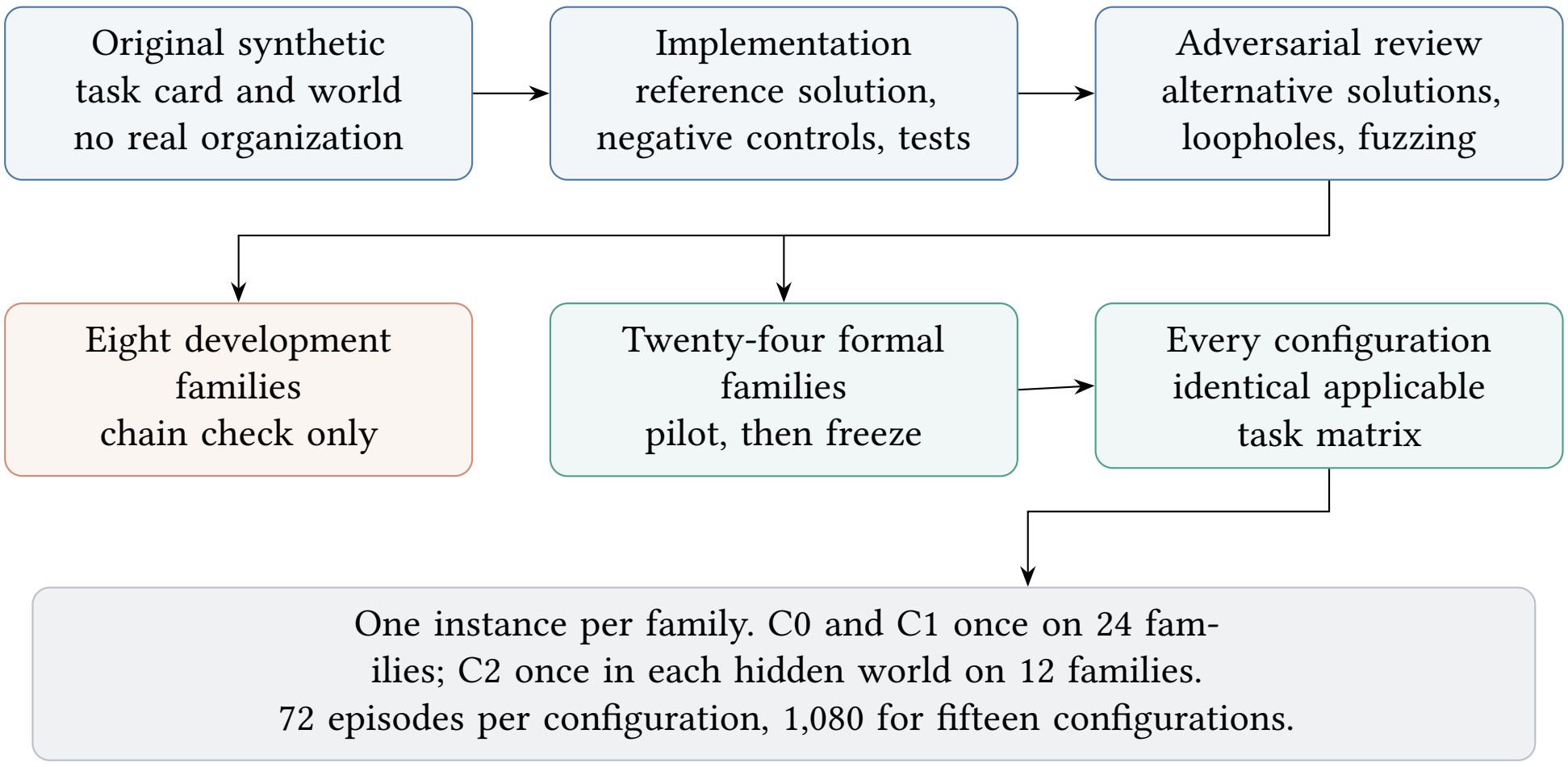
\begin{figure}[t]
\centering\begin{tikzpicture}[font=\small,>=Stealth,
 box/.style={draw=vizAxis,rounded corners,text width=.25\linewidth,align=center,minimum height=1.35cm,inner sep=5pt}]
\node[box,draw=vizP1,fill=vizP1!7] (source) {Original synthetic\\task card and world\\no real organization};
\node[box,draw=vizP1,fill=vizP1!7,right=.045\linewidth of source] (review) {Implementation\\reference solution,\\negative controls, tests};
\node[box,draw=vizP1,fill=vizP1!7,right=.045\linewidth of review] (structure) {Adversarial review\\alternative solutions,\\loopholes, fuzzing};
\node[box,draw=vizP2,fill=vizP2!9,below=1cm of source] (dev) {Eight development\\families\\chain check only};
\node[box,draw=vizP3,fill=vizP3!8,below=1cm of review] (test) {Twenty-four formal\\families\\pilot, then freeze};
\node[box,draw=vizP3,fill=vizP3!8,below=1cm of structure] (all) {Every configuration\\identical applicable\\task matrix};
\draw[->] (source.east)--(review.west);
\draw[->] (review.east)--(structure.west);
\draw[->] (structure.south)--++(0,-.45cm)-| (dev.north);
\draw[->] (structure.south)--++(0,-.45cm)-| (test.north);
\draw[->] (test.east)--(all.west);
\node[box,fill=vizMid,below=.9cm of test,text width=.86\linewidth] (matrix) {One instance per family. C0 and C1 once on 24 families; C2 once in each hidden world on 12 families.\\72 episodes per configuration, 1,080 for fifteen configurations.};
\draw[->] (all.south)--++(0,-.35cm)-| ([xshift=2cm]matrix.north);
\end{tikzpicture}
\Description{Each family starts as an original synthetic task card and is implemented with a reference solution, negative controls, and offline tests, then reviewed adversarially. Eight development families serve only the chain check; twenty-four formal families pass a pilot and are frozen. Every model then receives the same matrix of 72 episodes, 1,080 in total.}
\caption{Construction and evaluation separation. Each family is written as an original synthetic task card, implemented with a reference solution, negative controls, and offline tests, and reviewed adversarially. Development families serve only the chain check and registration; formal families pass a pilot and are frozen. Every configuration receives the same matrix.}
\label{fig:task-construction}
\end{figure}

With one instance per family, generalization in this study rests on the variety of families and groups, not on variation within a family. A family score is the outcome of a single task. A flaw in that task, or a task that is unusually hard or easy, has no second instance to compare against. Figure~\ref{fig:task-construction} shows the path from task card to the common matrix.

\subsection{Task cards}
Each family has a task card that is the binding specification for its implementation and for the second scorer. The card describes the scenario, the difference from the development family, the synthetic world with its deliberate distractors, the tools, the definitions of G, P, and E, the C1 correction and its trigger, the C2 operation and its verification route where applicable, a reference route, the negative controls, and known limitations. Known limitations are stated rather than designed away. For example, a card may note that the correction arrives before the agent usually commits, or that a tool's own validation makes one kind of policy violation impossible.

Tasks contain deliberate distractors that a careful agent must rule out. Examples are a train connection too short for the minimum transfer time, a cheap first-class fare forbidden by policy, a hotel outside the distance limit, a supplier quote that has expired, a duplicate invoice, a final-sale item, and a stale table version. A distractor counts only when the scorer's rule, not a preferred route, makes the distracted solution fail.

\subsection{Tools and common access}
All models receive the same tools and permissions for a task. Tools are specific to the family, for example searching rail journeys, reading a return policy, querying a table, reading a source file, running the public tests, or posting invoices. Every tool validates its arguments. A malformed call, an unknown identifier, or a refused operation returns an error message that says what was wrong, and the message becomes the agent's observation. A tool exception that is not such a validation error counts as a task-code defect (Section~\ref{sec:protocol}).

Tools expose only what the agent is allowed to see. The agent cannot read the scorer, the reference solution, hidden test cases, or the C2 world. In software families, model-written code runs only in a restricted executor in a separate process with CPU and wall-clock limits. Public tests are available to the agent; hidden tests and differential checks are used only by the scorer. A calculator or query tool removes unsupported mental arithmetic as a confound, but the agent still has to choose the right quantities, rules, and units.

\subsection{Fixed text transcription for image content}\label{sec:construction-perception}
One family, \code{inquiry-test-03}, involves a receipt photo, a device display photo, an app screenshot, and manual pages. No images are delivered and no vision model is used. Instead, the \code{view\_image} tool returns a fixed, pre-written text transcription of the requested region. The transcription is identical for every model and contains no answer that the task asks the agent to derive. The family therefore tests the use of image-derived information under a fixed transcription, not perception. The study makes no claim about native visual or multimodal capability, and Appendix~\ref{app:perception} describes the contract.

\subsection{Independent verification and negative controls}
Each scorer recomputes the expected outcome from the original task data and the final rules rather than calling tool code. A data scorer, for example, recomputes group means from the raw rows, and a code scorer runs public and hidden tests and, where specified, a differential test against the original program. Every family has negative controls. These are scripted wrong solutions, each built to fail on one named endpoint, such as applying the superseded C1 information, retrying blindly after an ambiguous reply, choosing the first listed offer, or misreporting a total. Section~\ref{sec:software-validation} reports the offline tests, the adversarial review, the pilot, and the second scorer.

\section{Experimental Protocol}\label{sec:protocol}
This section states the frozen protocol under which every formal episode was collected. The protocol was fixed in a run kit (version 13) before the first formal model request on 2026-11-21. Its runner and task code carry content fingerprints, and a formal run refuses to start when either differs from the frozen lock. Section~\ref{sec:collection} records how the collection was actually carried out, including deviations.

\subsection{Main matrix}
The inventory contains eight scenario groups with three task families each. Every family has one instance, so the twenty-four families define twenty-four distinct base tasks. Repetition depends on the condition. C0 and C1 run each task once. C2 applies to the twelve families in the four groups with genuine state-changing operations, and each of those tasks runs twice, once in each hidden effect world. Table~\ref{tab:main-matrix} gives the resulting counts.

\begin{table}[t]
\centering\small
\begin{tabular}{@{}lrrr@{}}\toprule
Component & Families & Episodes per family & Episodes\\\midrule
C0, per configuration & 24 & 1 & 24\\
C1, per configuration & 24 & 1 & 24\\
C2, per configuration & 12 & 2 (one per hidden world) & 24\\\midrule
All conditions, one configuration & 24 & & 72\\
Fifteen OpenHands--model configurations & 24 & & 1,080\\\bottomrule
\end{tabular}
\caption{Main matrix. Eight scenario groups with three families each and one instance per family give twenty-four base tasks. C2 runs twice per applicable task so that each hidden world occurs exactly once. Condition variants and C2 repeats do not add distinct base tasks.}\label{tab:main-matrix}
\end{table}

The 1,080 episodes are not 1,080 independent problems. Condition variants and the second C2 run reuse the same twenty-four base tasks. No task is run twice under the same condition and the same hidden world, so the study does not estimate rerun variability. Each family score rests on a single task, and within one configuration each C1 or C2 contrast has one observed pair per family. These are design costs accepted to make twenty-four new families feasible; Section~\ref{sec:discussion} discusses their consequences. On each device the five models run one after another. Within each configuration, the runner first completes the sixty episodes of the first round (all C0 and C1 episodes and the first C2 world), in a fixed order that groups the conditions of each family, and then the twelve second-world C2 episodes. The order is the same for every configuration.

\subsection{Configurations, operators, and devices}
Execution is divided by model, not by task. The prespecified assignment gives each of three operators five models on one registered device and all seventy-two task keys for each model. No operator chooses a different subset of tasks. Actual deployment deviations are documented in Section~\ref{sec:collection}. Table~\ref{tab:models} lists the fifteen configurations. All weights are Q4\_K\_M quantizations served by Ollama, and every model digest is checked against the frozen roster before its first episode.

\begin{table}[t]
\centering\footnotesize
\begin{tabularx}{\linewidth}{@{}llYcc@{}}\toprule
ID & Developer & Ollama tag & Default reasoning & Temp.\\\midrule
L01 & Qwen & \code{qwen3.6:35b-a3b-q4\_K\_M} & yes & 1.0\\
L02 & Google & \code{gemma4:26b-a4b-it-q4\_K\_M} & yes & 1.0\\
L03 & IBM & \code{granite4.2:30b-q4\_K\_M} & yes & 1.0\\
L04 & Liquid AI & \code{lfm2.5:24b-a2b-q4\_K\_M} & yes & 0.1\\
L05 & Cohere & \code{north-mini-code-1.0:q4\_K\_M} & yes & 1.0\\\midrule
L06 & Z.ai & \code{glm-4.7-flash:q4\_K\_M} & yes & 1.0\\
L07 & NVIDIA & \code{nemotron-3.5-lightning:30b-a3b-q4\_K\_M} & yes & 1.0\\
L08 & Google & \code{gemma4:31b-it-q4\_K\_M} & yes & 1.0\\
L09 & Qwen & \code{qwen3.6:27b-coding-mtp-q4\_K\_M} & yes & 0.6\\
L10 & Google & \code{gemma4:12b-it-q4\_K\_M} & yes & 1.0\\\midrule
L11 & Qwen & \code{qwen3.5:9b-q4\_K\_M} & yes & 1.0\\
L12 & Google & \code{gemma4:e4b-it-q4\_K\_M} & yes & 1.0\\
L13 & IBM & \code{granite4.2:8b-q4\_K\_M} & yes & 1.0\\
L14 & NVIDIA & \code{nemotron-3-nano:4b-q4\_K\_M} & yes & 1.0\\
L15 & Liquid AI & \code{lfm2.5:8b-a1b-q4\_K\_M} & no & 0.2\\\bottomrule
\end{tabularx}
\caption{Registered configurations. L01--L05 run on an Apple M4 Max (64 GB, macOS, Metal), L06--L10 on an Apple M5 Max (64 GB, macOS, Metal), and L11--L15 on rented NVIDIA RTX 3090 GPUs (24 GB, Linux, CUDA). ``Default reasoning'' records whether the model emitted reasoning output in its registration episode without any reasoning parameter being sent. ``Temp.'' is the temperature in the model's own Ollama parameters; other sampling values are registered with each run.}\label{tab:models}
\end{table}

The fifteen configurations are fifteen registered model--backend--hardware combinations. They are not fifteen independent model lineages: the roster contains four Gemma 4 models, three Qwen models, two Granite 4.2 sizes, and two LFM2.5 sizes. Each model runs on the device of its operator, so differences between models on different devices combine model, backend, and hardware differences. Time and resource comparisons are therefore made only within a platform.

\subsection{OpenHands framework and task tools}
Every configuration uses the official OpenHands Software Agent SDK, version 1.61.0, headlessly through its \code{Agent} and \code{Conversation} classes~\citep{wang2025openhands,openhandsSDK}. Ollama 0.40.14 serves the model. The SDK reaches Ollama through LiteLLM 1.105.7 with the \code{ollama\_chat} provider. The runner starts its own Ollama server on a separate port and sets the context length there, so the operator's own Ollama installation is not used.

The agent receives only the tools of its task. Each environment operation is registered as an SDK client tool whose executor calls the task environment and returns the environment's actual result or error message~\citep{openhandsCustomTools}. The SDK's default tools are disabled, so the agent has no terminal, file editor, browser, or network access. Skills, persistent memory, subagents, and context condensation are not enabled. A task's tool set is the same for every model and every condition. The configuration is therefore OpenHands with benchmark-specific tools, not the unmodified OpenHands product.

Each episode begins with a single user message: an instruction to solve the task using only the provided tools, start with \code{read\_task}, follow the specification notice and current rules, and end by calling \code{finish}, followed by the task's public introduction. The system prompt is the SDK's own. The agent loop then runs without further user turns. No reminder is added when a model replies with text only.

\subsection{Vendor-default reasoning and sampling}
The main policy is vendor-default reasoning: no reasoning parameter is sent. The runner sends neither Ollama's \code{think} field nor a reasoning-effort level, and sets the SDK's reasoning effort to an explicit null so that no SDK default is filled in. A metering proxy between the SDK and Ollama logs every request. If any request in a batch carries a reasoning parameter, the batch stops. Each model's default behavior was registered before freezing in one development episode per model (Table~\ref{tab:models}): fourteen models emitted reasoning by default and L15 did not.

Vendor default is a configuration choice, not a claim that computation is equated across models. A model that reasons by default spends more of its output budget and more time. Reasoning text is retained and counted toward the output budget. It is never executed as an action and is not accepted as evidence of task completion.

Sampling likewise uses each model's own Ollama parameters, including temperatures between 0.1 and 1.0. The runner adds only the output limit and a per-episode seed. The seed is the first four bytes, read big-endian, of the SHA-256 hash of a fixed prefix followed by the task key. The task key contains family, instance, repeat, and condition, so the same key receives the same seed for every model and different keys receive different seeds. Equal seeds do not make trajectories comparable across models or platforms. With non-zero temperatures and GPU numerics, they also do not guarantee identical output on different hardware.

\subsection{Resource contract}
Table~\ref{tab:budget} lists the frozen limits. They apply uniformly to all fifteen configurations and all tasks. The token and call limits are the widest of the candidate tiers listed before collection. The earlier plan of calibrating them on development episodes was cancelled on 2026-11-19. On 2026-11-21, before freezing, the two output limits were each raised by one tier after development episodes showed that the then-widest tier truncated successful work of models that reason by default. The limits are therefore fixed ceilings, not values calibrated to be just sufficient.

\begin{table}[t]
\centering\small
\begin{tabular}{@{}lr@{}}\toprule
Limit & Value\\\midrule
Active context & 32,768 tokens\\
Output per model request (reasoning and action) & 16,384 tokens\\
Output per episode & 98,304 tokens\\
Model requests per episode & 48\\
Executed tool calls per episode & 64\\
Time per model request & 3,710 s\\
Time per episode & 22,440 s\\\bottomrule
\end{tabular}
\caption{Frozen resource contract. Output limits include reasoning tokens. The two time limits are safeguards derived from the slowest registered model so that every model can use its full token budget; they are not expected durations.}\label{tab:budget}
\end{table}

The context limit cannot exceed the smallest native context in the roster and must fit on the 24 GB devices. The time limits are derived from the slowest measured generation and prompt-processing speeds in the registration episodes (L08: 6.9 and 363 tokens per second). The episode limit is at least 1.5 times the time that model needs to generate the whole output budget and process the largest observed prompt volume, and the request limit is derived the same way for one request. Because the time limits are set by the slowest model, hardware speed should not usually turn into a difference in completion. Episodes that end on a time limit are reported separately.

Budgets are counted from actual requests and tool executions, not from SDK iterations. The proxy counts every model request and its output tokens and refuses a request that would exceed a limit. The tool guard counts every tool call that reaches the environment. Tool calls that the SDK rejects before execution, such as calls to an unknown tool or with undeclared arguments, are logged separately. The SDK's own retries are disabled, so a failed request remains a failed request.

\subsection{Termination and infrastructure reruns}
Every episode receives exactly one terminal reason, assigned by a fixed rule from the request log and the episode summary. An episode ends as \code{submitted} when the agent calls \code{finish}. The other model-behavior reasons are:
\begin{itemize}
\item \code{text\_reply\_no\_tool\_call}: the last reply contains text but no tool call;
\item \code{output\_truncated\_no\_action}: the last reply was cut off by the output limit before any action;
\item \code{tool\_call\_parse\_error}: the server could not parse the model's tool call;
\item \code{stuck\_detected}: the SDK's stuck detector ended the conversation after repeated identical actions or errors;
\item \code{budget\_requests}, \code{budget\_episode\_output}, \code{budget\_tool\_calls}: a resource limit was reached;
\item \code{wall\_clock}, \code{request\_timeout}: a time limit was reached with a scorable state.
\end{itemize}
All of these are scored from the final state and count in every denominator. Following the rule fixed before collection, a text-only reply ends the episode as incomplete with its own terminal reason, without an added reminder, and its rate is reported for each model.

Infrastructure reasons are \code{server\_error}, \code{framework\_error}, \code{wall\_clock\_killed} (no scorable state remained), and \code{unclassified}. An episode with an infrastructure reason is rerun once with the same seed, and both records are kept. A \code{tool\_exception}, an exception raised by task code rather than a rejected argument, is a task-code defect rather than a model outcome. Formal episodes run strictly one at a time on each device. An interrupted run resumes at the first episode without a scored summary and refuses to resume if the code, dependencies, or environment differ from those at the start.

\subsection{Hidden C2 worlds}
For each C2 family, one of its two runs takes place in the world where the ambiguous operation took effect and the other in the world where it did not. Which run receives which world is derived from a private salt held by the coordinator. The salt is not in the repository or any package; the lock records only its SHA-256 hash, and a formal run with a different salt is refused. The assignment depends only on the task, so every configuration receives the same world for the same task key. Agents cannot infer the world from the seed or from any observation before the ambiguous reply, which is identical in both worlds.

\subsection{Freeze and what it binds}
The lock binds the runner and task-code fingerprints, the OpenHands SDK, LiteLLM, and Ollama versions, the resource contract, and the salt hash. The task inventory and episode order are generated from the planning package and are part of the runner fingerprint. Model weights are bound by the roster digests. The endpoints, condition contrasts, and denominators in Section~\ref{sec:analysis} were recorded in the decision document on 2026-11-13, before freezing. After the freeze, runner and task code were not changed. Any change would require a documented reason and a new freeze, and results under different locks would not be pooled.

\section{Task Validation and Freeze}\label{sec:software-validation}
Before any formal request, the task code and runner went through four kinds of checks. Offline tests check each family against its reference solution and negative controls. Adversarial review looks for loopholes and over-strict scoring. A pilot run with a model outside the roster exercises all formal task keys. A freeze then binds the result. This section specifies the scope of each validation layer. These checks are separate from the registered configurations and their formal results. The second scorer, described last, was applied to the formal records after collection.

\subsection{Offline family tests}
Each family is one module that defines its synthetic world, its tools, its scorer, a reference solution, and a list of negative controls. The reference solution uses only the tools an agent can call. For every family and every applicable condition, with C2 tested in both hidden worlds, the offline suite checks the following properties.
\begin{itemize}
\item The reference solution reaches $J=1$ without any tool error, in 10 to 18 tool calls, within the smallest candidate budget tier.
\item Calling \code{finish} immediately does not count as completion.
\item Each negative control fails on exactly the endpoint it targets. Formal families have four to ten negative controls each, 147 in total. They include, for every family, a control that applies the superseded C1 information and, for every C2 family, a blind retry in the committed world and a finish without verification in the uncommitted world.
\item The C1 correction is delivered exactly once and remains visible through \code{read\_task}.
\item In C2, the ambiguous reply and every observation before it are identical in both worlds, and the opening information does not reveal the world.
\item Tool names are unique across families and every tool has a description.
\end{itemize}
The suite has 861 checks. All of them pass on Python 3.9, 3.13, and 3.14. Agreement between a scorer and its reference solution establishes internal consistency. The separate scoring implementation described below additionally checks the recorded outcomes against the task cards.

\subsection{Adversarial review and interface testing}
After implementation, four reviewers examined the families adversarially. They wrote alternative legitimate solutions to find over-strict scoring, searched for shortcuts that pass the scorer without meeting the task, and checked the C1 and C2 mechanics. All findings were fixed and re-tested. Every tool was also fuzzed with ill-typed and extreme arguments (21,760 calls). The only exception raised was the intended \code{ValueError}, whose message the runner returns to the agent as an observation.

\subsection{Pilot over the formal task keys}
Before freezing, a model outside the roster (\code{qwen3:8b}) ran all 72 formal task keys through the actual runner. The pilot looked only for task-code defects: tool exceptions, unclassified terminations, infrastructure failures that persisted after a rerun, and C1 corrections that were not delivered. Its scores were neither inspected nor reported, and the pilot does not belong to any registered configuration. No such defect occurred. The pilot did show two error messages that did not say how to fix a rejected argument: the list of setting names in \code{inquiry-test-03} and the filter and aggregation errors of the data query tool. The model repeated the same malformed call several times in response. Both messages were rewritten without changing any scoring rule. All 72 pilot episodes were then replayed through the new code to confirm that only the expected observations changed, and the six affected task keys were rerun.

\subsection{Development stages and registration}
Each device ran two development stages with the actual runner. The chain check ran three development episodes covering a C1 correction and both C2 worlds. The registration stage ran one development C0 episode per model to record whether the model reasons by default and to measure its generation and prompt-processing speeds. These episodes use the eight development families, which are separated from the formal families in content and solution structure. They are kept apart from formal records and are not results. A planned budget calibration on development episodes was cancelled in favor of the fixed widest tier (Section~\ref{sec:protocol}).

\subsection{Freeze}
The coordinator froze the kit on 2026-11-21 at 18:53 (Mountain Time). The freeze took the three registration outputs as input, checked their completeness and code identity, derived the time limits from the slowest model, generated the C2 salt, wrote the lock, and built the formal kit. From the unpacked kit, the formal preflight reported zero problems for all three operators with 72 task keys for each of five models, a different salt was refused, and all 861 offline checks passed. After the freeze, neither runner nor task code changed.

\subsection{Second scorer}\label{sec:verification-status}
The protocol assigned a task-card-based second scorer to operators A and C, split by scenario group and compared with the original on the recorded formal states. On 2026-11-26, with that implementation outstanding, the coordinator assigned it to the AI coding assistant used in the coordinator's work (Claude Code, Anthropic), which had also helped build the task code and original scorers. The scoring logic was written separately; the implementations share task data and the restricted executor for code tasks. This is a separate implementation check, rather than an external or fully blinded replication.

The inputs were restricted to the task cards, the final-state snapshot and event log of each episode, and the world data of the task code, read by importing its data constants. For the three software families, the scorer runs the submitted code in the task's restricted executor and compares the outputs with its own expected outputs. The original scoring functions and the original per-episode scores were not read before the comparison. Three exposures to withheld information are recorded with the implementation. First, before writing the scorer, the assistant had seen the per-family pass counts of the original scores while preparing figures. During implementation, its own per-family goal counts matched these totals, and no code was changed in response. Second, it saw one original episode score while inspecting the record format. Third, the data constants it read include the code text of some reference implementations and the answer constants of one family. The code was frozen, with its hash recorded, after all twenty-four families were written and before the first comparison.

The first comparison over all 1,080 episodes agreed on every G and J value, on 1,079 P values, and on 1,069 E values. The twelve disagreements were resolved against the task cards (Appendix~\ref{app:verifier-validation}). In nine, the second scorer had misread the card. It was corrected and frozen again, and each change is logged with its reason. For three E values, the original scorer was more lenient than the card. In one purchasing episode, the report described a cheaper quote as more expensive. In two expense-review episodes, an earlier effective decision on one report cited a policy clause that does not apply to it. The coordinator adopted the card-based outcome for these three episodes. The frozen scorer cannot be changed, so the rescoring is applied in the analysis, and the episodes are listed in Table~\ref{tab:disagreements}. Only E changes. G and J are unchanged because each of these episodes had already failed G or P. Among the reported summaries, only the C0 evidence score of L08 changes, from 23 to 22 of 24 tasks.

After the corrections, the two scorers agree on all 1,080 G and J values, on all 930 applicable P values (both mark P as inapplicable in the other 150 episodes), and on 1,077 E values; the other three are the adopted rescoring. Two families with unusual outcome patterns were examined with priority: \code{data-test-01}, completed in C0 by one configuration, and \code{travel-test-03}, with more C2 than C0 successes. In both, the second scorer reconstructed every G value from the recorded states and events. No episode was rerun because of the comparison. The agreement establishes outcome consistency for the collected records. The shared task interpretation and implementation history remain relevant to validity: a specification error common to both implementations requires additional task review rather than agreement testing alone.

\section{Collection Record}\label{sec:collection}
This section records how the formal episodes were collected and where the collection departed from the protocol. It is written from the operators' run logs and from an independent check of each returned archive by the coordinator. The check runs the frozen report tool over the returned outputs and inspects every episode record. It confirms that all assigned episodes are present, that every episode carries the frozen runner and task-code fingerprints, that no reasoning parameter was sent, and that the context and output settings match the lock. Table~\ref{tab:collection} summarizes the state at the time of writing.

\begin{table}[t]
\centering\small
\begin{tabularx}{\linewidth}{@{}lYrr@{}}\toprule
Models & Hardware as run & Episodes & Infrastructure reruns\\\midrule
L01--L05 (A) & One Apple M4 Max, 64 GB, macOS & 360 / 360 & 0\\
L11--L15 (B) & Five rented RTX 3090 instances on two machines, 24 GB, Linux, two driver versions & 360 / 360 & 0\\
L06--L10 (C) & One Apple M5 Max, 64 GB, macOS & 360 / 360 & 0\\\bottomrule
\end{tabularx}
\caption{Collection state. All 1,080 retained episodes carry the frozen fingerprints, and none ended with an infrastructure reason or a task-code exception. The combined report confirms complete coverage of the assigned matrix. Interrupted and excluded records are described below.}\label{tab:collection}
\end{table}

\subsection{Apple M4 Max arm (L01--L05)}
All 360 episodes ran on one machine, one at a time, between 2026-11-21 23:09 and 2026-11-23 14:23 (Mountain Time), using 39.2 machine-hours. Because the machine was also the operator's workstation, the job ran at a reduced macOS scheduling class (utility quality of service). This class affects scheduling only, not the model or its settings. A first start at background priority was stopped after the first model's generation rate fell to about 14 tokens per second, which risked the frozen time limits. The interrupted episode had no scored summary, so the restarted run executed it from the beginning with the same seed. Its partial request log is retained under a separate name.

From their eighth episode on, L01 and L02 generated at roughly 55\% of the rate measured in their registration episodes (about 53 and 56 tokens per second). The cause was not identified, and no setting was changed during the run. L03 to L05 ran at 21, 80--100, and 58 tokens per second. The longest episode took 3,277 seconds, far below the 22,440-second limit, and no episode ended on a time limit.

\subsection{Apple M5 Max arm (L06--L10) and two host pauses}
All 360 episodes ran on one machine, one at a time, between 22:38 UTC on 2026-11-22 and 10:10 UTC on 2026-11-25, using 59.5 machine-hours with no idle time between episodes. The coordinator regenerated the arm with the real salt and obtained a manifest identical to C's, with the same task order as the other arms. Measured from the server's own generation timings, L06 to L10 generated at about 21, 54, 16, 19, and 32 tokens per second. Apart from the two requests described next, the longest request took 1,121 seconds. The longest episode took 5,983 seconds, inside the 22,440-second limit.

The request log records two durations for every request, the wall time measured by the metering proxy and the total duration reported by the inference server. In every other successful request of the three arms, the two durations agree to within a minute. In two requests of this arm they differ by much more. For request index 3 of \code{L09.data-test-02.i1.r1.C0} the wall time was 3,166 seconds and the server time 435 seconds. For request index 4 of \code{L10.data-test-01.i1.r1.C0} they were 3,228 and 811 seconds. The two events are more than seven hours apart and indicate host pauses of about 46 and 40 minutes. Operator C's accompanying note confirms that the L10 episode overlapped a lid-closed, battery-powered sleep from 17:05:55 to 17:47:24 Mountain Time on 2026-11-24. The L09 pause is identified from the timing discrepancy; its cause has no corresponding operator confirmation. The run instructions required the host to be kept awake, so both pauses are recorded as deviations.

The coordinator retained both episodes under the frozen rerun rule. The recorded inputs and seed were unchanged, and even with the pause included, neither request reached the 3,710-second request limit or the episode limit. Both episodes ended for model-behavior reasons, \code{submitted} and \code{stuck\_detected}, for which the frozen rules provide no rerun. Their recorded elapsed times include the pauses; the resource report marks both episodes and excludes their elapsed times from time comparisons.

\subsection{RTX 3090 arm (L11--L15) and the multi-instance deviation}
The protocol assigns each model to one registered device. Operator B instead ran the formal stage on five rented RTX 3090 instances in parallel, each with its own GPU and strictly single-stream, between 05:16 and 13:46 UTC on 2026-11-22 (35.6 machine-hours). Runner and task code were not modified. Placeholder records made the runner on each instance skip episodes assigned to other instances, and the five outputs were then merged. The coordinator regenerated the arm with the real salt and obtained a manifest identical to B's. This confirms the correct salt and the same C2 world assignment as the protocol. The longest single request took 2,080 seconds and the longest episode 2,090 seconds, both well inside the limits.

The five instances were on two physical machines: two instances on a machine in Taiwan with NVIDIA driver 580.173.02 and three on a machine in Argentina with driver 615.71.09. L12 and L14 ran entirely on one instance in Argentina. L11 ran 36 episodes in Taiwan and 36 in Argentina, L13 ran 35 and 37, and L15 ran 18 and 54. The instance of every episode is recorded in a per-episode machine map, and each instance's environment record is retained.

Operator B also compared duplicate runs. A handover between instances caused eleven episodes to run twice. The kept copy of each was chosen by the instance assignment table fixed before the runs, independent of the outcomes. The eleven other copies are excluded from analysis. Operator B reports retaining those copies, and the coordinator has requested them for archival completion; they are not part of the received 1,080-episode analysis archive. In B's comparison, the three duplicate pairs from the same instance were token-for-token identical. All eight pairs that crossed the two driver versions differed. In three of them the terminal reason differed, and in five the score or the correction-delivery flag differed. These pairs demonstrate fixed-seed agreement within the sampled instance and divergence across the sampled drivers.

The coordinator retained the arm with the documented deviation. For L11, L13, and L15, each model's results mix two driver configurations. L12 and L14 each ran on one instance, although it was rented rather than the originally registered device. The analysis therefore treats these as actual serving configurations and omits cross-platform time or resource comparisons with the RTX 3090 arm.

\subsection{Tool-call format of L14}
Sixty-four L14 episodes ended as \code{text\_reply\_no\_tool\_call}. In 60 of them the final reply's text was itself a tool call. It was written in an XML-like format, an opening \code{tool\_call} tag followed by \code{function} and \code{parameter} tags, often ending after the function name or one parameter. Ollama 0.40.14 did not parse it into a tool call. In the remaining four, the text was a bare JSON object of tool arguments without a function name. Across all 277 L14 replies, 202 contained tool calls that were parsed normally. The model therefore usually used a recognized format and occasionally switched to the unrecognized one, which ended the episode.

Under the frozen termination rule these episodes remain \code{text\_reply\_no\_tool\_call} and retain their observed scores. L14's results primarily characterize compatibility between the model's tool-call format and the parser of this Ollama version. Six L14 request logs end with a successful status line without reply content. These partial response captures are recorded as logging gaps; the associated episode terminal reasons are retained, with no reclassification based on those gaps.

\subsection{Output and time limits}
The per-episode output limit of 98,304 tokens ended 18 episodes of the RTX 3090 arm, 3 of the M4 Max arm, and 2 of the M5 Max arm. These episodes are retained as observed outcomes under the frozen budget. No episode in any arm ended on a request or episode time limit.

\subsection{Scoring after collection}
All three arms were returned, checked, and combined, and the combined report covers all 1,080 assigned episodes. The second scorer (Section~\ref{sec:verification-status}) was then applied to every formal record. It led to the rescoring of three evidence outcomes and to no rerun. Section~\ref{sec:results} reports the results.

\section{Estimands and Reporting Plan}\label{sec:analysis}

\subsection{When the analysis was fixed}
The primary endpoint, the two condition contrasts, their denominators, and the treatment of untriggered episodes were recorded in the study's decision document on 2026-11-13, before the freeze and before any formal request. This section restates them for the final design. The text was written on 2026-11-24, after two of the three arms had been collected and their summary scores viewed by the coordinator. It was updated on 2026-11-26, after the third arm had been collected and its scores also viewed. That update changed only the collection status and the marking of paused episodes in the resource report. The original endpoints, contrasts, weights, and assigned denominators remain unchanged. Their results in Section~\ref{sec:results} were computed on 2026-11-26, after the second-scorer comparison; the figures were selected after the scores had been seen. The C2 hidden-world split and the relation of completion to resource use are exploratory. On 2026-11-28, a further post-hoc reanalysis of the existing C2 events added operation-matched follow-up categories (Section~\ref{sec:c2-recovery}; Appendix~\ref{app:c2-operation-verification}). This addition uses the frozen records without new model calls and changes no original score or denominator.

\subsection{Primary endpoint: C0 goal completion}
Let $Y_{mfc}\in\{0,1\}$ be an endpoint for configuration $m$, family $f$, and condition $c\in\{\mathrm{C0},\mathrm{C1}\}$, and let $Y_{mf\mathrm{C2}w}$ be the C2 outcome in hidden world $w\in\{0,1\}$. Each of these is a single episode. Let $\mathcal G$ be the eight scenario groups and $\mathcal F_g$ the three families of group $g$. The primary score is the group-balanced C0 goal-completion rate
\begin{equation}
 S_G(m)=\frac{1}{|\mathcal G|}\sum_{g\in\mathcal G}\frac{1}{|\mathcal F_g|}\sum_{f\in\mathcal F_g} G_{mf\mathrm{C0}} .
 \label{eq:macro}
\end{equation}
Every group has three families, so $S_G(m)$ equals the plain mean over the twenty-four families; the group weighting matters only if a family becomes unscorable. The eight group scores and the twenty-four family outcomes are always reported beside the summary, so a broad average cannot hide concentrated success or failure. With twenty-four tasks, one task moves the summary by about 4.2 percentage points, and neighboring configurations may not be distinguishable.

\subsection{Condition contrasts}
Both contrasts compare each condition with the C0 episode of the same family:
\begin{equation}
 \Delta_1(m)=\frac{1}{24}\sum_{f\in\mathcal F}\bigl(Y_{mf\mathrm{C1}}-Y_{mf\mathrm{C0}}\bigr),\qquad
 \Delta_2(m)=\frac{1}{12}\sum_{f\in\mathcal F_{\mathrm{C2}}}\Bigl(\tfrac12\sum_{w\in\{0,1\}}Y_{mf\mathrm{C2}w}-Y_{mf\mathrm{C0}}\Bigr).
 \label{eq:contrast}
\end{equation}
$\mathcal F_{\mathrm{C2}}$ contains the twelve families of the travel, service, finance, and software groups. The C2 contrast therefore compares those twelve families with their own C0 episodes, and the restricted C0 baseline is shown next to it. Subtracting the twenty-four-family C0 score from a twelve-family C2 score would mix the intervention with a change of task population.

Within one configuration, each C1 contrast rests on one pair per family and each C2 contrast on one C0 episode and two C2 episodes per family. No rerun variability is estimated. Per-configuration contrasts are therefore reported as exact counts of improved, unchanged, and worsened families rather than as tests. The main interpretation of the conditions is pooled across configurations. It reports, over all configurations, how many family pairs changed from success to failure and how many from failure to success. Positive, zero, and negative contrasts are all possible, and no direction is built into scoring or presentation.

\subsection{Assigned denominators and trigger diagnostics}
All assigned episodes stay in their condition's denominator, including C1 episodes in which the correction was never delivered and C2 episodes in which the ambiguous operation was never attempted. Reaching the trigger depends on the agent, so restricting a contrast to triggered episodes would compare a different, self-selected group for each configuration. Trigger and delivery are reported as diagnostics with their own denominators. For C1 these are the share of episodes in which the correction was delivered. For C2 they are the share in which the ambiguous reply occurred, a tool-name query proxy, and the share of committed-world episodes with a process violation, which includes every duplicated effect. The query proxy detects a successful call to a designated verification-tool name after the ambiguous reply and before another successful call of the designated operation's name. It does not match the affected object or inspect the returned state, so it is not an operation-level verification rate. A triggered-subset contrast may be shown as a labeled diagnostic but never replaces the assigned-condition contrast.

\subsection{Process, evidence, and joint outcomes}
P, E, and J are reported with their applicable denominators. E is defined for all 24 families and P for 19. P does not apply to \code{data-test-01}, \code{data-test-03}, and the three \code{inquiry} families, which ask for a submitted answer and have no process constraint; in these five families J reduces to $G\wedge E$. On records where G, P, and E are all defined, outcomes are partitioned into joint success, goal reached with a process violation, goal reached with conforming process but insufficient evidence, and goal not reached. A diagnostic always names the task predicate it comes from. A failed goal is not by itself attributed to a reasoning deficit, and a correct final state is not by itself taken as supported by the agent's observations.

\subsection{Terminal reasons and resources}
Terminal reasons (Section~\ref{sec:protocol}) are reported per configuration, including the rate of text-only replies, which the frozen rule requires. For text-only terminations, the final reply is classified as a tool call in an unparsed format (XML-like or bare JSON), other text such as a final summary, or an empty reply. This classification is descriptive. It does not change any score, and it is reported because the L14 parser incompatibility (Section~\ref{sec:collection}) shows that the same terminal reason can arise in different ways.

Resource use is reported per episode as model requests, executed tool calls, output tokens, reasoning characters, and elapsed time. Failed and truncated episodes are included. Time and throughput are compared only within a platform. For the RTX 3090 arm they are not used for any comparison, because machines differed within a model. Requests whose wall time far exceeds the server's own duration are flagged as host pauses. The two such episodes (Section~\ref{sec:collection}) are marked, and their elapsed times are not used in any time comparison.

\subsection{Sensitivity and multiplicity}
A family-resampling analysis resamples whole families within each scenario group and carries all of a family's episodes together~\citep{efron1979bootstrap}. It shows how much the summaries depend on the particular family mix. With three families per group and one instance per family, it is not a confidence statement about a population of future tasks. Exact family outcomes and leave-one-family-out summaries are reported beside it.

Two further sensitivity views are reported. The first separates L11, L13, and L15 by driver version. Because their episodes were split between drivers by instance availability rather than at random, this view is descriptive only. The second shows the pooled condition contrasts with L14 excluded, because its outcomes mainly reflect the parser incompatibility. The primary summaries always include all configurations.

The design yields fifteen C1 and fifteen C2 contrasts and many pairwise configuration differences. They are presented as estimates with exact counts, not as a family of significance tests. If a test is reported, its assumptions are stated and a Holm adjustment is applied across the declared contrast family~\citep{holm1979multiple}.

\subsection{Coverage and completeness}
Every summary is checked against the assigned ledger of 1,080 episodes. Completed failures, infrastructure terminations, never-started episodes, and inapplicable cells are distinct states. An episode that exhausts its budget is an observed failure, not a missing value. If an episode remains unscorable after its infrastructure rerun, summaries report lower and upper bounds that set it to failure and to success (Appendix~\ref{app:analysis-details}). In the collected data, no episode is missing or unscorable.

\section{Results}\label{sec:results}
This section reports the 1,080 formal episodes as scored after the second-scorer comparison (Section~\ref{sec:verification-status}), including the three evidence outcomes rescored there. All numbers are computed by the analysis script from the returned outputs, and Appendix~\ref{app:results} gives the full tables and further figures. The summaries follow Section~\ref{sec:analysis}, and views that the plan does not list are marked as exploratory. A configuration is one model on its registered platform (Section~\ref{sec:discussion}), so a difference between configurations on different platforms combines model, backend, and hardware. No episode is missing or unscorable, so all bounds equal the point estimates.

\subsection{Goal completion in C0}
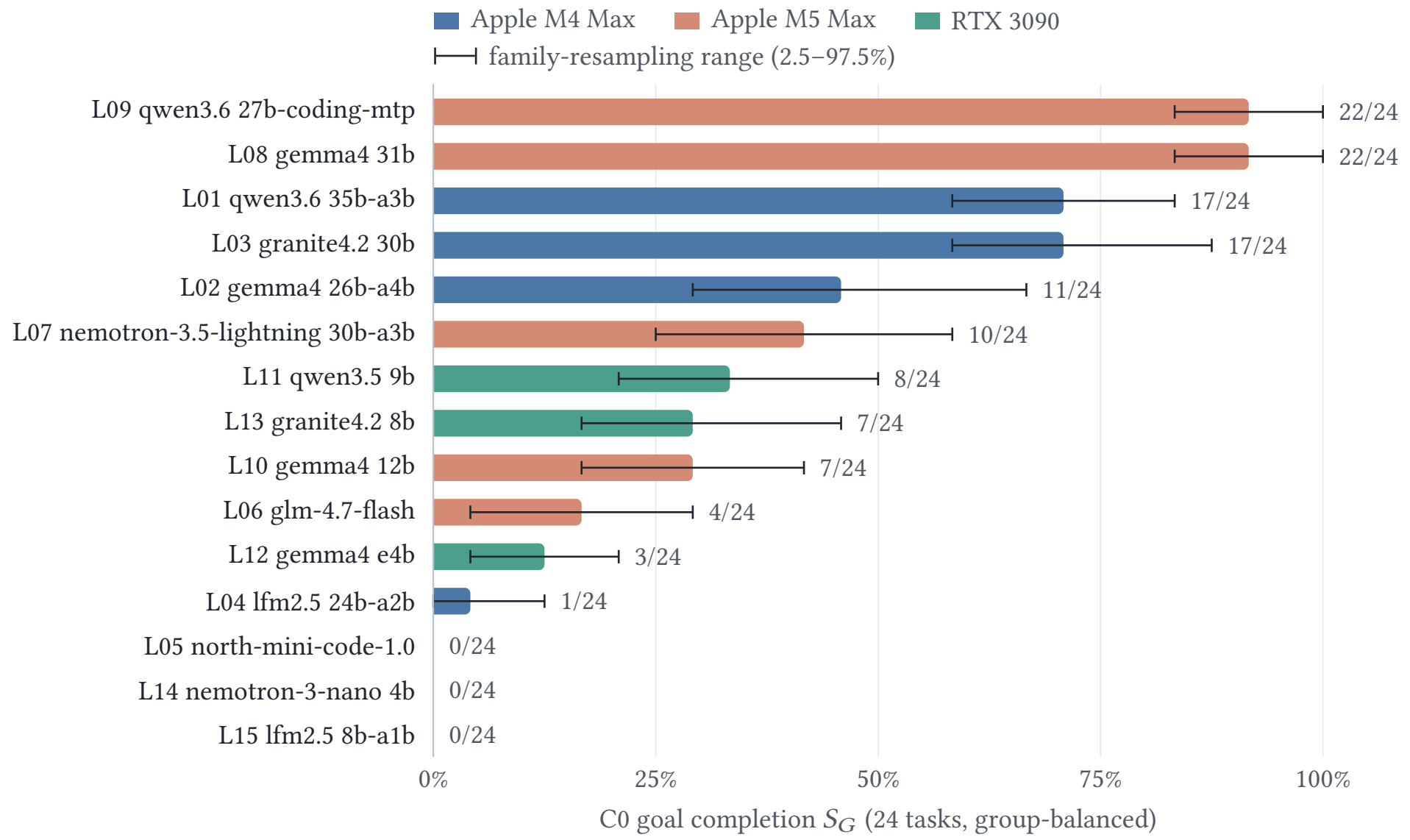
\begin{figure}[tbp]
\centering
\begin{tikzpicture}[x=1cm,y=1cm,line cap=butt]
\draw[vizgrid] (0,-6) -- (0,0.05);
\node[viztick,anchor=north] at (0,-6.08) {0\%};
\draw[vizgrid] (2,-6) -- (2,0.05);
\node[viztick,anchor=north] at (2,-6.08) {25\%};
\draw[vizgrid] (4,-6) -- (4,0.05);
\node[viztick,anchor=north] at (4,-6.08) {50\%};
\draw[vizgrid] (6,-6) -- (6,0.05);
\node[viztick,anchor=north] at (6,-6.08) {75\%};
\draw[vizgrid] (8,-6) -- (8,0.05);
\node[viztick,anchor=north] at (8,-6.08) {100\%};
\node[viztick,anchor=north] at (4,-6.42) {C0 goal completion $S_G$ (24 tasks, group-balanced)};
\fill[vizP2] (0,-0.32) [rounded corners=0.05cm] -- (7.333,-0.32) -- (7.333,-0.08) [sharp corners] -- (0,-0.08) -- cycle;
\draw[vizInk,line width=0.5pt] (6.667,-0.2) -- (8,-0.2);
\draw[vizInk,line width=0.5pt] (6.667,-0.26) -- (6.667,-0.14);
\draw[vizInk,line width=0.5pt] (8,-0.26) -- (8,-0.14);
\node[viztick,anchor=west] at (8.1,-0.2) {22/24};
\fill[vizP2] (0,-0.72) [rounded corners=0.05cm] -- (7.333,-0.72) -- (7.333,-0.48) [sharp corners] -- (0,-0.48) -- cycle;
\draw[vizInk,line width=0.5pt] (6.667,-0.6) -- (8,-0.6);
\draw[vizInk,line width=0.5pt] (6.667,-0.66) -- (6.667,-0.54);
\draw[vizInk,line width=0.5pt] (8,-0.66) -- (8,-0.54);
\node[viztick,anchor=west] at (8.1,-0.6) {22/24};
\fill[vizP1] (0,-1.12) [rounded corners=0.05cm] -- (5.667,-1.12) -- (5.667,-0.88) [sharp corners] -- (0,-0.88) -- cycle;
\draw[vizInk,line width=0.5pt] (4.667,-1) -- (6.667,-1);
\draw[vizInk,line width=0.5pt] (4.667,-1.06) -- (4.667,-0.94);
\draw[vizInk,line width=0.5pt] (6.667,-1.06) -- (6.667,-0.94);
\node[viztick,anchor=west] at (6.767,-1) {17/24};
\fill[vizP1] (0,-1.52) [rounded corners=0.05cm] -- (5.667,-1.52) -- (5.667,-1.28) [sharp corners] -- (0,-1.28) -- cycle;
\draw[vizInk,line width=0.5pt] (4.667,-1.4) -- (7,-1.4);
\draw[vizInk,line width=0.5pt] (4.667,-1.46) -- (4.667,-1.34);
\draw[vizInk,line width=0.5pt] (7,-1.46) -- (7,-1.34);
\node[viztick,anchor=west] at (7.1,-1.4) {17/24};
\fill[vizP1] (0,-1.92) [rounded corners=0.05cm] -- (3.667,-1.92) -- (3.667,-1.68) [sharp corners] -- (0,-1.68) -- cycle;
\draw[vizInk,line width=0.5pt] (2.333,-1.8) -- (5.333,-1.8);
\draw[vizInk,line width=0.5pt] (2.333,-1.86) -- (2.333,-1.74);
\draw[vizInk,line width=0.5pt] (5.333,-1.86) -- (5.333,-1.74);
\node[viztick,anchor=west] at (5.433,-1.8) {11/24};
\fill[vizP2] (0,-2.32) [rounded corners=0.05cm] -- (3.333,-2.32) -- (3.333,-2.08) [sharp corners] -- (0,-2.08) -- cycle;
\draw[vizInk,line width=0.5pt] (2,-2.2) -- (4.667,-2.2);
\draw[vizInk,line width=0.5pt] (2,-2.26) -- (2,-2.14);
\draw[vizInk,line width=0.5pt] (4.667,-2.26) -- (4.667,-2.14);
\node[viztick,anchor=west] at (4.767,-2.2) {10/24};
\fill[vizP3] (0,-2.72) [rounded corners=0.05cm] -- (2.667,-2.72) -- (2.667,-2.48) [sharp corners] -- (0,-2.48) -- cycle;
\draw[vizInk,line width=0.5pt] (1.667,-2.6) -- (4,-2.6);
\draw[vizInk,line width=0.5pt] (1.667,-2.66) -- (1.667,-2.54);
\draw[vizInk,line width=0.5pt] (4,-2.66) -- (4,-2.54);
\node[viztick,anchor=west] at (4.1,-2.6) {8/24};
\fill[vizP3] (0,-3.12) [rounded corners=0.05cm] -- (2.333,-3.12) -- (2.333,-2.88) [sharp corners] -- (0,-2.88) -- cycle;
\draw[vizInk,line width=0.5pt] (1.333,-3) -- (3.667,-3);
\draw[vizInk,line width=0.5pt] (1.333,-3.06) -- (1.333,-2.94);
\draw[vizInk,line width=0.5pt] (3.667,-3.06) -- (3.667,-2.94);
\node[viztick,anchor=west] at (3.767,-3) {7/24};
\fill[vizP2] (0,-3.52) [rounded corners=0.05cm] -- (2.333,-3.52) -- (2.333,-3.28) [sharp corners] -- (0,-3.28) -- cycle;
\draw[vizInk,line width=0.5pt] (1.333,-3.4) -- (3.333,-3.4);
\draw[vizInk,line width=0.5pt] (1.333,-3.46) -- (1.333,-3.34);
\draw[vizInk,line width=0.5pt] (3.333,-3.46) -- (3.333,-3.34);
\node[viztick,anchor=west] at (3.433,-3.4) {7/24};
\fill[vizP2] (0,-3.92) [rounded corners=0.05cm] -- (1.333,-3.92) -- (1.333,-3.68) [sharp corners] -- (0,-3.68) -- cycle;
\draw[vizInk,line width=0.5pt] (0.333,-3.8) -- (2.333,-3.8);
\draw[vizInk,line width=0.5pt] (0.333,-3.86) -- (0.333,-3.74);
\draw[vizInk,line width=0.5pt] (2.333,-3.86) -- (2.333,-3.74);
\node[viztick,anchor=west] at (2.433,-3.8) {4/24};
\fill[vizP3] (0,-4.32) [rounded corners=0.05cm] -- (1,-4.32) -- (1,-4.08) [sharp corners] -- (0,-4.08) -- cycle;
\draw[vizInk,line width=0.5pt] (0.333,-4.2) -- (1.667,-4.2);
\draw[vizInk,line width=0.5pt] (0.333,-4.26) -- (0.333,-4.14);
\draw[vizInk,line width=0.5pt] (1.667,-4.26) -- (1.667,-4.14);
\node[viztick,anchor=west] at (1.767,-4.2) {3/24};
\fill[vizP1] (0,-4.72) [rounded corners=0.05cm] -- (0.333,-4.72) -- (0.333,-4.48) [sharp corners] -- (0,-4.48) -- cycle;
\draw[vizInk,line width=0.5pt] (0,-4.6) -- (1,-4.6);
\draw[vizInk,line width=0.5pt] (0,-4.66) -- (0,-4.54);
\draw[vizInk,line width=0.5pt] (1,-4.66) -- (1,-4.54);
\node[viztick,anchor=west] at (1.1,-4.6) {1/24};
\node[viztick,anchor=west] at (0.1,-5) {0/24};
\node[viztick,anchor=west] at (0.1,-5.4) {0/24};
\node[viztick,anchor=west] at (0.1,-5.8) {0/24};
\draw[vizaxis] (0,-6) -- (0,0.05);
\node[vizlab,anchor=east] at (-0.12,-0.2) {L09 qwen3.6 27b-coding-mtp};
\node[vizlab,anchor=east] at (-0.12,-0.6) {L08 gemma4 31b};
\node[vizlab,anchor=east] at (-0.12,-1) {L01 qwen3.6 35b-a3b};
\node[vizlab,anchor=east] at (-0.12,-1.4) {L03 granite4.2 30b};
\node[vizlab,anchor=east] at (-0.12,-1.8) {L02 gemma4 26b-a4b};
\node[vizlab,anchor=east] at (-0.12,-2.2) {L07 nemotron-3.5-lightning 30b-a3b};
\node[vizlab,anchor=east] at (-0.12,-2.6) {L11 qwen3.5 9b};
\node[vizlab,anchor=east] at (-0.12,-3) {L13 granite4.2 8b};
\node[vizlab,anchor=east] at (-0.12,-3.4) {L10 gemma4 12b};
\node[vizlab,anchor=east] at (-0.12,-3.8) {L06 glm-4.7-flash};
\node[vizlab,anchor=east] at (-0.12,-4.2) {L12 gemma4 e4b};
\node[vizlab,anchor=east] at (-0.12,-4.6) {L04 lfm2.5 24b-a2b};
\node[vizlab,anchor=east] at (-0.12,-5) {L05 north-mini-code-1.0};
\node[vizlab,anchor=east] at (-0.12,-5.4) {L14 nemotron-3-nano 4b};
\node[vizlab,anchor=east] at (-0.12,-5.8) {L15 lfm2.5 8b-a1b};
\node[fill=vizP1,minimum width=0.22cm,minimum height=0.15cm,inner sep=0,anchor=west] (lgs1) at (0,0.62) {};
\node[viztick,anchor=west,xshift=0.06cm] (lgl1) at (lgs1.east) {Apple M4 Max};
\node[fill=vizP2,minimum width=0.22cm,minimum height=0.15cm,inner sep=0,anchor=west,xshift=0.3cm] (lgs2) at (lgl1.east) {};
\node[viztick,anchor=west,xshift=0.06cm] (lgl2) at (lgs2.east) {Apple M5 Max};
\node[fill=vizP3,minimum width=0.22cm,minimum height=0.15cm,inner sep=0,anchor=west,xshift=0.3cm] (lgs3) at (lgl2.east) {};
\node[viztick,anchor=west,xshift=0.06cm] (lgl3) at (lgs3.east) {RTX 3090};
\node[inner sep=0,anchor=west] (lgs4) at (0,0.28) {\textcolor{vizInk}{\rule[0.3ex]{0.5pt}{1ex}\rule[0.75ex]{0.35cm}{0.5pt}\rule[0.3ex]{0.5pt}{1ex}}};
\node[viztick,anchor=west,xshift=0.06cm] (lgl4) at (lgs4.east) {family-resampling range (2.5--97.5\%)};
\end{tikzpicture}
\caption{C0 goal completion $S_G$ by configuration, sorted. The bar is the share of the 24 tasks whose goal was reached in C0, and the label gives the count. The whisker is the 2.5--97.5\% range over 10,000 family resamplings within scenario groups, a measure of dependence on the family mix rather than a confidence interval. Color marks the platform.}\label{fig:leaderboard}
\end{figure}

Figure~\ref{fig:leaderboard} and Table~\ref{tab:leaderboard} give the primary score. Two configurations, L08 and L09, reached the goal in 22 of the 24 tasks ($S_G=0.92$), and L01 and L03 in 17 ($S_G=0.71$). L02 and L07 followed with 11 and 10 tasks. Six configurations reached between one and eight tasks, and L05, L14, and L15 reached none. The family-resampling ranges of neighboring configurations overlap. For example, the range of L03 (0.58 to 0.88) reaches into that of L08 and L09 (0.83 to 1.00). Leaving out any one family changes no configuration's $S_G$ by more than 0.042, consistent with the weight of one task; this sensitivity check does not establish that neighboring ranks are distinct. The RTX 3090 arm ran small models that fit on 24 GB GPUs, and none of its configurations reached more than eight tasks. Each platform ran different models, so this pattern describes the evaluated configurations rather than an isolated hardware effect.

\begin{table}[tbp]
\centering
{\footnotesize\setlength{\tabcolsep}{3.5pt}
\begin{tabular}{@{}lrrrrrrl@{}}\toprule
Configuration & C0 & C1 & $\Delta_1$ & C0 (12) & C2 & $\Delta_2$ & \multicolumn{1}{c}{C0 \textcolor{vizInk2}{(top)} and C1}\\
 & \textcolor{vizInk2}{of 24} & \textcolor{vizInk2}{of 24} & & \textcolor{vizInk2}{of 12} & \textcolor{vizInk2}{of 24} & & \\\midrule
L09 qwen3.6 27b-coding-mtp & 22 & 18 & $-$0.167 & 11 & 11 & $-$0.458 & \tikz[baseline=-0.6ex]{\fill[vizMid] (0,-0.13) rectangle (2.3,0.13);\fill[vizP2] (0,0.015) rectangle (2.108,0.12);\fill[vizP2!40!white] (0,-0.12) rectangle (1.725,-0.015);}\\
L08 gemma4 31b & 22 & 20 & $-$0.083 & 11 & 12 & $-$0.417 & \tikz[baseline=-0.6ex]{\fill[vizMid] (0,-0.13) rectangle (2.3,0.13);\fill[vizP2] (0,0.015) rectangle (2.108,0.12);\fill[vizP2!40!white] (0,-0.12) rectangle (1.917,-0.015);}\\
L01 qwen3.6 35b-a3b & 17 & 14 & $-$0.125 & 7 & 12 & $-$0.083 & \tikz[baseline=-0.6ex]{\fill[vizMid] (0,-0.13) rectangle (2.3,0.13);\fill[vizP1] (0,0.015) rectangle (1.629,0.12);\fill[vizP1!40!white] (0,-0.12) rectangle (1.342,-0.015);}\\
L03 granite4.2 30b & 17 & 12 & $-$0.208 & 8 & 10 & $-$0.250 & \tikz[baseline=-0.6ex]{\fill[vizMid] (0,-0.13) rectangle (2.3,0.13);\fill[vizP1] (0,0.015) rectangle (1.629,0.12);\fill[vizP1!40!white] (0,-0.12) rectangle (1.15,-0.015);}\\
L02 gemma4 26b-a4b & 11 & 7 & $-$0.167 & 7 & 10 & $-$0.167 & \tikz[baseline=-0.6ex]{\fill[vizMid] (0,-0.13) rectangle (2.3,0.13);\fill[vizP1] (0,0.015) rectangle (1.054,0.12);\fill[vizP1!40!white] (0,-0.12) rectangle (0.671,-0.015);}\\
L07 nemotron-3.5-lightning 30b-a3b & 10 & 6 & $-$0.167 & 7 & 8 & $-$0.250 & \tikz[baseline=-0.6ex]{\fill[vizMid] (0,-0.13) rectangle (2.3,0.13);\fill[vizP2] (0,0.015) rectangle (0.958,0.12);\fill[vizP2!40!white] (0,-0.12) rectangle (0.575,-0.015);}\\
L11 qwen3.5 9b & 8 & 4 & $-$0.167 & 5 & 3 & $-$0.292 & \tikz[baseline=-0.6ex]{\fill[vizMid] (0,-0.13) rectangle (2.3,0.13);\fill[vizP3] (0,0.015) rectangle (0.767,0.12);\fill[vizP3!40!white] (0,-0.12) rectangle (0.383,-0.015);}\\
L13 granite4.2 8b & 7 & 4 & $-$0.125 & 5 & 5 & $-$0.208 & \tikz[baseline=-0.6ex]{\fill[vizMid] (0,-0.13) rectangle (2.3,0.13);\fill[vizP3] (0,0.015) rectangle (0.671,0.12);\fill[vizP3!40!white] (0,-0.12) rectangle (0.383,-0.015);}\\
L10 gemma4 12b & 7 & 6 & $-$0.042 & 4 & 5 & $-$0.125 & \tikz[baseline=-0.6ex]{\fill[vizMid] (0,-0.13) rectangle (2.3,0.13);\fill[vizP2] (0,0.015) rectangle (0.671,0.12);\fill[vizP2!40!white] (0,-0.12) rectangle (0.575,-0.015);}\\
L06 glm-4.7-flash & 4 & 3 & $-$0.042 & 2 & 1 & $-$0.125 & \tikz[baseline=-0.6ex]{\fill[vizMid] (0,-0.13) rectangle (2.3,0.13);\fill[vizP2] (0,0.015) rectangle (0.383,0.12);\fill[vizP2!40!white] (0,-0.12) rectangle (0.287,-0.015);}\\
L12 gemma4 e4b & 3 & 0 & $-$0.125 & 2 & 2 & $-$0.083 & \tikz[baseline=-0.6ex]{\fill[vizMid] (0,-0.13) rectangle (2.3,0.13);\fill[vizP3] (0,0.015) rectangle (0.287,0.12);}\\
L04 lfm2.5 24b-a2b & 1 & 2 & +0.042 & 1 & 2 & 0.000 & \tikz[baseline=-0.6ex]{\fill[vizMid] (0,-0.13) rectangle (2.3,0.13);\fill[vizP1] (0,0.015) rectangle (0.096,0.12);\fill[vizP1!40!white] (0,-0.12) rectangle (0.192,-0.015);}\\
L05 north-mini-code-1.0 & 0 & 1 & +0.042 & 0 & 0 & 0.000 & \tikz[baseline=-0.6ex]{\fill[vizMid] (0,-0.13) rectangle (2.3,0.13);\fill[vizP1!40!white] (0,-0.12) rectangle (0.096,-0.015);}\\
L14 nemotron-3-nano 4b & 0 & 0 & 0.000 & 0 & 0 & 0.000 & \tikz[baseline=-0.6ex]{\fill[vizMid] (0,-0.13) rectangle (2.3,0.13);}\\
L15 lfm2.5 8b-a1b & 0 & 0 & 0.000 & 0 & 0 & 0.000 & \tikz[baseline=-0.6ex]{\fill[vizMid] (0,-0.13) rectangle (2.3,0.13);}\\
\bottomrule\end{tabular}}
\caption{Goal completion with exact counts, sorted as in Figure~\ref{fig:leaderboard}. C0 and C1: tasks with the goal reached, of 24. C0 (12): C0 on the twelve state-changing tasks, the baseline of $\Delta_2$. C2: goal reached in the 24 C2 episodes (twelve tasks, two hidden worlds). $\Delta_1$ and $\Delta_2$ follow Equation~\eqref{eq:contrast}. Bars: C0 (upper, platform color) and C1 (lower, lighter).}\label{tab:leaderboard}
\end{table}

\begin{figure}[tbp]
\centering
\input{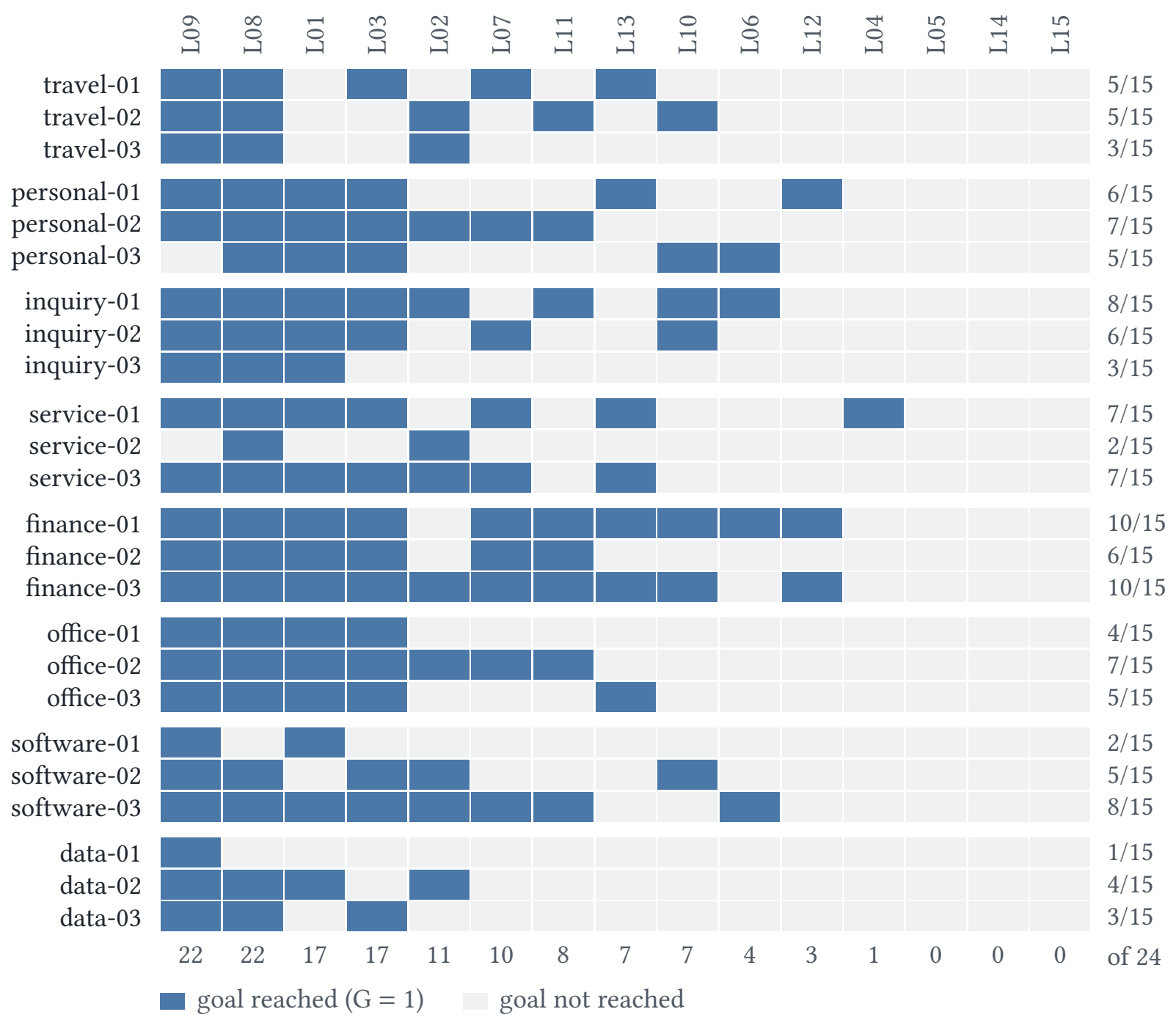}
\caption{Goal reached in C0 for every task and configuration. Columns are sorted as in Figure~\ref{fig:leaderboard}, and rows are grouped by scenario group. Right margin: configurations reaching the goal of the task. Bottom margin: tasks reached by the configuration.}\label{fig:heat-c0}
\end{figure}

Figure~\ref{fig:heat-c0} shows the outcome of every task. Every task was completed by at least one configuration. \code{data-test-01} was completed only by L09, and \code{service-test-02} and \code{software-test-01} by two configurations each. \code{finance-test-01} and \code{finance-test-03} were completed most often, by ten configurations each. Pooled over configurations, the finance group had the most completions (26 of 45 configuration--task pairs) and the data group the fewest (8 of 45). The other groups ranged from 13 (travel) to 18 (personal). The group profile of each configuration is in Figure~\ref{fig:group-profile}.

Few reached goals failed on process or evidence (Table~\ref{tab:endpoints}, Figure~\ref{fig:partition}). In C0, $S_J$ equals $S_G$ for ten configurations and is lower by one or two tasks for the other five. Over all conditions, 930 records have G, P, and E defined, and 276 of them reached the goal. Of these 276, 265 were joint successes, 3 violated a process constraint, and 8 had insufficient evidence. The other 654 records did not reach the goal. P on its own is uninformative for configurations that acted little. L14 has $S_P=1.00$ with $S_G=0$, because the process constraints forbid certain actions and doing nothing satisfies them.

\subsection{Corrections}
\begin{figure}[tbp]
\centering
\input{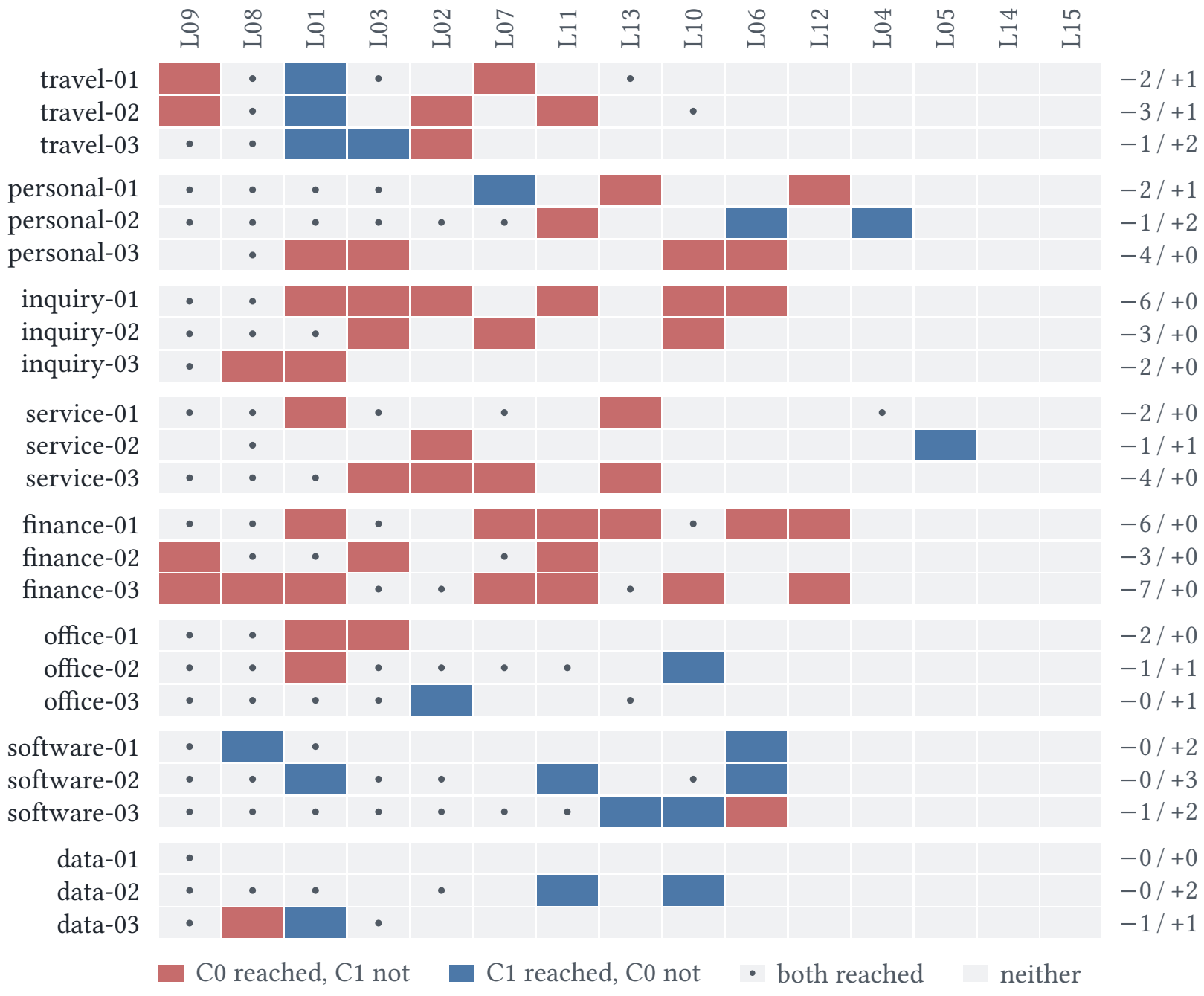}
\caption{Change from C0 to C1 for every task and configuration. Red: the goal was reached in C0 but not after the correction. Blue: the reverse. A dot marks a task reached in both conditions. Columns are sorted as in Figure~\ref{fig:leaderboard}. Right margin: worsened and improved configurations for the task.}\label{fig:heat-c1}
\end{figure}

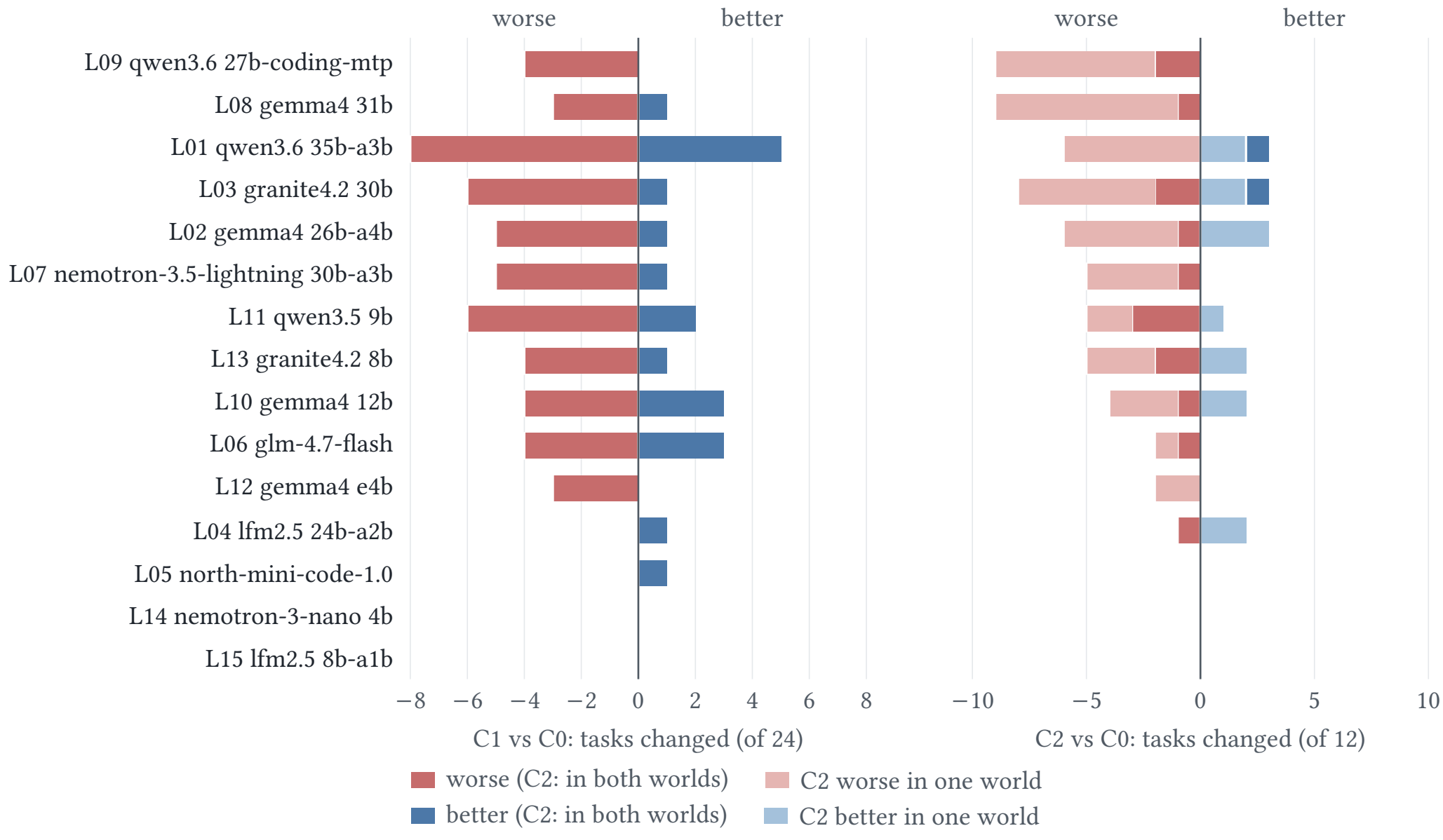
\begin{figure}[tbp]
\centering
\begin{tikzpicture}[x=1cm,y=1cm,line cap=butt]
\draw[vizgrid] (0,-6) -- (0,0.05);
\node[viztick,anchor=north] at (0,-6.08) {$-$8};
\draw[vizgrid] (0.538,-6) -- (0.538,0.05);
\node[viztick,anchor=north] at (0.538,-6.08) {$-$6};
\draw[vizgrid] (1.075,-6) -- (1.075,0.05);
\node[viztick,anchor=north] at (1.075,-6.08) {$-$4};
\draw[vizgrid] (1.612,-6) -- (1.612,0.05);
\node[viztick,anchor=north] at (1.612,-6.08) {$-$2};
\draw[vizgrid] (2.15,-6) -- (2.15,0.05);
\node[viztick,anchor=north] at (2.15,-6.08) {0};
\draw[vizgrid] (2.688,-6) -- (2.688,0.05);
\node[viztick,anchor=north] at (2.688,-6.08) {2};
\draw[vizgrid] (3.225,-6) -- (3.225,0.05);
\node[viztick,anchor=north] at (3.225,-6.08) {4};
\draw[vizgrid] (3.762,-6) -- (3.762,0.05);
\node[viztick,anchor=north] at (3.762,-6.08) {6};
\draw[vizgrid] (4.3,-6) -- (4.3,0.05);
\node[viztick,anchor=north] at (4.3,-6.08) {8};
\node[viztick,anchor=north] at (2.15,-6.42) {C1 vs C0: tasks changed (of 24)};
\node[viztick,anchor=south] at (1.075,0.12) {worse};
\node[viztick,anchor=south] at (3.225,0.12) {better};
\fill[vizRed] (1.075,-0.32) rectangle (2.15,-0.08);
\draw[vizgap] (1.075,-0.33) -- (1.075,-0.07);
\fill[vizRed] (1.344,-0.72) rectangle (2.15,-0.48);
\draw[vizgap] (1.344,-0.73) -- (1.344,-0.47);
\fill[vizBlue45] (2.15,-0.72) rectangle (2.419,-0.48);
\draw[vizgap] (2.15,-0.73) -- (2.15,-0.47);
\fill[vizRed] (0,-1.12) rectangle (2.15,-0.88);
\draw[vizgap] (0,-1.13) -- (0,-0.87);
\fill[vizBlue45] (2.15,-1.12) rectangle (3.494,-0.88);
\draw[vizgap] (2.15,-1.13) -- (2.15,-0.87);
\fill[vizRed] (0.538,-1.52) rectangle (2.15,-1.28);
\draw[vizgap] (0.538,-1.53) -- (0.538,-1.27);
\fill[vizBlue45] (2.15,-1.52) rectangle (2.419,-1.28);
\draw[vizgap] (2.15,-1.53) -- (2.15,-1.27);
\fill[vizRed] (0.806,-1.92) rectangle (2.15,-1.68);
\draw[vizgap] (0.806,-1.93) -- (0.806,-1.67);
\fill[vizBlue45] (2.15,-1.92) rectangle (2.419,-1.68);
\draw[vizgap] (2.15,-1.93) -- (2.15,-1.67);
\fill[vizRed] (0.806,-2.32) rectangle (2.15,-2.08);
\draw[vizgap] (0.806,-2.33) -- (0.806,-2.07);
\fill[vizBlue45] (2.15,-2.32) rectangle (2.419,-2.08);
\draw[vizgap] (2.15,-2.33) -- (2.15,-2.07);
\fill[vizRed] (0.538,-2.72) rectangle (2.15,-2.48);
\draw[vizgap] (0.538,-2.73) -- (0.538,-2.47);
\fill[vizBlue45] (2.15,-2.72) rectangle (2.688,-2.48);
\draw[vizgap] (2.15,-2.73) -- (2.15,-2.47);
\fill[vizRed] (1.075,-3.12) rectangle (2.15,-2.88);
\draw[vizgap] (1.075,-3.13) -- (1.075,-2.87);
\fill[vizBlue45] (2.15,-3.12) rectangle (2.419,-2.88);
\draw[vizgap] (2.15,-3.13) -- (2.15,-2.87);
\fill[vizRed] (1.075,-3.52) rectangle (2.15,-3.28);
\draw[vizgap] (1.075,-3.53) -- (1.075,-3.27);
\fill[vizBlue45] (2.15,-3.52) rectangle (2.956,-3.28);
\draw[vizgap] (2.15,-3.53) -- (2.15,-3.27);
\fill[vizRed] (1.075,-3.92) rectangle (2.15,-3.68);
\draw[vizgap] (1.075,-3.93) -- (1.075,-3.67);
\fill[vizBlue45] (2.15,-3.92) rectangle (2.956,-3.68);
\draw[vizgap] (2.15,-3.93) -- (2.15,-3.67);
\fill[vizRed] (1.344,-4.32) rectangle (2.15,-4.08);
\draw[vizgap] (1.344,-4.33) -- (1.344,-4.07);
\fill[vizBlue45] (2.15,-4.72) rectangle (2.419,-4.48);
\draw[vizgap] (2.15,-4.73) -- (2.15,-4.47);
\fill[vizBlue45] (2.15,-5.12) rectangle (2.419,-4.88);
\draw[vizgap] (2.15,-5.13) -- (2.15,-4.87);
\draw[vizInk2,line width=0.5pt] (2.15,-6) -- (2.15,0.05);
\draw[vizgrid] (5.3,-6) -- (5.3,0.05);
\node[viztick,anchor=north] at (5.3,-6.08) {$-$10};
\draw[vizgrid] (6.375,-6) -- (6.375,0.05);
\node[viztick,anchor=north] at (6.375,-6.08) {$-$5};
\draw[vizgrid] (7.45,-6) -- (7.45,0.05);
\node[viztick,anchor=north] at (7.45,-6.08) {0};
\draw[vizgrid] (8.525,-6) -- (8.525,0.05);
\node[viztick,anchor=north] at (8.525,-6.08) {5};
\draw[vizgrid] (9.6,-6) -- (9.6,0.05);
\node[viztick,anchor=north] at (9.6,-6.08) {10};
\node[viztick,anchor=north] at (7.45,-6.42) {C2 vs C0: tasks changed (of 12)};
\node[viztick,anchor=south] at (6.375,0.12) {worse};
\node[viztick,anchor=south] at (8.525,0.12) {better};
\fill[vizRed] (7.02,-0.32) rectangle (7.45,-0.08);
\draw[vizgap] (7.02,-0.33) -- (7.02,-0.07);
\fill[vizRedLight] (5.515,-0.32) rectangle (7.02,-0.08);
\draw[vizgap] (5.515,-0.33) -- (5.515,-0.07);
\fill[vizRed] (7.235,-0.72) rectangle (7.45,-0.48);
\draw[vizgap] (7.235,-0.73) -- (7.235,-0.47);
\fill[vizRedLight] (5.515,-0.72) rectangle (7.235,-0.48);
\draw[vizgap] (5.515,-0.73) -- (5.515,-0.47);
\fill[vizRedLight] (6.16,-1.12) rectangle (7.45,-0.88);
\draw[vizgap] (6.16,-1.13) -- (6.16,-0.87);
\fill[vizBlue25] (7.45,-1.12) rectangle (7.88,-0.88);
\draw[vizgap] (7.45,-1.13) -- (7.45,-0.87);
\fill[vizBlue45] (7.88,-1.12) rectangle (8.095,-0.88);
\draw[vizgap] (7.88,-1.13) -- (7.88,-0.87);
\fill[vizRed] (7.02,-1.52) rectangle (7.45,-1.28);
\draw[vizgap] (7.02,-1.53) -- (7.02,-1.27);
\fill[vizRedLight] (5.73,-1.52) rectangle (7.02,-1.28);
\draw[vizgap] (5.73,-1.53) -- (5.73,-1.27);
\fill[vizBlue25] (7.45,-1.52) rectangle (7.88,-1.28);
\draw[vizgap] (7.45,-1.53) -- (7.45,-1.27);
\fill[vizBlue45] (7.88,-1.52) rectangle (8.095,-1.28);
\draw[vizgap] (7.88,-1.53) -- (7.88,-1.27);
\fill[vizRed] (7.235,-1.92) rectangle (7.45,-1.68);
\draw[vizgap] (7.235,-1.93) -- (7.235,-1.67);
\fill[vizRedLight] (6.16,-1.92) rectangle (7.235,-1.68);
\draw[vizgap] (6.16,-1.93) -- (6.16,-1.67);
\fill[vizBlue25] (7.45,-1.92) rectangle (8.095,-1.68);
\draw[vizgap] (7.45,-1.93) -- (7.45,-1.67);
\fill[vizRed] (7.235,-2.32) rectangle (7.45,-2.08);
\draw[vizgap] (7.235,-2.33) -- (7.235,-2.07);
\fill[vizRedLight] (6.375,-2.32) rectangle (7.235,-2.08);
\draw[vizgap] (6.375,-2.33) -- (6.375,-2.07);
\fill[vizRed] (6.805,-2.72) rectangle (7.45,-2.48);
\draw[vizgap] (6.805,-2.73) -- (6.805,-2.47);
\fill[vizRedLight] (6.375,-2.72) rectangle (6.805,-2.48);
\draw[vizgap] (6.375,-2.73) -- (6.375,-2.47);
\fill[vizBlue25] (7.45,-2.72) rectangle (7.665,-2.48);
\draw[vizgap] (7.45,-2.73) -- (7.45,-2.47);
\fill[vizRed] (7.02,-3.12) rectangle (7.45,-2.88);
\draw[vizgap] (7.02,-3.13) -- (7.02,-2.87);
\fill[vizRedLight] (6.375,-3.12) rectangle (7.02,-2.88);
\draw[vizgap] (6.375,-3.13) -- (6.375,-2.87);
\fill[vizBlue25] (7.45,-3.12) rectangle (7.88,-2.88);
\draw[vizgap] (7.45,-3.13) -- (7.45,-2.87);
\fill[vizRed] (7.235,-3.52) rectangle (7.45,-3.28);
\draw[vizgap] (7.235,-3.53) -- (7.235,-3.27);
\fill[vizRedLight] (6.59,-3.52) rectangle (7.235,-3.28);
\draw[vizgap] (6.59,-3.53) -- (6.59,-3.27);
\fill[vizBlue25] (7.45,-3.52) rectangle (7.88,-3.28);
\draw[vizgap] (7.45,-3.53) -- (7.45,-3.27);
\fill[vizRed] (7.235,-3.92) rectangle (7.45,-3.68);
\draw[vizgap] (7.235,-3.93) -- (7.235,-3.67);
\fill[vizRedLight] (7.02,-3.92) rectangle (7.235,-3.68);
\draw[vizgap] (7.02,-3.93) -- (7.02,-3.67);
\fill[vizRedLight] (7.02,-4.32) rectangle (7.45,-4.08);
\draw[vizgap] (7.02,-4.33) -- (7.02,-4.07);
\fill[vizRed] (7.235,-4.72) rectangle (7.45,-4.48);
\draw[vizgap] (7.235,-4.73) -- (7.235,-4.47);
\fill[vizBlue25] (7.45,-4.72) rectangle (7.88,-4.48);
\draw[vizgap] (7.45,-4.73) -- (7.45,-4.47);
\draw[vizInk2,line width=0.5pt] (7.45,-6) -- (7.45,0.05);
\node[vizlab,anchor=east] at (-0.12,-0.2) {L09 qwen3.6 27b-coding-mtp};
\node[vizlab,anchor=east] at (-0.12,-0.6) {L08 gemma4 31b};
\node[vizlab,anchor=east] at (-0.12,-1) {L01 qwen3.6 35b-a3b};
\node[vizlab,anchor=east] at (-0.12,-1.4) {L03 granite4.2 30b};
\node[vizlab,anchor=east] at (-0.12,-1.8) {L02 gemma4 26b-a4b};
\node[vizlab,anchor=east] at (-0.12,-2.2) {L07 nemotron-3.5-lightning 30b-a3b};
\node[vizlab,anchor=east] at (-0.12,-2.6) {L11 qwen3.5 9b};
\node[vizlab,anchor=east] at (-0.12,-3) {L13 granite4.2 8b};
\node[vizlab,anchor=east] at (-0.12,-3.4) {L10 gemma4 12b};
\node[vizlab,anchor=east] at (-0.12,-3.8) {L06 glm-4.7-flash};
\node[vizlab,anchor=east] at (-0.12,-4.2) {L12 gemma4 e4b};
\node[vizlab,anchor=east] at (-0.12,-4.6) {L04 lfm2.5 24b-a2b};
\node[vizlab,anchor=east] at (-0.12,-5) {L05 north-mini-code-1.0};
\node[vizlab,anchor=east] at (-0.12,-5.4) {L14 nemotron-3-nano 4b};
\node[vizlab,anchor=east] at (-0.12,-5.8) {L15 lfm2.5 8b-a1b};
\node[fill=vizRed,minimum width=0.22cm,minimum height=0.15cm,inner sep=0,anchor=west] (lgs1) at (0,-6.95) {};
\node[viztick,anchor=west,xshift=0.06cm] (lgl1) at (lgs1.east) {worse (C2: in both worlds)};
\node[fill=vizRedLight,minimum width=0.22cm,minimum height=0.15cm,inner sep=0,anchor=west,xshift=0.3cm] (lgs2) at (lgl1.east) {};
\node[viztick,anchor=west,xshift=0.06cm] (lgl2) at (lgs2.east) {C2 worse in one world};
\node[fill=vizBlue45,minimum width=0.22cm,minimum height=0.15cm,inner sep=0,anchor=west] (lgs3) at (0,-7.29) {};
\node[viztick,anchor=west,xshift=0.06cm] (lgl3) at (lgs3.east) {better (C2: in both worlds)};
\node[fill=vizBlue25,minimum width=0.22cm,minimum height=0.15cm,inner sep=0,anchor=west,xshift=0.3cm] (lgs4) at (lgl3.east) {};
\node[viztick,anchor=west,xshift=0.06cm] (lgl4) at (lgs4.east) {C2 better in one world};
\end{tikzpicture}
\caption{Tasks whose goal outcome changed between C0 and each condition, per configuration: worsened to the left, improved to the right. For C2, the lighter shades mark a change in only one of the two hidden worlds.}\label{fig:transitions}
\end{figure}

Figures~\ref{fig:heat-c1} and~\ref{fig:transitions} compare C1 with C0 task by task. $\Delta_1$ was negative for eleven configurations, positive for L04 and L05 (one task each), and zero for L14 and L15, which reached no goal in either condition. Pooled over the 360 configuration--task pairs, 52 changed from success to failure, 20 from failure to success, and 288 were unchanged, and the mean $\Delta_1$ was $-0.089$. In other words, 52 of the 129 C0 successes were not reached again in C1, and 20 of the 231 C0 failures were reached in C1.

Each pair has one run per condition, so an observed transition includes both condition response and run-to-run variation. The imbalance of 52 worsened versus 20 improved pairs documents lower completion under C1 in this collected matrix. It is a descriptive condition contrast: the design does not estimate repeated-run variability or isolate a causal effect for an individual pair. Pairs also share tasks across configurations, so the pooled counts are not treated as independent trials or used for a significance test (Section~\ref{sec:analysis}).

The changes were concentrated in a few tasks. In \code{finance-test-03}, \code{finance-test-01}, and \code{inquiry-test-01}, seven, six, and six configurations went from success to failure, and none went the other way. The three software tasks moved in the opposite direction, with seven changes to success and one to failure. The correction was delivered in 329 of the 360 C1 episodes. Most of the 31 undelivered corrections were in L14 (16) and L15 (7). These episodes remain in the C1 denominator.

\subsection{Ambiguous operation replies}
On the twelve state-changing tasks, the configurations reached the goal in 70 of 180 C0 episodes and in 81 of 360 C2 episodes. $\Delta_2$ was negative for eleven configurations and zero for L04, L05, L14, and L15, which reached at most one of these goals in C0. The mean $\Delta_2$ was $-0.164$. The two strongest C0 configurations lost the most. L09 went from 11 of 12 tasks in C0 to 11 of 24 C2 episodes ($\Delta_2=-0.458$), and L08 from 11 of 12 to 12 of 24 ($\Delta_2=-0.417$). Over the 180 configuration--task cells, the per-task difference $d$ was $-1$ in 15 cells, $-\tfrac12$ in 47, $0$ in 102, $\tfrac12$ in 14, and $1$ in 2 (Figure~\ref{fig:heat-c2}).

The trigger diagnostics are in Table~\ref{tab:diagnostics} and Figure~\ref{fig:diagnostics}. The ambiguous reply occurred in \CtwoTriggeredEpisodes{} of the \CtwoAssignedEpisodes{} C2 episodes. The original tool-name query proxy was positive in \CtwoToolQueryEpisodes{} of these episodes: a successful call to a designated verification-tool name appeared after the reply and before another successful call of the designated operation's name. This proxy does not establish which object was queried or what state information was returned, and therefore is not an operation-level verification rate. The operation-matched analysis below uses event arguments and frozen tool contracts to distinguish target-covering queries from related retries. In the committed world, 17 of the 157 episodes with the reply violated a process constraint, which bounds the rate of duplicated effects from above.

An exploratory split by hidden world, outside the analysis plan, localizes the completion difference. In the world where the operation took effect, the configurations reached the goal in 67 of 180 episodes, close to their 70 of 180 in C0. In the world where it did not take effect, they reached it in 14 of 180, and eight of the twelve tasks had no success in that world. Reaching the goal there requires calling the operation again, which would duplicate it in the other world unless the agent first establishes its state. The completion split alone does not distinguish object-specific verification, blind retry, and termination without either action. In particular, the tool-name proxy cannot identify the prevalence of a strategy that accepts the ambiguous reply without checking the affected operation. \code{travel-test-03} is the only task with more C2 than C0 successes (11 of 30 C2 episodes against 3 of 15 C0 episodes), and 6 of its 11 C2 successes came in the world where the operation had not taken effect. The second scorer agreed on all of its goal outcomes.

\subsection{Operation-matched follow-up (exploratory)}\label{sec:c2-recovery}
A post-hoc reconstruction follows the object or overlapping object scope of the operation that returned the ambiguous reply, rather than stopping at the next call with the same tool name. It classifies observed tool-event order, using successful calls and the frozen handlers' query scopes (Appendix~\ref{app:c2-operation-verification}). Table~\ref{tab:c2-operation-verification} partitions all \CtwoAssignedEpisodes{} assigned C2 episodes. Among the \CtwoTriggeredEpisodes{} triggered episodes, \CtwoVerificationFirstEpisodes{} had a successful target-covering state query before any successful related retry, \CtwoRetryFirstEpisodes{} had a successful related retry first, and \CtwoNoFollowUpEpisodes{} had neither qualifying follow-up. There were \CtwoIndeterminateEpisodes{} indeterminate classifications and \CtwoNotTriggeredEpisodes{} untriggered episodes. A target-covering query is an invocation whose frozen contract exposes the relevant operation state. The event order does not establish that the agent understood the response, received it before deciding a later action, or verified every payload field. A related retry can change payload values and is not synonymous with a duplicated committed effect.

\begin{table}[tbp]
\centering\small
\begin{tabular}{@{}lrrr@{}}\toprule
First observed recovery behavior & Committed & Not committed & Total\\\midrule
State query before any related retry & 33 & 30 & 63\\
Related retry before any state query & 5 & 7 & 12\\
Neither state query nor related retry & 119 & 116 & 235\\
Ordering or target indeterminate & 0 & 0 & 0\\
Ambiguous reply did not occur & 23 & 27 & 50\\
\midrule All assigned C2 episodes & 180 & 180 & 360\\
\bottomrule\end{tabular}

\caption{Post-hoc operation-matched follow-up on the frozen C2 event records. Counts are episodes, partitioned by first qualifying successful query or related retry, with untriggered and indeterminate episodes retained. A query must cover the original target under the frozen tool contract; a related retry acts on the same or overlapping object scope. Rejected calls do not count as successful follow-up. The categories describe observed tool-event order, not response comprehension, safe recovery, or changes to G/P/E/J.}\label{tab:c2-operation-verification}
\end{table}

The new classification and the original proxy are not interchangeable. Of the original 60 proxy-positive episodes, 48 are also state-query-first under operation matching and 12 are not. Operation matching identifies 15 additional state-query-first episodes that the tool-name proxy missed. These differences arise because a query may concern another object and because a disjoint operation need not end the original object's follow-up window.

The neither-follow-up category is not an abandonment label: 214 of its \CtwoNoFollowUpEpisodes{} episodes ended with \code{submitted}, 20 with \code{text\_reply\_no\_tool\_call}, and one with \code{stuck\_detected}. It may include rejected query or retry attempts, which remain in the event log. Within this category, goal completion was 52/119 in the committed world and 0/116 in the uncommitted world. This descriptive split is consistent with the task requirement to apply an operation that has not taken effect. It does not estimate the causal benefit of verification or by itself show what an agent believed about the ambiguous reply.

\begin{figure}[tbp]
\centering
{\small\textbf{(a) Observed follow-up across all assigned episodes}\par}
\begin{tikzpicture}[x=1cm,y=1cm,line cap=butt]
\draw[vizgrid] (0,-1.49) -- (0,0.24);
\node[viztick,anchor=north] at (0,-1.53) {0\%};
\draw[vizgrid] (2.25,-1.49) -- (2.25,0.24);
\node[viztick,anchor=north] at (2.25,-1.53) {25\%};
\draw[vizgrid] (4.5,-1.49) -- (4.5,0.24);
\node[viztick,anchor=north] at (4.5,-1.53) {50\%};
\draw[vizgrid] (6.75,-1.49) -- (6.75,0.24);
\node[viztick,anchor=north] at (6.75,-1.53) {75\%};
\draw[vizgrid] (9,-1.49) -- (9,0.24);
\node[viztick,anchor=north] at (9,-1.53) {100\%};
\node[viztick,anchor=north] at (4.5,-1.96) {Share of all assigned C2 episodes in each world};
\node[vizlab,anchor=east] at (-0.16,0) {Effect present};
\fill[vizQuery] (0,-0.23) rectangle (1.65,0.23);
\draw[vizgap] (0,-0.24) -- (0,0.24);
\node[vizlab,anchor=center,text=white] at (0.825,0) {33};
\fill[vizRetry] (1.65,-0.23) rectangle (1.9,0.23);
\draw[vizgap] (1.65,-0.24) -- (1.65,0.24);
\draw[vizInk2,line width=0.35pt] (1.775,0.23) -- (1.775,0.45);
\node[vizlab,anchor=south] at (1.775,0.48) {5};
\fill[vizNoFollow] (1.9,-0.23) rectangle (7.85,0.23);
\draw[vizgap] (1.9,-0.24) -- (1.9,0.24);
\node[vizlab,anchor=center,text=white] at (4.875,0) {119};
\fill[vizUntriggered] (7.85,-0.23) rectangle (9,0.23);
\draw[vizgap] (7.85,-0.24) -- (7.85,0.24);
\node[vizlab,anchor=center,text=vizInk] at (8.425,0) {23};
\node[viztick,anchor=west] at (9.14,0) {n=180};
\node[vizlab,anchor=east] at (-0.16,-1.15) {No effect};
\fill[vizQuery] (0,-1.38) rectangle (1.5,-0.92);
\draw[vizgap] (0,-1.39) -- (0,-0.91);
\node[vizlab,anchor=center,text=white] at (0.75,-1.15) {30};
\fill[vizRetry] (1.5,-1.38) rectangle (1.85,-0.92);
\draw[vizgap] (1.5,-1.39) -- (1.5,-0.91);
\draw[vizInk2,line width=0.35pt] (1.675,-0.92) -- (1.675,-0.7);
\node[vizlab,anchor=south] at (1.675,-0.67) {7};
\fill[vizNoFollow] (1.85,-1.38) rectangle (7.65,-0.92);
\draw[vizgap] (1.85,-1.39) -- (1.85,-0.91);
\node[vizlab,anchor=center,text=white] at (4.75,-1.15) {116};
\fill[vizUntriggered] (7.65,-1.38) rectangle (9,-0.92);
\draw[vizgap] (7.65,-1.39) -- (7.65,-0.91);
\node[vizlab,anchor=center,text=vizInk] at (8.325,-1.15) {27};
\node[viztick,anchor=west] at (9.14,-1.15) {n=180};
\node[fill=vizQuery,minimum width=0.22cm,minimum height=0.15cm,inner sep=0,anchor=west] (lgs1) at (0,1.3) {};
\node[viztick,anchor=west,xshift=0.06cm] (lgl1) at (lgs1.east) {State query first};
\node[fill=vizRetry,minimum width=0.22cm,minimum height=0.15cm,inner sep=0,anchor=west,xshift=0.3cm] (lgs2) at (lgl1.east) {};
\node[viztick,anchor=west,xshift=0.06cm] (lgl2) at (lgs2.east) {Related retry first};
\node[fill=vizNoFollow,minimum width=0.22cm,minimum height=0.15cm,inner sep=0,anchor=west] (lgs3) at (0,0.96) {};
\node[viztick,anchor=west,xshift=0.06cm] (lgl3) at (lgs3.east) {Neither successful follow-up};
\node[fill=vizUntriggered,minimum width=0.22cm,minimum height=0.15cm,inner sep=0,anchor=west,xshift=0.3cm] (lgs4) at (lgl3.east) {};
\node[viztick,anchor=west,xshift=0.06cm] (lgl4) at (lgs4.east) {Not triggered};
\end{tikzpicture}
\par\medskip
{\small\textbf{(b) Goal reached within each triggered-episode category}\par}
\begin{tikzpicture}[x=1cm,y=1cm,line cap=butt]
\draw[vizgrid] (0,-3.22) -- (0,0.2);
\node[viztick,anchor=north] at (0,-3.3) {0\%};
\draw[vizgrid] (1.925,-3.22) -- (1.925,0.2);
\node[viztick,anchor=north] at (1.925,-3.3) {25\%};
\draw[vizgrid] (3.85,-3.22) -- (3.85,0.2);
\node[viztick,anchor=north] at (3.85,-3.3) {50\%};
\draw[vizgrid] (5.775,-3.22) -- (5.775,0.2);
\node[viztick,anchor=north] at (5.775,-3.3) {75\%};
\draw[vizgrid] (7.7,-3.22) -- (7.7,0.2);
\node[viztick,anchor=north] at (7.7,-3.3) {100\%};
\node[viztick,anchor=north] at (3.85,-3.7) {Goal reached within each observed category};
\node[vizlab,anchor=east] at (-0.18,-0.3) {State query first};
\filldraw[fill=vizCommitted,draw=white,line width=0.5pt] (3.5,-0.12) circle (0.075);
\node[viztick,anchor=west] at (3.64,-0.12) {15/33};
\filldraw[fill=vizUncommitted,draw=white,line width=0.5pt] (3.005,-0.555) rectangle (3.155,-0.405);
\node[viztick,anchor=west] at (3.22,-0.48) {12/30};
\node[vizlab,anchor=east] at (-0.18,-1.4) {Related retry first};
\filldraw[fill=vizCommitted,draw=white,line width=0.5pt] (0,-1.22) circle (0.075);
\node[viztick,anchor=west] at (0.14,-1.22) {0/5};
\filldraw[fill=vizUncommitted,draw=white,line width=0.5pt] (2.125,-1.655) rectangle (2.275,-1.505);
\node[viztick,anchor=west] at (2.34,-1.58) {2/7};
\node[vizlab,anchor=east] at (-0.18,-2.5) {Neither successful follow-up};
\filldraw[fill=vizCommitted,draw=white,line width=0.5pt] (3.365,-2.32) circle (0.075);
\node[viztick,anchor=west] at (3.505,-2.32) {52/119};
\filldraw[fill=vizUncommitted,draw=white,line width=0.5pt] (-0.075,-2.755) rectangle (0.075,-2.605);
\node[viztick,anchor=west] at (0.14,-2.68) {0/116};
\node[circle,fill=vizCommitted,minimum size=0.13cm,inner sep=0,anchor=west] (lgs1) at (0,0.65) {};
\node[viztick,anchor=west,xshift=0.06cm] (lgl1) at (lgs1.east) {Effect present};
\node[fill=vizUncommitted,minimum size=0.12cm,inner sep=0,anchor=west,xshift=0.3cm] (lgs2) at (lgl1.east) {};
\node[viztick,anchor=west,xshift=0.06cm] (lgl2) at (lgs2.east) {No effect};
\end{tikzpicture}
\caption{Post-hoc C2 behavior and outcomes. (a) Each hidden world contains 180 assigned episodes; segment labels give counts, including episodes without an ambiguous reply. (b) Exact successes/episodes within the three triggered follow-up categories; circle: effect present, square: no effect. Category membership differs in task and configuration composition. These descriptive rates do not estimate the causal benefit of querying or retrying, response comprehension, or safe recovery. No classification was indeterminate.}\label{fig:c2-follow-up}
\end{figure}
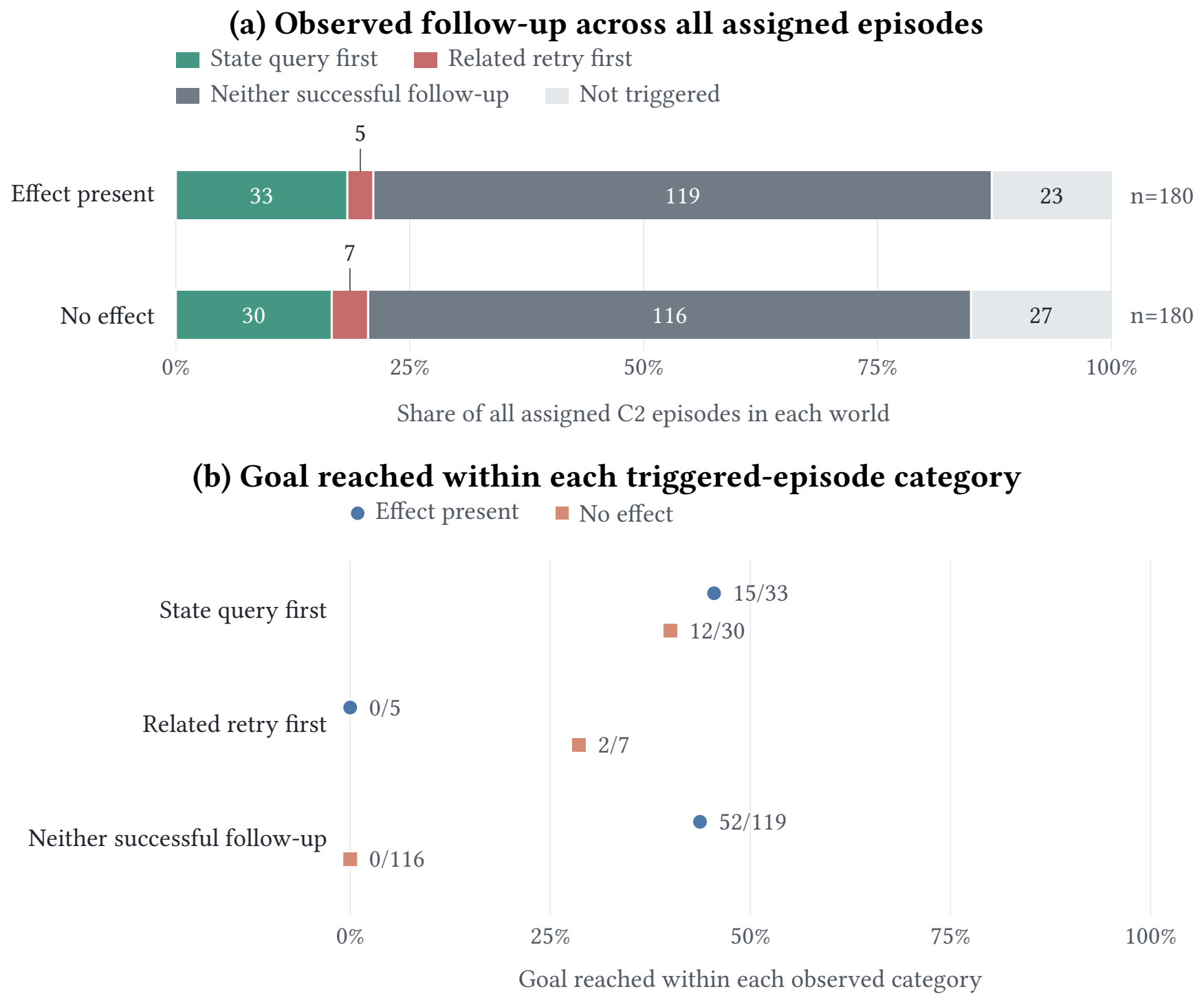

Figure~\ref{fig:c2-follow-up} separates the frequency of observed follow-up from completion within each category. In particular, the retry-first categories contain only five and seven episodes; their rates describe these records rather than a strategy-level reliability estimate.

\subsection{Terminal reasons and tool-call formats}
Of the 1,080 episodes, 823 ended with a submission (Table~\ref{tab:terminal}, Figure~\ref{fig:terminal}). Text-only replies ended 165 episodes, almost all of them in four configurations: L14 (64), L05 (53), L15 (29), and L04 (15). These are the four configurations with the lowest $S_G$, and no other configuration had more than one such episode. Of the 165 final replies, 63 were tool calls in the XML-like format (60 from L14 and 3 from L15), 4 were bare JSON objects (all from L14), and 98 were other text, which includes final summaries (Figure~\ref{fig:text-replies}). The text-only terminations of L14 therefore mostly reflect the parser incompatibility described in Section~\ref{sec:collection}, whereas the final replies of L05 and L04 were all other text. The other episodes ended on stuck detection (35, of which 10 in L13 and 6 in L10), the per-episode output limit (23, of which 10 in L13 and 7 in L11), truncated output without an action (12), the request limit (11), and a tool-call parse error (11). No episode ended on a time limit or for an infrastructure reason.

\subsection{Resources within platforms}
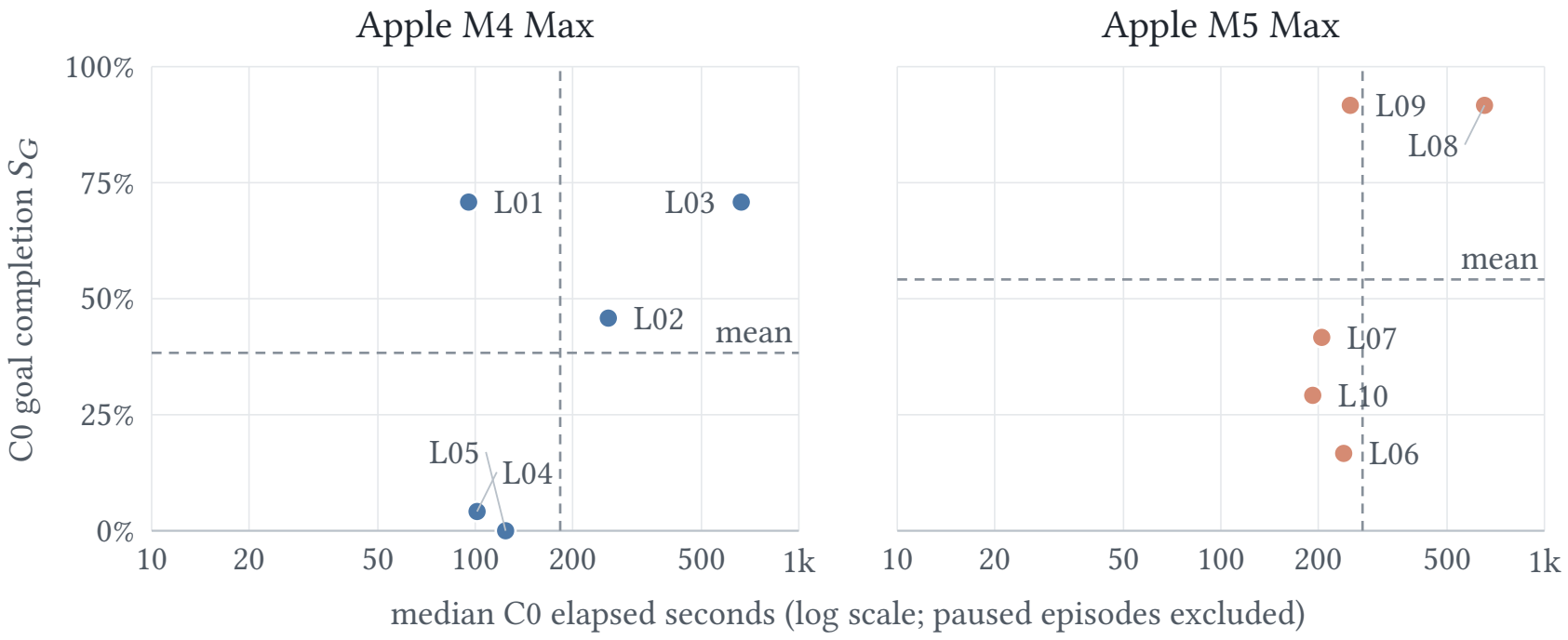
\begin{figure}[tbp]
\centering
\begin{tikzpicture}[x=1cm,y=1cm,line cap=butt]
\draw[vizgrid] (0,0) -- (4.6,0);
\node[viztick,anchor=east] at (-0.08,0) {0\%};
\draw[vizgrid] (0,0.825) -- (4.6,0.825);
\node[viztick,anchor=east] at (-0.08,0.825) {25\%};
\draw[vizgrid] (0,1.65) -- (4.6,1.65);
\node[viztick,anchor=east] at (-0.08,1.65) {50\%};
\draw[vizgrid] (0,2.475) -- (4.6,2.475);
\node[viztick,anchor=east] at (-0.08,2.475) {75\%};
\draw[vizgrid] (0,3.3) -- (4.6,3.3);
\node[viztick,anchor=east] at (-0.08,3.3) {100\%};
\draw[vizgrid] (0,0) -- (0,3.3);
\node[viztick,anchor=north] at (0,-0.08) {10};
\draw[vizgrid] (0.692,0) -- (0.692,3.3);
\node[viztick,anchor=north] at (0.692,-0.08) {20};
\draw[vizgrid] (1.608,0) -- (1.608,3.3);
\node[viztick,anchor=north] at (1.608,-0.08) {50};
\draw[vizgrid] (2.3,0) -- (2.3,3.3);
\node[viztick,anchor=north] at (2.3,-0.08) {100};
\draw[vizgrid] (2.992,0) -- (2.992,3.3);
\node[viztick,anchor=north] at (2.992,-0.08) {200};
\draw[vizgrid] (3.908,0) -- (3.908,3.3);
\node[viztick,anchor=north] at (3.908,-0.08) {500};
\draw[vizgrid] (4.6,0) -- (4.6,3.3);
\node[viztick,anchor=north] at (4.6,-0.08) {1k};
\draw[vizaxis] (0,0) -- (4.6,0);
\draw[vizMuted,line width=0.5pt,dash pattern=on 2pt off 1.5pt] (0,1.265) -- (4.6,1.265) (2.904,0) -- (2.904,3.3);
\node[viztick,anchor=south east] at (4.6,1.295) {mean};
\node[viztitle,anchor=south] at (2.3,3.4) {Apple M4 Max};
\filldraw[fill=vizP1,draw=white,line width=0.5pt] (2.253,2.337) circle (0.07);
\node[viztick,anchor=west] at (2.393,2.337) {L01};
\filldraw[fill=vizP1,draw=white,line width=0.5pt] (4.191,2.337) circle (0.07);
\node[viztick,anchor=east] at (4.051,2.337) {L03};
\filldraw[fill=vizP1,draw=white,line width=0.5pt] (3.246,1.512) circle (0.07);
\node[viztick,anchor=west] at (3.386,1.512) {L02};
\filldraw[fill=vizP1,draw=white,line width=0.5pt] (2.313,0.137) circle (0.07);
\draw[vizAxis,line width=0.35pt] (2.313,0.137) -- (2.453,0.417);
\node[viztick,anchor=west] at (2.453,0.417) {L04};
\filldraw[fill=vizP1,draw=white,line width=0.5pt] (2.516,0) circle (0.07);
\draw[vizAxis,line width=0.35pt] (2.516,0) -- (2.376,0.56);
\node[viztick,anchor=east] at (2.376,0.56) {L05};
\draw[vizgrid] (5.3,0) -- (9.9,0);
\draw[vizgrid] (5.3,0.825) -- (9.9,0.825);
\draw[vizgrid] (5.3,1.65) -- (9.9,1.65);
\draw[vizgrid] (5.3,2.475) -- (9.9,2.475);
\draw[vizgrid] (5.3,3.3) -- (9.9,3.3);
\draw[vizgrid] (5.3,0) -- (5.3,3.3);
\node[viztick,anchor=north] at (5.3,-0.08) {10};
\draw[vizgrid] (5.992,0) -- (5.992,3.3);
\node[viztick,anchor=north] at (5.992,-0.08) {20};
\draw[vizgrid] (6.908,0) -- (6.908,3.3);
\node[viztick,anchor=north] at (6.908,-0.08) {50};
\draw[vizgrid] (7.6,0) -- (7.6,3.3);
\node[viztick,anchor=north] at (7.6,-0.08) {100};
\draw[vizgrid] (8.292,0) -- (8.292,3.3);
\node[viztick,anchor=north] at (8.292,-0.08) {200};
\draw[vizgrid] (9.208,0) -- (9.208,3.3);
\node[viztick,anchor=north] at (9.208,-0.08) {500};
\draw[vizgrid] (9.9,0) -- (9.9,3.3);
\node[viztick,anchor=north] at (9.9,-0.08) {1k};
\draw[vizaxis] (5.3,0) -- (9.9,0);
\draw[vizMuted,line width=0.5pt,dash pattern=on 2pt off 1.5pt] (5.3,1.787) -- (9.9,1.787) (8.607,0) -- (8.607,3.3);
\node[viztick,anchor=south east] at (9.9,1.817) {mean};
\node[viztitle,anchor=south] at (7.6,3.4) {Apple M5 Max};
\filldraw[fill=vizP2,draw=white,line width=0.5pt] (8.52,3.025) circle (0.07);
\node[viztick,anchor=west] at (8.66,3.025) {L09};
\filldraw[fill=vizP2,draw=white,line width=0.5pt] (9.474,3.025) circle (0.07);
\draw[vizAxis,line width=0.35pt] (9.474,3.025) -- (9.334,2.745);
\node[viztick,anchor=east] at (9.334,2.745) {L08};
\filldraw[fill=vizP2,draw=white,line width=0.5pt] (8.317,1.375) circle (0.07);
\node[viztick,anchor=west] at (8.457,1.375) {L07};
\filldraw[fill=vizP2,draw=white,line width=0.5pt] (8.253,0.963) circle (0.07);
\node[viztick,anchor=west] at (8.393,0.963) {L10};
\filldraw[fill=vizP2,draw=white,line width=0.5pt] (8.473,0.55) circle (0.07);
\node[viztick,anchor=west] at (8.613,0.55) {L06};
\node[viztick,anchor=south,rotate=90] at (-0.75,1.65) {C0 goal completion $S_G$};
\node[viztick,anchor=north] at (4.95,-0.45) {median C0 elapsed seconds (log scale; paused episodes excluded)};
\end{tikzpicture}
\caption{Exploratory C0 quality--time view. Goal completion uses all 24 C0 episodes. Median C0 elapsed time excludes one paused episode each for L09 and L10, leaving 23 timing observations for these configurations and 24 for the others. Compare within a platform only; the multi-instance RTX 3090 arm is omitted. Dashed lines mark within-platform means (geometric on the logarithmic time axis).}\label{fig:time-score}
\end{figure}

Figure~\ref{fig:time-score} pairs C0 completion with median C0 elapsed time on the two single-machine platforms. Both axes refer to the same baseline task inventory; the two paused episodes are omitted from timing only. On the M5 Max, L09 and L08 both reached 22 tasks, with median C0 times of 251 and 653 seconds; L09's timing has 23 observations and L08's has 24. On the M4 Max, L01 and L03 both reached 17, with 95 and 664 seconds over 24 observations each. Figure~\ref{fig:tokens-score} uses mean output tokens across all 24 C0 episodes, including failures. These exploratory configured-system views are not controlled hardware or compute-efficiency comparisons.

Figure~\ref{fig:c0-output-per-goal} adds output-only accounting per goal reached: total output tokens across all 24 assigned C0 episodes divided by the number of goals reached. Thus failures contribute usage rather than disappearing from the numerator. For L08 and L09 this is 17,763 and 6,606 tokens per reached goal, and for L01 and L03 it is 11,104 and 27,137. The ratio is undefined for the three zero-goal configurations. Runtime-reported tokens use model-specific tokenizers; neither this ratio nor elapsed time represents total compute, energy, input-token use, or monetary cost. All-condition resource distributions remain in Figures~\ref{fig:tokens} and~\ref{fig:effort}.

\subsection{Sensitivity views}
Without L14, the pooled contrasts are nearly unchanged: 52 worsened and 20 improved C1 pairs, a mean $\Delta_1$ of $-0.095$, and a mean $\Delta_2$ of $-0.176$ (Table~\ref{tab:contrasts}). The driver split is descriptive only (Figure~\ref{fig:drivers}). L11 reached the goal in 6 of its 15 C0 episodes on driver 580.173.02 and in 2 of 9 on driver 615.71.09, L13 in 4 of 15 and 3 of 9, and L15 in none on either. The family-resampling ranges of the two contrasts are in Figure~\ref{fig:contrasts}.

\section{Interpretation, Validity, and Research Use}\label{sec:discussion}

\subsection{What a score describes}
A score in this study describes a registered configuration: one model at one quantization, served by Ollama 0.40.14 on its documented platform, driven by the OpenHands SDK 1.61.0 with task-specific tools, using its default reasoning and sampling, under one frozen resource contract. This is the unit of comparison, incorporating the model weights and the serving environment. Differences between configurations on different platforms combine model, backend, and hardware; changes to a serving stack, framework, reasoning setting, or budget define a separate comparison.

\subsection{Reading the results}
Three patterns distinguish the evaluated configurations. First, completion is concentrated: four configurations reached at least 17 of the 24 goals in C0, while nine reached at most eight. Second, the C1 matrix contains more success-to-failure transitions than the reverse, although many corrections arrive before an agent commits to an action and therefore do not require undoing an effect (Section~\ref{sec:task-validity}). This identifies correction uptake as a useful diagnostic alongside baseline completion. Third, the exploratory C2 split separates 67 committed-world completions from 14 uncommitted-world completions. The operation-matched reanalysis adds a concrete behavioral distinction: \CtwoNoFollowUpEpisodes{} of \CtwoTriggeredEpisodes{} triggered episodes contain neither a successful target-covering query nor a successful related retry, despite 214 of them ending with a submission. Goal completion in that category is 52/119 when committed and 0/116 when uncommitted. The counts establish follow-up behavior in the recorded event order, while leaving response comprehension and causal benefit distinct. Evaluating both worlds reveals a distinction that a committed-world-only test would miss. These findings characterize the documented configurations on the twenty-four-task population.

\subsection{Design scope and measurement resolution}
The fixed matrix defines the resolution and interpretation of the measurements.
\begin{itemize}
\item \textbf{Twenty-four tasks.} One task contributes about 4.2 percentage points to the primary score. Exact task counts and family-resampling ranges provide the appropriate resolution for interpreting neighboring configurations.
\item \textbf{One instance per family.} Each family contributes one fixed problem. Cross-family coverage is measured through the variety of objectives and scenario groups; within-family instance variation is outside the collected matrix.
\item \textbf{One run per task and condition.} C0 and C1 each contain one run per task, and C2 contains one run per hidden world. The resulting contrasts describe the collected paired outcomes, without estimating repeated-run variability. Registered non-zero sampling temperatures remain part of each configuration.
\item \textbf{Fixed budget.} The limits are the widest of the candidate tiers. The output limits were raised once before freezing after development episodes showed truncation. Episodes that reached the frozen output limit remain in the reported results.
\item \textbf{Vendor defaults.} Default reasoning is a configuration policy rather than equalized computation. Reasoning output and time are included in resource accounting.
\item \textbf{Fixed configuration factors.} Budget size, framework, reasoning policy, text-only termination, and serving platform follow the registered configuration. Their individual effects require distinct controlled comparisons; the current study compares the complete configurations.
\end{itemize}

\subsection{Serving-stack compatibility}
The collection showed that the serving stack can dominate a model's outcome. Most of L14's text-only terminations were tool calls in a format that this Ollama version did not parse (Section~\ref{sec:collection}). Under the frozen rule these count as model outcomes, but they mostly measure the compatibility between the model's tool-call format and the parser rather than the model's ability to call tools. The rule ending an episode at the first text-only reply, without a reminder, was fixed in advance precisely so that it would not be tuned to outcomes. It also means that a model that writes a final summary instead of calling \code{finish} is scored as incomplete even when its last state is correct. Terminal reasons and the classification of text-only replies are reported so that readers can separate these cases.

\subsection{Hardware and the multi-instance deviation}
The three arms ran on different platforms, and one arm ran on five rented instances on two machines with different GPU driver versions. In B's duplicate comparison, the sampled same-instance pairs agreed token for token, while the cross-driver pairs diverged despite fixed seeds. Three models therefore have results from two driver configurations. The analysis reports their driver split descriptively because episode assignment to machines was not randomized. In another arm, the host paused two requests, most likely by sleeping; the recorded inputs and seeds were unchanged, and both episodes remained within their limits. Time and resource comparisons are restricted to one platform, with paused episodes marked and omitted from elapsed-time comparisons.

\subsection{Task validity}\label{sec:task-validity}
The original synthetic tasks provide explicit rules, bounded state, and checkable outcomes without relying on any company's systems. Booking, order, ledger, and ticket worlds expose the operations declared in their cards; the software families use compact packages, and the image-based family supplies a fixed transcription. These choices define a controlled evaluation setting. Applying the findings to larger production systems or native visual interaction requires evaluation in those settings. The cards also specify intervention scope: many C1 corrections arrive before an agent commits to an action, so C1 primarily measures correction uptake, while C2 makes the first call of one designated operation ambiguous.

Scorer validity is supported by offline tests, negative controls, adversarial review, and the separate reconstruction of formal outcomes. The second scorer reproduced every goal outcome and identified three evidence outcomes for task-card-based revision. Section~\ref{sec:verification-status} specifies its shared implementation history, information access, and adjudication procedure. The two implementations agree on the collected goal outcomes; common specification or interpretation errors remain a distinct target for task review.

\subsection{Research use}
The benchmark provides a documented task population for comparing configured agents, examining responses to corrections and ambiguous replies, and diagnosing outcomes from state and event records. Its scores measure the declared task predicates rather than universal general intelligence, security assurance, or deployment certification. The insurance and medication documents evaluate reading comprehension; the results do not constitute legal, medical, or financial advice.

\section{Research Materials and Access}\label{sec:availability}
The article and its accompanying source package provide the task specifications, condition definitions, scoring criteria, configuration descriptions, result tables, and figures. The accompanying publication package also includes machine-readable aggregate results and verification receipts binding the reported results to the frozen study and the adopted scoring adjudications.

The authors retain the versioned task environments, runner, scoring implementations, and episode archives. Requests for access to these additional research materials can be directed to \texttt{research@accentrust.com}; distribution is handled by the authors following the applicable confidentiality and third-party licensing checks. Executable code and raw episode archives follow this author-mediated access route separately from the publication package. The coordinator's private assignment salt and operational machine records are excluded from the publication package. All task data are synthetic; no production customer records, company application code, or private business-system architecture are part of the study.

\section{Conclusion}\label{sec:conclusion}
AGIBenchmark combines twenty-four original synthetic tasks, matched information conditions, and state-based outcome checks for agents handling daily, enterprise, and technical requests. All tasks run under a final-information baseline and an explicit correction with matched final obligations. Twelve state-changing tasks additionally run under ambiguous replies in two hidden effect worlds. The completed study comprises 1,080 episodes from fifteen open-weight models through a common OpenHands framework, task tools, and frozen resource contract. Collection deviations and serving-stack behavior are documented, and a separate scoring implementation reconstructed every goal outcome and led to three evidence rescorings. Under C0, two configurations reached 22 of the 24 goals and three reached none. C1 contained 52 success-to-failure and 20 failure-to-success transitions. The exploratory C2 breakdown found 67 of 180 completions when the operation had taken effect and 14 of 180 when it had not, compared with 70 of 180 matched C0 episodes. A post-hoc operation-matched reconstruction identified \CtwoVerificationFirstEpisodes{} state-query-first episodes, \CtwoRetryFirstEpisodes{} related-retry-first episodes, and \CtwoNoFollowUpEpisodes{} with neither qualifying successful follow-up among the \CtwoTriggeredEpisodes{} episodes with an ambiguous reply. These exact outcomes establish a configured-agent baseline and distinguish ordinary completion, correction uptake, and verification of uncertain effects as separate evaluation dimensions.

\bibliographystyle{ACM-Reference-Format}
\bibliography{references}
\clearpage
\appendix
\section{Worked Task Example}
\label{app:examples}
This appendix follows one formal family, \code{travel-test-01}, through its world, tools, conditions, and scoring. It illustrates the level of specification behind every family in Appendix~\ref{app:task-family-catalog}. The example describes the task contract, not a model trajectory.

\subsection{Request and world}
A fictional traveler is going from Harbourton to Calderford for a conference that starts at 14:00 on the first day and ends at 17:00 two days later. She needs an outbound train and a hotel, and she already holds a return ticket. After a timetable change, that ticket leaves only four minutes to change trains, while the travel rules require at least twelve. The agent must book the outbound journey and the hotel, fix the return, and send her the itinerary.

The simulated booking system has three stations, one of them a junction for connections. On the final outbound date there are six journeys, direct or with a change, each in first or second class and as a non-refundable Saver or a fully refundable Flex fare. Three distractors are built in. The cheapest journey has an 8-minute connection, another cheap journey arrives too late, and a first-class fare looks like good value but is forbidden by the travel policy. Six return journeys exist on the last day. Two are legal targets for the disrupted ticket: a later departure with a 24-minute connection and a direct train. One leaves too early, one is first class, and one arrives too late. Five hotels differ in distance from the venue, breakfast, and price, and three meet the rule of being within 1.5 km with breakfast. The budget is 290.00 and counts only new spending: new fares, the hotel total, and any fare lost by cancelling a Saver ticket. A free rebooking caused by the timetable change costs nothing. Under the final information, exactly three outbound journeys combined with one of the hotels for two nights fit the budget.

\subsection{Tools}
The agent has \code{read\_task}, \code{list\_trip\_bookings}, \code{search\_rail\_journeys}, \code{search\_hotels}, \code{book\_rail\_journey}, \code{rebook\_rail\_booking}, \code{cancel\_rail\_booking}, \code{book\_hotel\_stay}, \code{cancel\_hotel\_stay}, \code{send\_trip\_itinerary}, and \code{finish}. Rebooking is allowed only for a booking affected by the timetable change, on the same line and day, in second class, and free of charge. Every tool validates its arguments and returns an error message explaining any rejection.

\subsection{Scoring}
\begin{itemize}
\item \textbf{G}: the agent called \code{finish} and sent an itinerary, and the final state holds exactly one valid outbound ticket, one valid return ticket, and one hotel stay. The outbound ticket must be on the final date, arrive before 13:00, and have every transfer of at least 12 minutes. The return ticket must leave after 17:30, arrive before 23:00, and have transfers of at least 12 minutes. The hotel must be within 1.5 km with breakfast for the right nights, and new spending must not exceed 290.00. Every admissible combination is accepted.
\item \textbf{P}: no first-class ticket was ever booked, the agent never held two tickets for the same direction or two hotel stays at once, and the original return ticket was not cancelled, because the rules say to rebook it.
\item \textbf{E}: the last itinerary sent matches the final bookings and dates, and its reported new spending equals the spending computed by the rules within 0.01. This includes the fare lost if a paid Saver ticket was cancelled.
\end{itemize}

\subsection{Conditions}
In C0, the final dates are known from the start. In C1, the task first states that the conference starts one day later, so the outbound journey and the hotel would move by a day. At the first call of a search or booking tool, the result carries a correction notice moving the conference back to the final date, and \code{read\_task} shows the corrected dates from then on. A solution that keeps the old dates fails G. In C2, the first outbound booking returns \code{submitted} with the message that it was passed to the operator and the result is not yet confirmed. The text is the same in both worlds. In the committed world the ticket exists, and booking again leaves two outbound tickets, which violates P and also fails G. In the uncommitted world there is no ticket, and finishing without checking leaves the outbound leg missing, so G fails. \code{list\_trip\_bookings} shows the held bookings and the result of every request.

\subsection{Reference route and negative controls}
The reference route reads the task and existing bookings, searches outbound and return journeys and hotels, and picks the cheapest legal combination. It books the outbound journey, checking the booking list after an ambiguous reply in C2, then rebooks the return to the earliest legal journey, books the hotel, checks the bookings to compute the spending, sends the itinerary, and finishes. It needs 11 to 13 tool calls. The seven negative controls are listed below; each is a scripted wrong solution that must fail on its named endpoint:
\begin{enumerate}
\item keeping the old dates after the C1 correction (G);
\item booking again without checking in the committed C2 world (P);
\item treating \code{submitted} as success in the uncommitted C2 world (G);
\item booking the cheap first-class fare, which meets time and budget so that only P fails (P);
\item leaving the broken return ticket unchanged (G);
\item always taking the first listed option, which exceeds the budget (G);
\item misreporting the total in the itinerary (E).
\end{enumerate}

\subsection{Known limitations}
The card records four limitations. The correction arrives at the first search, usually before anything is booked, so C1 tests searching again under the correction rather than undoing a booking. The rebooking tool rejects first-class and cross-line targets, so a policy violation can only occur through a new booking. The rules point to the existing return ticket, so the agent must compute the broken connection but is told to look. In the committed C2 world, a blind rebooking fails G as well as P.

\section{Records and Reporting Contract}
\label{app:reporting}
This appendix lists what is recorded for every formal episode and what the results report must contain. The purpose is that every summary can be traced back to episode records and every episode record to the frozen code that produced it.

\subsection{Per-episode records}
Each episode produces four files and one line in the arm's result list.
\begin{itemize}
\item \textbf{Environment record}: the final state snapshot and the ordered event log of the task environment, including every tool call that reached it with its arguments, any error it raised, and whether the C1 correction was delivered at that call. The tool results themselves are in the request log, where they appear as observations in the next request. The record is written before scoring, so a scorer failure cannot lose the state.
\item \textbf{Trace}: the tool calls as seen by the agent, including calls the SDK rejected before they reached the environment.
\item \textbf{Summary}: score, terminal-relevant flags, counts, timing, seed, and settings. It is written last and atomically, and its presence marks the episode as scored.
\item \textbf{Request log}: one line per model request with the full request, the full response including reasoning text and tool calls, the HTTP status, the elapsed time, and the token counts reported by the server.
\end{itemize}
The result line contains the task key, model and device identifiers, the C2 world, G, P, E, J, the C1 delivery flag, the terminal reason, and the counts of requests, failed and refused requests, tool calls, and requests without a tool call. It also has output tokens in total and per request, the peak prompt size, reasoning characters, throughput, the options actually sent, the reasoning parameters sent (always empty under vendor default), the server's context length, model memory, the runner and task-code fingerprints, the package and Ollama versions, and the preflight result.

\subsection{Per-arm records}
Each arm also contains its manifest of 72 task keys per model with their worlds, the frozen budget, the model digests, one registration probe per model with capabilities, template hash, and default parameters, and an environment record of the host and software. The RTX 3090 arm adds a per-episode instance map and an environment record for each instance. Each returned archive is identified by its SHA-256 hash in the coordinator's check record.

\subsection{What the results report contains}
\begin{enumerate}
\item Coverage against the assigned ledger of 1,080 episodes, by configuration and condition, with every non-scored state named.
\item The C0 goal-completion summary per configuration with its eight group scores and twenty-four family outcomes.
\item The C1 and C2 contrasts per configuration as counts of improved, unchanged, and worsened families, the restricted C0 baseline beside C2, and the pooled counts across configurations.
\item C1 delivery rates, the rate of ambiguous replies actually produced in C2, the original tool-name query proxy after the reply, and committed-world process violations, each with its denominator. The post-hoc operation-matched follow-up classification is reported separately, with its target and successful-call rules, untriggered episodes, and any indeterminate records.
\item P, E, and J with their applicable denominators, and the partition of G into J and its two gap terms.
\item Terminal reasons per configuration, with the classification of text-only replies.
\item Resource use per configuration, compared only within a platform, with episodes affected by host pauses marked.
\item The sensitivity views of Section~\ref{sec:analysis}: family resampling, leave-one-family-out, the driver split, and the pooled contrasts without L14.
\item The second-scorer comparison: agreements, disagreements, their resolution, and any rescoring.
\end{enumerate}
Nothing in the report is selected by outcome. A configuration's weak group is shown as prominently as its strong one, and a contrast of zero or the opposite sign is reported in the same form as any other.

\section{Task Family Catalogue}
\label{app:task-family-catalog}
This appendix summarizes the twenty-four formal families. Each entry gives the task, the goal predicate G, the C1 correction with its trigger, and, for the twelve state-changing families, the C2 operation with its verification route. The binding specification of each family is its task card, which also defines P, E, the reference route, the negative controls, and known limitations. People, organizations, places, products, and documents are fictional. Under C1, the solution that follows the superseded information always fails G, and in C2 a blind retry in the committed world violates P while finishing without verification in the uncommitted world fails G.

\subsection{Travel and scheduling (daily; C0, C1, C2)}
\paragraph{\code{travel-test-01}: conference trip with a disrupted return.} A traveler needs an outbound train and a hotel for a conference, and her existing return ticket no longer leaves enough time to change trains after a timetable change. The agent searches journeys and hotels, books the outbound leg, rebooks the return, books the hotel, and sends the itinerary. \emph{G}: the final state holds exactly one valid outbound ticket (arriving before 13:00 on the first day, every transfer at least 12 minutes), one valid return ticket (leaving after 17:30, arriving before 23:00, transfers at least 12 minutes), and one hotel stay within 1.5 km with breakfast. New spending must not exceed 290.00, and the itinerary must have been sent. All admissible combinations are accepted. Distractors include an 8-minute connection, a cheap first-class fare forbidden by policy, and a hotel outside the distance limit. \emph{C1}: the conference start moves one day earlier, delivered at the first search or booking call. \emph{C2}: the first rail booking returns \code{submitted}; the booking list shows the true state.

\paragraph{\code{travel-test-02}: project review across time zones.} An organizer needs a 60-minute review on one of three days with four required and two optional participants in three time zones. The agent checks calendars, reserves a room, sends invitations, and confirms with the organizer. \emph{G}: exactly one valid room reservation of 60 minutes on the half-hour grid, in a slot where every required participant is free and within working hours. The slot must maximize the number of available optional participants, with ties accepted. The room must have video and enough capacity, and the reservation's invitation must cover all six invitees. \emph{C1}: one remote participant's calendar was wrong, so the slot that was uniquely optimal under the old calendar is no longer feasible; delivered at the first query of that participant's busy times. \emph{C2}: the first invitation returns \code{pending}; the invitation list shows the true state.

\paragraph{\code{travel-test-03}: two-day itinerary on a budget.} A traveler wants a weekend plan that includes three required attractions and, if the budget allows, a favorite extra one. The plan must respect opening hours, travel times, and timed-entry slots. The agent reserves the timed entries and saves the plan. \emph{G}: the last saved plan includes the required attractions, at least five distinct attractions with at least two per day, and a feasible schedule from 09:00 with every travel time and opening window respected. It must also return to the lodging within the daily limit, hold exactly the reservations it uses, and cost at most 85.00. The extra attraction must be included whenever a compliant plan with it exists, which under the final data is not the case. \emph{C1}: the budget falls from 95.00 to 85.00, which excludes the extra attraction; delivered at the first slot check or reservation. \emph{C2}: the first timed-entry reservation returns \code{processing}; the reservation list shows the true state and the fees paid.

\subsection{Personal affairs and purchasing (daily; C0, C1)}
\paragraph{\code{personal-test-01}: bills and subscriptions.} A household asks for its card statement to be reconciled against its bills and subscription terms. The agent disputes wrong charges in a simulated account portal, cancels the plans the household wants stopped, and sends a report. \emph{G}: every required dispute is filed with the correct reason and amount, with exactly one of a duplicated pair disputed. The two plans to be stopped are cancelled, and the report has been sent. \emph{C1}: a subscription's price increase takes effect a month later than first stated, which makes one charge disputable; delivered at the first read of that plan's terms.

\paragraph{\code{personal-test-02}: constrained purchase.} A user needs a monitor, a stand, and a docking station that fit together, arrive in time, and stay within budget, bought in one order at the lowest total. The agent compares a frozen catalogue, places one order, and explains the purchase. \emph{G}: the last order contains exactly one item of each kind, satisfies every compatibility and delivery constraint, and costs the lowest total among all compliant combinations, ties accepted. The scorer enumerates all combinations. \emph{C1}: the user's charger is 90 W, not 65 W, so the dock must deliver at least 90 W; delivered at the first product-detail lookup.

\paragraph{\code{personal-test-03}: insurance and medication leaflet.} Given a home-insurance policy and a medication leaflet, the agent answers ten questions about specific situations strictly from the documents and cites the clauses. \emph{G}: all ten answers are correct, with numbers to two decimals, yes or no, or \code{not\_stated} when the documents do not say. The task tests reading comprehension only and scores no personal legal or medical conclusion. \emph{C1}: a policy section is superseded and the water-damage deductible rises from 250 to 400; delivered at the first read of that section.

\subsection{Everyday inquiry (daily; C0, C1)}
\paragraph{\code{inquiry-test-01}: residential parking permit.} A user who recently moved wants a parking permit for a leased electric car. From a frozen municipal guide, the agent determines eligibility, zone, steps, required documents, fee, and deadline, and cites a section for each. \emph{G}: all six fields are correct, with every legal combination of proof documents accepted. \emph{C1}: the guide page is revised and a lease agreement alone is no longer accepted; delivered at the first opening of that page, which still returns the old text with the notice.

\paragraph{\code{inquiry-test-02}: ferry trip with conflicting sources.} For a ferry trip with a bicycle on a given Saturday, twelve sources disagree. Applying the task's source-priority rule, the agent gives the value valid on the day for five claims and cites the deciding source. \emph{G}: all five values are correct under type-tolerant normalization of times, numbers, and yes or no. \emph{C1}: a publisher corrects one source so that it no longer covers the travel date, which flips the bicycle answer; delivered at the first read of that source.

\paragraph{\code{inquiry-test-03}: thermostat receipt, screenshots, and manual.} From a receipt photo, a display photo, an app screenshot, manual pages, and a return policy, the agent answers five questions. The questions cover the amount paid, two return deadlines, the manual step for a display fault, and the settings to change before enabling an eco schedule, and each answer cites its image region or document. Image content is delivered as fixed text transcriptions (Appendix~\ref{app:perception}). \emph{G}: all five answers are correct under the stated normalization. \emph{C1}: one screenshot turns out to show the wrong device and is replaced, which changes the settings answer; delivered at the first view of that screenshot.

\subsection{Customer service and orders (enterprise; C0, C1, C2)}
\paragraph{\code{service-test-01}: returns and refunds.} A customer requests returns, an exchange, and refunds for five items of one order. The agent applies the store's policy to each item, authorizes returns or exchanges, submits refunds, and replies. \emph{G}: returns are authorized for exactly the eligible lines with the right resolution, the exchange goes to the correct variant, and completed refunds cover exactly the refundable lines with the correct amounts. The customer must have received a reply. \emph{C1}: the return window is 45 days, not 30, which makes one item refundable; delivered at the first read of the policy. \emph{C2}: the first refund returns \code{processing}; the refund list or order lookup shows whether it completed or failed.

\paragraph{\code{service-test-02}: order change and reshipment.} A customer asks to change an unpicked order (more units, one line removed, a new address, a delivery method chosen by rule) and to reship a missing item from a delivered order. \emph{G}: the changed order has exactly the right lines, address, and method. Payments and adjustments must equal the order total computed from the final state, and exactly one reshipped unit goes to the new address. Stock levels must match the final state, and the customer must have been updated. \emph{C1}: a stock count releases two held units, so the full requested quantity is available; delivered at the first stock query or quantity change. \emph{C2}: the first reshipment returns \code{pending}; the order details show whether it shipped or was cancelled.

\paragraph{\code{service-test-03}: ticket triage and escalation.} Six overnight tickets must be categorized, prioritized, assigned to an agent of the right team who speaks the language and has capacity, and escalated where the rules require. \emph{G}: every ticket has the correct category and priority, its assignee belongs to the right team and speaks the language, and no agent exceeds capacity. The set of escalation targets of every ticket must be correct. \emph{C1}: a customer update turns one access ticket into an account-wide outage that requires a different team and an incident escalation; delivered at the first ticket listing or read. \emph{C2}: the first escalation returns \code{submitted}; the ticket's escalations and status show the true state.

\subsection{Finance and procurement (enterprise; C0, C1, C2)}
\paragraph{\code{finance-test-01}: three-way invoice matching.} Six supplier invoices must be matched line by line against purchase orders and goods receipts. The agent lists every discrepancy, posts only fully matching invoices, and submits a matching report. \emph{G}: the set of invoices actually posted equals the matching set, and the last report's discrepancy set is exact. \emph{C1}: a goods receipt is corrected for rejected units, which turns one invoice into a quantity discrepancy; delivered at the first read of that receipt. \emph{C2}: the first posting returns \code{submitted}; the invoices' ledger status and journal entries show the true state.

\paragraph{\code{finance-test-02}: expense review.} Five pending expense reports must be reviewed against a synthetic travel policy. Each receives a decision (full approval, partial approval, or return), an approved amount, and the clauses applied. One report in the queue was already reimbursed and must not be touched. \emph{G}: every pending report has an effective decision whose type and amount match the policy. \emph{C1}: one nightly limit rises from 140.00 to 150.00, which changes one partial approval; delivered at the first read of that report. \emph{C2}: the first decision returns \code{submitted}; the report's status and decisions show the true state.

\paragraph{\code{finance-test-03}: supplier selection and order.} For office chairs, six quotes must be compared on validity, delivery time, supplier registration, minimum order, and tiered prices. The agent orders from the cheapest compliant quote within budget and explains the choice. \emph{G}: exactly one effective purchase order, from the cheapest compliant quote at the final quantity, with the correct tier price and total. \emph{C1}: the requested quantity rises from 36 to 44, which moves a different quote into a cheaper tier; delivered at the first quote lookup. \emph{C2}: the first purchase order returns \code{pending}; the purchase-order list shows the true state.

\subsection{Office documents and collaboration (enterprise; C0, C1)}
\paragraph{\code{office-test-01}: meeting notes to tasks.} Seventeen lines of meeting notes must become tracked tasks with owners, due dates, dependencies, and source lines. Existing tasks changed in the meeting are updated rather than duplicated, and a recap is sent. \emph{G}: exactly the five required new tasks exist with the right owners, due dates, and dependencies, and the existing task carries its new due date and is otherwise unchanged. \emph{C1}: one deadline is ``in ten days'', not ``in two weeks''; delivered at the first read of the notes.

\paragraph{\code{office-test-02}: revision against a fact sheet.} A draft annual report must be revised against an approved fact sheet. Only values that disagree are changed, new values follow the house format, protected sections stay untouched, and a revision log is submitted. \emph{G}: all nine paragraphs of the final text equal the expected text after whitespace normalization. \emph{C1}: a new fact-sheet revision corrects a date that previously matched the draft; delivered at the first read of the fact sheet.

\paragraph{\code{office-test-03}: weekly report from a table.} A weekly sales report must be computed from a versioned order table by the reporting conventions, filled into a template, previewed, and published once with the data version and included orders. \emph{G}: the last published values equal those computed from the final table version within the stated tolerances. \emph{C1}: the order table is corrected (one region and one amount), which changes the top region and revenue; delivered at the first read of the table.

\subsection{Software engineering (technical; C0, C1, C2)}
\paragraph{\code{software-test-01}: defect repair.} A small delimited-text parsing package returns a stray quotation mark when a quoted label ends a line. The agent must find the cause, repair the package so that the whole contract holds, run the public tests, and submit one change request. \emph{G}: protected files are unchanged, and public tests, 13 hidden tests, and three conversion cases produce exactly the contracted results. \emph{C1}: the rule for blank numbers changes from zero to missing; delivered at the first successful write of an editable file. \emph{C2}: the first change request returns \code{processing}; the change-request list shows the true state.

\paragraph{\code{software-test-02}: interface migration.} A small shop package must move money from floating-point currency units to integer cents. The agent adds a cents-based price function, migrates every caller, handles the legacy function, runs the tests, and submits one change request. \emph{G}: protected files are unchanged, and public tests and 19 hidden tests produce exactly the contracted results. A syntax-tree check must also confirm the required handling of the legacy identifier. \emph{C1}: the legacy function must be removed, not kept as a shim; delivered at the first successful write. \emph{C2}: as in \code{software-test-01}.

\paragraph{\code{software-test-03}: constrained refactoring.} In a grade-book package, several functions repeat the same entry-parsing loop. The agent moves the parsing into one shared function, makes the listed functions use it while every public function keeps its behavior, runs the tests, and submits one change request. \emph{G}: protected files are unchanged and the structural requirements hold on the syntax tree. Retained functions must match the original program on hidden inputs in a differential test, and the shared parser must meet its contract. A mutation test must also confirm that the listed functions actually depend on it. \emph{C1}: a fourth function joins the list to be consolidated; delivered at the first successful write. \emph{C2}: as in \code{software-test-01}.

\subsection{Data analysis (technical; C0, C1)}
\paragraph{\code{data-test-01}: multi-table reconciliation.} Water samples must be counted and averaged by river basin. The site table is versioned, so a site that moved between basins has two rows, and units must be harmonized, missing values excluded, and unmatched samples listed. \emph{G}: the last submission has exactly the four non-empty basins, each with the exact member set, count, and mean within 0.01, and the exact unmatched set. The scorer recomputes from the raw rows. \emph{C1}: the site-assignment rule changes from the current version to the version effective at sampling time; delivered at the first query touching the site table.

\paragraph{\code{data-test-02}: cleaning under invariants.} A daily station log must be cleaned by five rules, and the result submitted as a change set. Only required edits are allowed: unusual but valid values are not errors, and protected rows stay unchanged. \emph{G}: applying the last change set to the original table yields exactly the expected rows and cells, and the change set itself has exactly the required entries with correct values and rule numbers. \emph{C1}: the duplicate rule keeps the latest entry instead of the earliest; delivered at the first query of the log.

\paragraph{\code{data-test-03}: statistical summary and chart.} Stream temperatures must be summarized by elevation band with a mean and a 95\% normal interval, and a bar chart with error bars specified whose marks equal the table values. \emph{G}: the method, the band set, the counts, and the estimates and bounds within 0.01 must be correct. The chart fields and units must be correct, with one mark per band matching the table. \emph{C1}: the band boundaries change, which keeps the labels but changes every value; delivered at the first query of the logger table.

\section{Evaluator Validation and Negative Controls}\label{app:verifier-validation}
This appendix details the checks behind the scorers summarized in Section~\ref{sec:software-validation}. It covers the checks completed before the freeze and the second scorer applied after collection.

\subsection{Scorer design}
Each scorer recomputes the expected outcome from the original task data and the final rules, never from the agent's statements and never by calling tool code. The data scorers recompute groups, means, and intervals from the raw rows. The code scorers run the public tests, hidden tests, and where specified a differential test against the original program in a restricted executor. One refactoring family adds a mutation test that replaces the shared parser to confirm that the listed functions really depend on it. Booking, order, ledger, and ticket scorers read the final state of the simulated system. Document scorers compare normalized paragraphs or fields with the expected text. The checks admit the solutions permitted by each task's predicates, including admissible itineraries, minimal-cost purchase combinations, valid sets of proof documents, and refactorings satisfying the declared behavioral and structural checks.

\subsection{Completed checks}
Table~\ref{tab:verifier-controls} lists the checks completed before the freeze. Table~\ref{tab:negative-counts} gives the number of negative controls per family.

\begin{table}[t]
\centering\small
\begin{tabularx}{\linewidth}{@{}lY@{}}\toprule
Check & Coverage and result\\\midrule
Reference solutions & Every family and applicable condition, with C2 in both worlds: $J=1$, no tool error, 10--18 tool calls.\\
Immediate finish & Every family and condition: calling \code{finish} first does not complete the task.\\
Negative controls & 147 scripted wrong solutions, each failing on its named endpoint.\\
C1 delivery & The correction is delivered once and stays visible in \code{read\_task}.\\
C2 equality & The ambiguous reply and all earlier observations are identical across worlds; the introduction does not reveal the world.\\
Tool interface & Unique tool names with descriptions; fuzzing with 21,760 ill-typed or extreme calls raised only validation errors.\\
Adversarial review & Four reviewers searched for over-strict scoring, loopholes, and condition defects; all findings fixed.\\
Pilot & A model outside the roster ran all 72 formal keys; no tool exception, unclassified termination, or undelivered correction.\\
Offline suite & 861 checks pass on Python 3.9, 3.13, and 3.14.\\\bottomrule
\end{tabularx}
\caption{Validation checks completed before the freeze, separate from the registered configurations. They establish task-code consistency and interface behavior; the second implementation adds a task-card-based reconstruction of the recorded scores.}\label{tab:verifier-controls}
\end{table}

\begin{table}[t]
\centering\small
\begin{tabular}{@{}lccc@{}}\toprule
Group & Family 01 & Family 02 & Family 03\\\midrule
travel & 7 & 7 & 6\\
personal & 5 & 5 & 5\\
inquiry & 4 & 5 & 4\\
service & 6 & 8 & 6\\
finance & 7 & 7 & 8\\
office & 6 & 5 & 5\\
software & 7 & 8 & 10\\
data & 6 & 5 & 5\\\midrule
Total & \multicolumn{3}{c}{147}\\\bottomrule
\end{tabular}
\caption{Negative controls per formal family. Every family has a control applying the superseded C1 information; every C2 family has a blind retry in the committed world and a finish without verification in the uncommitted world.}\label{tab:negative-counts}
\end{table}

\subsection{Second scorer}
The protocol assigned the second scorer to two operators, A for the \code{personal}, \code{inquiry}, \code{office}, and \code{data} groups and C for \code{travel}, \code{service}, \code{finance}, and \code{software}. The implemented procedure, including the changed implementation assignment, input access, and shared dependencies, is documented in Section~\ref{sec:verification-status}. Each family has its own scoring module. It recomputes the expected outcome from the task card and the world data, for example the admissible itineraries, the three-way invoice match, the clauses that apply to each expense report, or the group means of the raw rows. It then evaluates G, P, and E with their applicability on the final state and event log of every formal episode. The code scorers run the submitted workspace in the task's restricted executor, one temporary directory per episode, and compare its outputs with the scorer's own expected outputs. They add the structural checks, the differential test against the original program, and the mutation test that the task cards specify.

The comparison first targeted cases where a scorer may be too lenient or too strict. Examples are an answer that is right for a wrong reason, a stale value that happens to equal the final one, a correct state with a misreported total, a duplicate effect later reversed, an episode that finished without triggering the correction, and the two C2 worlds. Families with unusual outcome patterns in the collected data were examined with priority. This examination concerned scorer correctness only. A scorer was to be changed only where it disagreed with the task specification, never because an outcome looked surprising. Table~\ref{tab:second-scorer} gives the agreement, and Table~\ref{tab:disagreements} the resolution of every disagreement.

\begin{table}[t]
\centering\small
\begin{tabular}{@{}lrrrr@{}}\toprule
Comparison & G & P & E & J\\\midrule
First (code frozen before comparison) & 1,080 & 1,079 & 1,069 & 1,080\\
Final (after the logged corrections) & 1,080 & 1,080 & 1,077 & 1,080\\\bottomrule
\end{tabular}
\caption{Episodes, of 1,080, on which the second scorer and the original scorer agree. Agreement on P includes the 150 episodes in which both mark P as inapplicable. The three final E disagreements are the original scorer's lenient outcomes; the coordinator adopted the second scorer's outcome for them.}\label{tab:second-scorer}
\end{table}

\begin{table}[t]
\centering\footnotesize
\begin{tabularx}{\linewidth}{@{}lllY@{}}\toprule
Family & Episodes & Endpoint & Resolution\\\midrule
\code{office-test-01} & L05 C0 & P & Original correct. The agent changed a protected task and changed it back in the next step; the card counts any change, while the second scorer had compared only the final state.\\
\code{inquiry-test-02} & L12 C0 & E & Original correct. E judges each submitted verdict, and every submitted verdict cited its deciding source; the missing verdicts fail G, not E.\\
\code{finance-test-03} & seven C1 episodes & E & Original correct. The agents ordered at the superseded quantity, and their stated reasons for rejecting other quotes were true at that quantity; the second scorer had judged them at the final quantity.\\
\code{finance-test-03} & L05 C2 & E & Second scorer correct. The report labeled a cheaper compliant quote as more expensive.\\
\code{finance-test-02} & L08 C0, L08 C2 & E & Second scorer correct. An earlier effective decision on one report cited a clause that does not apply to it; the card requires the clauses of every effective decision to be correct.\\\bottomrule
\end{tabularx}
\caption{Resolution of the twelve disagreements of the first comparison. In the first three rows, nine outcomes in all, the second scorer was corrected and frozen again. The last two rows are the three adopted rescorings, applied in the analysis because the frozen scorer cannot change; each of these episodes already failed G or P, so G and J are unchanged.}\label{tab:disagreements}
\end{table}

The comparison establishes consistency on the collected records and documents every adjudicated disagreement. Its information-access and shared-implementation scope are specified in Section~\ref{sec:verification-status}.

\section{Analysis Details}
\label{app:analysis-details}

\subsection{Group profile and summary}
For configuration $m$ and an endpoint $Y\in\{G,P,E,J\}$, the group score in condition C0 is
\begin{equation}
 S_Y(m,g)=\frac{1}{|\mathcal F_g\cap\mathcal A_Y|}\sum_{f\in\mathcal F_g\cap\mathcal A_Y} Y_{mf\mathrm{C0}},
\end{equation}
where $\mathcal A_Y$ is the set of families for which $Y$ is defined. The summary $S_Y(m)$ averages the group scores over the groups in which $Y$ is defined, as in Equation~\eqref{eq:macro} for $G$. For $G$, $E$, and $J$, all twenty-four families are included. For $P$, the five families without a process constraint are excluded, so the \code{inquiry} group has no P score and the \code{data} group's P score rests on one family. The family-level outcomes are always shown with the summaries.

\subsection{Pooled condition counts}
For C1, let $n^{-}_1(m)$ be the number of families with $Y_{mf\mathrm{C0}}=1$ and $Y_{mf\mathrm{C1}}=0$, and $n^{+}_1(m)$ the number with the reverse. Then $\Delta_1(m)=(n^{+}_1(m)-n^{-}_1(m))/24$. For C2, the per-family difference $d_{mf}=\tfrac12(Y_{mf\mathrm{C2}0}+Y_{mf\mathrm{C2}1})-Y_{mf\mathrm{C0}}$ takes values in $\{-1,-\tfrac12,0,\tfrac12,1\}$. The report gives the count of families at each value. Pooled counts sum these over configurations. They are reported for all configurations and, as a sensitivity view, without L14. Because the same twenty-four families appear in every configuration, pooled counts are not independent across configurations, and they are not treated as a binomial sample.

\subsection{Bounds for unscorable episodes}
If an episode remains unscorable after its infrastructure rerun, its endpoint is unknown. The report then gives two summaries, one with the unknown endpoints set to 0 and one with them set to 1. For a contrast, the lower bound sets the condition episode to 0 and its C0 episode to 1, and the upper bound does the reverse. Known failures, including budget exhaustion and text-only termination, are never treated as unknown. In the collected data there is no unscorable episode, so both bounds equal the point estimate.

\subsection{Family resampling}
The resampling view draws, independently within each scenario group, three families with replacement from the group's three, and recomputes every summary and contrast with all of a family's episodes carried together. It uses 10,000 draws, and the 2.5th and 97.5th percentiles are reported as a range of family-mix sensitivity. With three families per group there are only ten distinct multisets per group, so the range is coarse and is not a confidence interval for a population of tasks. Leave-one-family-out summaries are reported beside it, recomputing each summary twenty-four times with one family removed.

\subsection{Diagnostic denominators}
\begin{itemize}
\item C1 delivery: all C1 episodes; numerator, episodes in which the correction notice was delivered.
\item C2 reply: all C2 episodes; numerator, episodes in which the designated operation was called and returned the ambiguous status.
\item C2 tool-name query proxy: C2 episodes with the ambiguous reply; numerator, episodes with a successful call to a designated verification-tool name after the reply and before another successful call of the designated operation's name. The proxy checks tool names and order, not whether the query concerns the affected object or returns its relevant state.
\item C2 process violation in the committed world: committed-world C2 episodes with the ambiguous reply; numerator, episodes with $P=0$. In every C2 family, P forbids the duplicate effect that a blind retry creates, so this rate bounds the duplicate rate from above; it also counts other process violations.
\item Text-only replies: episodes ending as \code{text\_reply\_no\_tool\_call}; the final reply is classified as an XML-like tool call, a bare JSON object, other text (which includes final summaries), or empty.
\end{itemize}
The C2 diagnostics are derived from the environment's event log, not from the agent's messages.

\subsection{Post-hoc operation-matched follow-up}\label{app:c2-operation-verification}
This exploratory reanalysis was added on 2026-11-28 after all formal scores had been observed. It reads the existing C2 event sequences and frozen tool contracts, without executing an agent, changing a score, or replacing the original assigned denominators. The trigger is the first successful call of the family's designated operation, which receives the C2 ambiguous reply. The follow-up window runs from that event to episode termination, retaining unrelated actions in the trace.

Operation identity follows the affected scope: for example a journey or timed-entry slot; invitation recipients on a reservation; refund lines on an order; invoice identifiers; an expense report; a supplier quote; or the software submission title together with the submission-time workspace. Overlapping object sets count as related. A successful state query must cover the original scope under its frozen handler, rather than merely have a familiar tool name. Complete listing tools can establish absence in the uncommitted world. Queries for disjoint objects and unrelated operations do not end the original operation's window. A successful related retry may alter payload values, so the classifier distinguishes related invocations from exact-request repeats. Neither category by itself establishes a duplicated effect or a process violation; those remain separately scored.

The first qualifying event determines a state-query-first or related-retry-first category. If neither occurs, the episode enters the neither-successful-follow-up category, including episodes that submit a result or contain rejected attempts. Validation-error calls do not qualify as successful queries or retries. An identity or ordering needed for classification that cannot be reconstructed yields an indeterminate outcome rather than an assumed absence. For software submissions, the workspace identity is reconstructed from immutable change-request snapshots and successful writes; an unrecoverable submission-time workspace is treated as unknown. Episodes without the trigger remain a separate category. These categories partition the assigned C2 records, and Table~\ref{tab:c2-operation-verification} reports them by hidden world.

The classification concerns observed tool-event order. Event records contain arguments and validation errors, while query coverage is established from the frozen handler contract. It does not assert response comprehension, verification of every payload value, or that a query result was received before a later action was chosen. Its descriptive comparisons are post-hoc and do not estimate the causal benefit of querying or retrying.

\section{Observation and Comparison Properties}\label{app:properties}
These properties connect task construction, observation boundaries, and aggregation. They follow from the declared contracts and serve as checks on the implementation. A property of the specification does not determine how an agent will perform.

\subsection{Identical assignments and different trajectories}
A task key contains family, instance, repeat, and condition, but not model or operator. The manifest attaches configuration and device. Matched configurations therefore receive identical intended inputs, corrections, and worlds. The equality concerns experimental obligations, not actions: reading a document again or taking a longer route is behavior within the common interface. The manifests of the three arms are identical in task keys, order, and worlds.

\subsection{Group balance}
With three families in every group and one episode per family and condition, the group-balanced C0 summary equals the plain mean over twenty-four families. This equality is a calculation check, not a reason to omit group profiles. It fails as soon as a family is unscorable or an endpoint is undefined for some families, as P is for five, and the group-balanced estimand is then computed directly.

\subsection{Non-applicability is not missing data}
Families without a state-changing operation have no C2 episode, and their cells are inapplicable. An assigned C2 episode that is not scorable is different: it stays in the assigned ledger and enters the bounds of Appendix~\ref{app:analysis-details}. Replacing either state by zero changes the estimand, for different reasons. Both terms of the C2 contrast are restricted to the same twelve families. As a calculation check, permuting outcomes of the other twelve families must leave the C2 contrast unchanged while the full C0 summary can change.

\subsection{Matched goals and investigation cost}
C0 and C1 have the same final goal, but C1 changes the preceding information. Rereading, recomputing, or revising can be part of successful adaptation, so equal obligations do not imply equal work, and correction-related calls are not subtracted. A C1 variant that added a requirement absent from C0 would compare different tasks. The offline tests therefore score the C1 reference solution with the same final scorer as C0 and require the superseded-information control to fail.

\subsection{Indistinguishable histories in C2}
Consider the two worlds of one C2 task, with the same request, the same final requirements, and the same complete history through the ambiguous reply. The operation took effect in one world and not in the other. If every observation up to that point is identical, no side channel reveals the world, and the agent's sampling is independent of the world, then the distribution of the agent's next action is the same in both worlds until an authorized observation distinguishes them. The seed depends only on the task key, and the world is assigned to the repeat through the private salt. Neither the seed nor the repeat number therefore reveals the world to an agent that does not know the salt. The two runs of a family use different seeds, so their trajectories can differ for reasons unrelated to the world, which is one reason the per-family C2 outcome is read together with the event log.

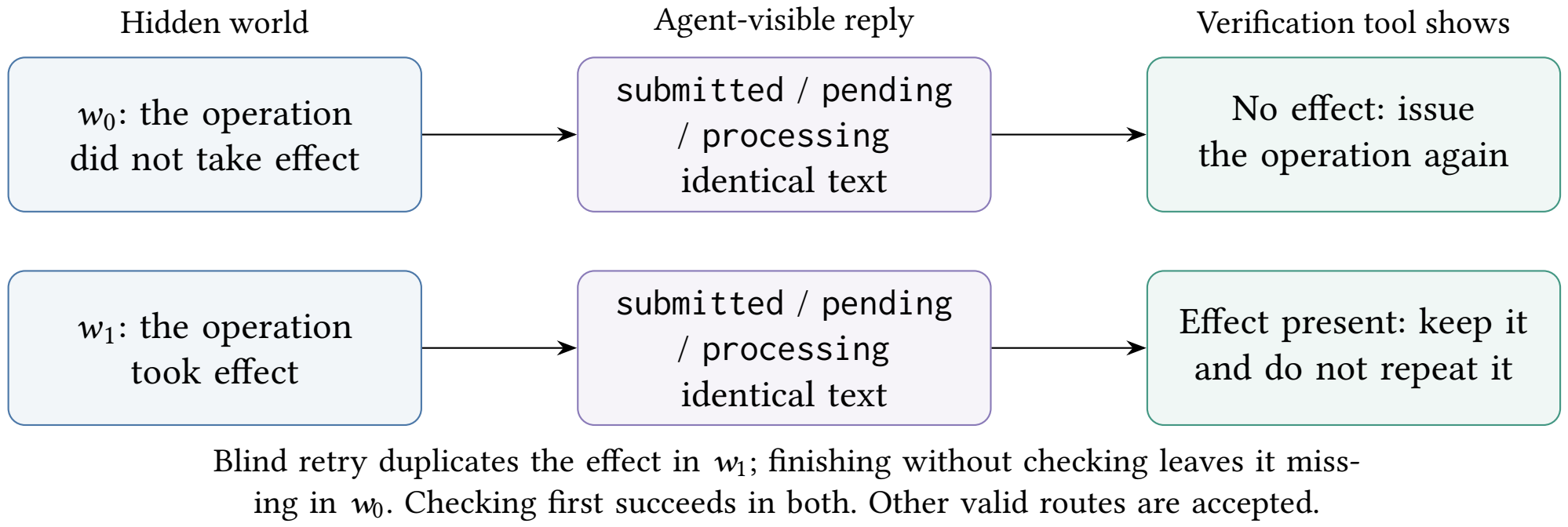
\begin{figure}[t]
\centering\begin{tikzpicture}[x=\linewidth,y=1cm,font=\small,>=Stealth,
 box/.style={draw,rounded corners,align=center,text width=.23\linewidth,minimum height=1.3cm,inner sep=4pt}]
\node[font=\footnotesize,align=center] at (.155,.95) {Hidden world};
\node[font=\footnotesize,align=center] at (.5,.95) {Agent-visible reply};
\node[font=\footnotesize,align=center] at (.845,.95) {Verification tool shows};
\node[box,draw=vizP1,fill=vizP1!7] (s0) at (.155,0) {$w_0$: the operation\\did not take effect};
\node[box,draw=vizC1,fill=vizC1!8] (o0) at (.5,0) {\code{submitted} / \code{pending}\\/ \code{processing}\\identical text};
\node[box,draw=vizC2,fill=vizC2!8] (q0) at (.845,0) {No effect: issue\\the operation again};
\node[box,draw=vizP1,fill=vizP1!7] (s1) at (.155,-1.8) {$w_1$: the operation\\took effect};
\node[box,draw=vizC1,fill=vizC1!8] (o1) at (.5,-1.8) {\code{submitted} / \code{pending}\\/ \code{processing}\\identical text};
\node[box,draw=vizC2,fill=vizC2!8] (q1) at (.845,-1.8) {Effect present: keep it\\and do not repeat it};
\draw[->] (s0.east)--(o0.west);\draw[->] (s1.east)--(o1.west);
\draw[->] (o0.east)--(q0.west);\draw[->] (o1.east)--(q1.west);
\node[align=center,font=\footnotesize,text width=.92\linewidth] at (.5,-2.95) {Blind retry duplicates the effect in $w_1$; finishing without checking leaves it missing in $w_0$. Checking first succeeds in both. Other valid routes are accepted.};
\end{tikzpicture}
\Description{Two hidden worlds differ only in whether the first call of the designated operation took effect. The agent receives an identical ambiguous status in both. A verification tool reveals the true state: in the first world the operation must be issued again; in the second the existing effect must be kept without repetition. Blind retry fails in one world and finishing without checking fails in the other.}
\caption{Contract for state-changing families. The same ambiguous reply corresponds to different hidden effects. Checking the state with the verification tool distinguishes them; blind retry fails in the committed world and finishing without checking fails in the other.}\label{fig:uncertain-worlds}
\end{figure}

This argument identifies the leakage paths that the construction excludes: an identifier present in only one world, a reply that differs in wording, length, or timing by world, a readable assignment file, or a seed derived from the world. Comparing reply strings alone would miss differences elsewhere in the history, so the offline test compares the complete serialized observation histories. Figure~\ref{fig:uncertain-worlds} shows the contract.

\subsection{Why both worlds are needed}
If only one world were used, a fixed strategy could succeed without any verification: always retrying succeeds when the first call never takes effect, and never retrying succeeds when it always does. Running each C2 task once in each world makes both fixed strategies fail in exactly one of the two runs. A strategy that checks before acting succeeds in both. The world-averaged C2 outcome of a family is therefore 1 only for a configuration that handles both worlds, $\tfrac12$ for one that handles exactly one, and 0 otherwise.

\section{Execution Architecture and Evidence Lifecycle}\label{app:architecture}
This appendix describes how an episode is executed and how its evidence is kept. All three operators use the same frozen kit and the same commands; only the operator letter differs.

\subsection{Pipeline}
\begin{enumerate}
\item \textbf{Manifest.} The study tool generates the arm's manifest from the frozen allocation: the five models of the device, the 72 task keys for each, their order, and the C2 worlds derived from the private salt.
\item \textbf{Preflight.} Before the first episode, the batch runner checks the runner and task-code fingerprints against the lock, the SDK, LiteLLM, and Ollama versions, the model digests, the frozen budget, and the salt hash, and refuses to start on any mismatch.
\item \textbf{Inference server.} The runner starts its own Ollama 0.40.14 server on a separate port, with the frozen context length, and verifies the server's reported context and model memory for every model.
\item \textbf{Metering proxy.} A local proxy between the SDK and the server records every request and response, counts requests and output tokens, refuses a request that would exceed a budget, and stops the batch if any request carries a reasoning parameter.
\item \textbf{Episode process.} Each episode runs in its own process. It builds a fresh task environment for the family, condition, and world, registers the family's tools as SDK client tools, creates an OpenHands \code{Agent} with only those tools and no default tools, and runs one \code{Conversation} in an empty temporary workspace. The episode time limit is enforced by a timer in the process and, as a backstop, by the batch runner.
\item \textbf{Scoring.} When the conversation ends, the process writes the environment record, scores it, writes the trace, and finally writes the summary atomically. A scorer exception is recorded as a task-code defect rather than a model outcome.
\item \textbf{Classification.} The batch runner assigns the terminal reason from the summary and the request log, reruns an infrastructure termination once with the same seed, and appends the result line.
\end{enumerate}

\subsection{Isolation}
The agent can act only through the family's tools. The SDK's default tools are disabled, so there is no terminal, file editor, or browser. The temporary workspace is empty and removed after the episode. The environment object holds the hidden state and the scorer data in the episode process, outside anything a tool returns. Model-written code in the software families runs in a restricted executor in a separate process with CPU and wall-clock limits. No information passes between episodes: every episode has a fresh session, environment, and workspace. The inference server may keep a model loaded between episodes, but it keeps no conversation state.

\subsection{Interruption and resumption}
A batch can be interrupted at any time. Rerunning the same command resumes at the first episode without a summary. An episode without a summary is rerun from the beginning with the same seed, and any partial request log is kept under a separate name. Resumption is refused if the code, dependencies, or environment differ from those recorded at the start. Formal batches run one episode at a time; concurrency is allowed only in development stages.

\subsection{Evidence lifecycle}
At the end of an arm, the operator archives the arm's output directory and sends it to the coordinator with its hash. The coordinator unpacks each archive into a fresh directory, runs the frozen report tool over it, inspects every episode record, and writes a check record with the archive hashes and findings. The salt file never enters a package or the repository; only its hash is recorded. Development outputs, the pilot, and the formal arms are stored separately, and a formal record is never relabelled as development data or overwritten.

\section{Fixed Transcription Contract}\label{app:perception}
One formal family, \code{inquiry-test-03}, works with image content: a store receipt, a photo of a thermostat display, an app settings screenshot, two manual pages, and an unrelated receipt as a distractor. This appendix states how that content reaches the agent and what the family can and cannot measure.

\subsection{Interface}
The agent lists the available images with \code{list\_images} and requests one region of one image with \code{view\_image}. Examples of regions are the receipt's header, item lines, totals, and footer, the display's screen and status icons, the app's mode and system panels, and the manual's error table, pairing steps, and eco requirements. The tool returns a fixed, pre-written English transcription of that region. No image file is delivered and no vision model or OCR runs. The transcription is the same for every model, every condition, and every call, except for the C1 replacement described below. A text document, the return policy, is read with a separate tool. Answers cite their sources as an image region or a document identifier.

\subsection{What the transcription contains}
The transcriptions contain what a careful reader would see in each region: item names and prices, a discount line, dates, a blinking error code, panel settings, and manual steps. They do not contain derived answers. The agent must still compute the amount actually paid after a discount, apply the return windows and a membership extension to the purchase date, match the displayed code to the right manual step while avoiding a similar distractor code, and determine which settings must change. Wording can still carry cues that an image would not. The display transcription, for instance, names the code explicitly. This is recorded as a known limitation.

\subsection{C1 in this family}
Initially, the screenshot labeled as the living-room thermostat actually shows a different device. At the first view of any region of that screenshot, the result still carries the old transcription together with a notice that the screenshot was wrong and has been replaced, and later views return the transcription of the correct screenshot. Answers based on the old transcription fail G. Citations of the old version do not count as reading the current one, so they also fail E. The old screenshot's header names a different room, so a careful agent can notice the mismatch before the notice. This does not affect scoring.

\subsection{Scope of claims}
The family measures the use of image-derived information under a fixed transcription: reading, combining, computing, and citing. It does not measure perception, native multimodal ability, or robustness to imperfect recognition, and no result of this family is reported as a visual capability. A study of native perception would need actual images, a declared perception path for each configuration, and its own comparison design.

\clearpage
\section{Additional Results}\label{app:results}
\begin{figure}[tbp]
\centering
\begin{tikzpicture}[x=1cm,y=1cm,line cap=butt]
\draw[vizgrid] (0,-6) -- (0,0.05);
\node[viztick,anchor=north] at (0,-6.08) {0};
\draw[vizgrid] (0.975,-6) -- (0.975,0.05);
\node[viztick,anchor=north] at (0.975,-6.08) {50};
\draw[vizgrid] (1.95,-6) -- (1.95,0.05);
\node[viztick,anchor=north] at (1.95,-6.08) {100};
\node[viztick,anchor=south,text width=2.55cm,align=center] at (0.975,0.08) {C1 correction delivered};
\filldraw[fill=vizP2,draw=white,line width=0.5pt] (1.95,-0.2) circle (0.06);
\filldraw[fill=vizP2,draw=white,line width=0.5pt] (1.95,-0.6) circle (0.06);
\filldraw[fill=vizP1,draw=white,line width=0.5pt] (1.95,-1) circle (0.06);
\filldraw[fill=vizP1,draw=white,line width=0.5pt] (1.95,-1.4) circle (0.06);
\filldraw[fill=vizP1,draw=white,line width=0.5pt] (1.95,-1.8) circle (0.06);
\filldraw[fill=vizP2,draw=white,line width=0.5pt] (1.95,-2.2) circle (0.06);
\filldraw[fill=vizP3,draw=white,line width=0.5pt] (1.95,-2.6) circle (0.06);
\filldraw[fill=vizP3,draw=white,line width=0.5pt] (1.787,-3) circle (0.06);
\filldraw[fill=vizP2,draw=white,line width=0.5pt] (1.869,-3.4) circle (0.06);
\filldraw[fill=vizP2,draw=white,line width=0.5pt] (1.869,-3.8) circle (0.06);
\filldraw[fill=vizP3,draw=white,line width=0.5pt] (1.625,-4.2) circle (0.06);
\filldraw[fill=vizP1,draw=white,line width=0.5pt] (1.95,-4.6) circle (0.06);
\filldraw[fill=vizP1,draw=white,line width=0.5pt] (1.95,-5) circle (0.06);
\filldraw[fill=vizP3,draw=white,line width=0.5pt] (0.65,-5.4) circle (0.06);
\filldraw[fill=vizP3,draw=white,line width=0.5pt] (1.381,-5.8) circle (0.06);
\draw[vizgrid] (2.6,-6) -- (2.6,0.05);
\node[viztick,anchor=north] at (2.6,-6.08) {0};
\draw[vizgrid] (3.575,-6) -- (3.575,0.05);
\node[viztick,anchor=north] at (3.575,-6.08) {50};
\draw[vizgrid] (4.55,-6) -- (4.55,0.05);
\node[viztick,anchor=north] at (4.55,-6.08) {100};
\node[viztick,anchor=south,text width=2.55cm,align=center] at (3.575,0.08) {C2 ambiguous reply reached};
\filldraw[fill=vizP2,draw=white,line width=0.5pt] (4.55,-0.2) circle (0.06);
\filldraw[fill=vizP2,draw=white,line width=0.5pt] (4.55,-0.6) circle (0.06);
\filldraw[fill=vizP1,draw=white,line width=0.5pt] (4.55,-1) circle (0.06);
\filldraw[fill=vizP1,draw=white,line width=0.5pt] (4.55,-1.4) circle (0.06);
\filldraw[fill=vizP1,draw=white,line width=0.5pt] (4.388,-1.8) circle (0.06);
\filldraw[fill=vizP2,draw=white,line width=0.5pt] (4.55,-2.2) circle (0.06);
\filldraw[fill=vizP3,draw=white,line width=0.5pt] (4.469,-2.6) circle (0.06);
\filldraw[fill=vizP3,draw=white,line width=0.5pt] (4.306,-3) circle (0.06);
\filldraw[fill=vizP2,draw=white,line width=0.5pt] (4.388,-3.4) circle (0.06);
\filldraw[fill=vizP2,draw=white,line width=0.5pt] (4.306,-3.8) circle (0.06);
\filldraw[fill=vizP3,draw=white,line width=0.5pt] (4.469,-4.2) circle (0.06);
\filldraw[fill=vizP1,draw=white,line width=0.5pt] (4.55,-4.6) circle (0.06);
\filldraw[fill=vizP1,draw=white,line width=0.5pt] (4.144,-5) circle (0.06);
\filldraw[fill=vizP3,draw=white,line width=0.5pt] (2.6,-5.4) circle (0.06);
\filldraw[fill=vizP3,draw=white,line width=0.5pt] (3.819,-5.8) circle (0.06);
\draw[vizgrid] (5.2,-6) -- (5.2,0.05);
\node[viztick,anchor=north] at (5.2,-6.08) {0};
\draw[vizgrid] (6.175,-6) -- (6.175,0.05);
\node[viztick,anchor=north] at (6.175,-6.08) {50};
\draw[vizgrid] (7.15,-6) -- (7.15,0.05);
\node[viztick,anchor=north] at (7.15,-6.08) {100};
\node[viztick,anchor=south,text width=2.55cm,align=center] at (6.175,0.08) {tool query proxy};
\filldraw[fill=vizP2,draw=white,line width=0.5pt] (5.444,-0.2) circle (0.06);
\filldraw[fill=vizP2,draw=white,line width=0.5pt] (5.688,-0.6) circle (0.06);
\filldraw[fill=vizP1,draw=white,line width=0.5pt] (5.85,-1) circle (0.06);
\filldraw[fill=vizP1,draw=white,line width=0.5pt] (5.281,-1.4) circle (0.06);
\filldraw[fill=vizP1,draw=white,line width=0.5pt] (5.732,-1.8) circle (0.06);
\filldraw[fill=vizP2,draw=white,line width=0.5pt] (5.688,-2.2) circle (0.06);
\filldraw[fill=vizP3,draw=white,line width=0.5pt] (5.963,-2.6) circle (0.06);
\filldraw[fill=vizP3,draw=white,line width=0.5pt] (5.571,-3) circle (0.06);
\filldraw[fill=vizP2,draw=white,line width=0.5pt] (5.643,-3.4) circle (0.06);
\filldraw[fill=vizP2,draw=white,line width=0.5pt] (5.479,-3.8) circle (0.06);
\filldraw[fill=vizP3,draw=white,line width=0.5pt] (5.37,-4.2) circle (0.06);
\filldraw[fill=vizP1,draw=white,line width=0.5pt] (5.362,-4.6) circle (0.06);
\filldraw[fill=vizP1,draw=white,line width=0.5pt] (5.713,-5) circle (0.06);
\node[viztick,anchor=west] at (5.25,-5.4) {n/a};
\filldraw[fill=vizP3,draw=white,line width=0.5pt] (5.2,-5.8) circle (0.06);
\draw[vizgrid] (7.8,-6) -- (7.8,0.05);
\node[viztick,anchor=north] at (7.8,-6.08) {0};
\draw[vizgrid] (8.775,-6) -- (8.775,0.05);
\node[viztick,anchor=north] at (8.775,-6.08) {50};
\draw[vizgrid] (9.75,-6) -- (9.75,0.05);
\node[viztick,anchor=north] at (9.75,-6.08) {100};
\node[viztick,anchor=south,text width=2.55cm,align=center] at (8.775,0.08) {committed world: P violated};
\filldraw[fill=vizP2,draw=white,line width=0.5pt] (7.8,-0.2) circle (0.06);
\filldraw[fill=vizP2,draw=white,line width=0.5pt] (7.963,-0.6) circle (0.06);
\filldraw[fill=vizP1,draw=white,line width=0.5pt] (7.8,-1) circle (0.06);
\filldraw[fill=vizP1,draw=white,line width=0.5pt] (7.8,-1.4) circle (0.06);
\filldraw[fill=vizP1,draw=white,line width=0.5pt] (7.8,-1.8) circle (0.06);
\filldraw[fill=vizP2,draw=white,line width=0.5pt] (8.125,-2.2) circle (0.06);
\filldraw[fill=vizP3,draw=white,line width=0.5pt] (8.125,-2.6) circle (0.06);
\filldraw[fill=vizP3,draw=white,line width=0.5pt] (8.155,-3) circle (0.06);
\filldraw[fill=vizP2,draw=white,line width=0.5pt] (7.963,-3.4) circle (0.06);
\filldraw[fill=vizP2,draw=white,line width=0.5pt] (8.19,-3.8) circle (0.06);
\filldraw[fill=vizP3,draw=white,line width=0.5pt] (7.8,-4.2) circle (0.06);
\filldraw[fill=vizP1,draw=white,line width=0.5pt] (7.963,-4.6) circle (0.06);
\filldraw[fill=vizP1,draw=white,line width=0.5pt] (8.017,-5) circle (0.06);
\node[viztick,anchor=west] at (7.85,-5.4) {n/a};
\filldraw[fill=vizP3,draw=white,line width=0.5pt] (8.883,-5.8) circle (0.06);
\node[viztick,anchor=north] at (4.875,-6.42) {percent of each panel's denominator};
\node[vizlab,anchor=east] at (-0.12,-0.2) {L09 qwen3.6 27b-coding-mtp};
\node[vizlab,anchor=east] at (-0.12,-0.6) {L08 gemma4 31b};
\node[vizlab,anchor=east] at (-0.12,-1) {L01 qwen3.6 35b-a3b};
\node[vizlab,anchor=east] at (-0.12,-1.4) {L03 granite4.2 30b};
\node[vizlab,anchor=east] at (-0.12,-1.8) {L02 gemma4 26b-a4b};
\node[vizlab,anchor=east] at (-0.12,-2.2) {L07 nemotron-3.5-lightning 30b-a3b};
\node[vizlab,anchor=east] at (-0.12,-2.6) {L11 qwen3.5 9b};
\node[vizlab,anchor=east] at (-0.12,-3) {L13 granite4.2 8b};
\node[vizlab,anchor=east] at (-0.12,-3.4) {L10 gemma4 12b};
\node[vizlab,anchor=east] at (-0.12,-3.8) {L06 glm-4.7-flash};
\node[vizlab,anchor=east] at (-0.12,-4.2) {L12 gemma4 e4b};
\node[vizlab,anchor=east] at (-0.12,-4.6) {L04 lfm2.5 24b-a2b};
\node[vizlab,anchor=east] at (-0.12,-5) {L05 north-mini-code-1.0};
\node[vizlab,anchor=east] at (-0.12,-5.4) {L14 nemotron-3-nano 4b};
\node[vizlab,anchor=east] at (-0.12,-5.8) {L15 lfm2.5 8b-a1b};
\node[fill=vizP1,minimum width=0.22cm,minimum height=0.15cm,inner sep=0,anchor=west] (lgs1) at (0,-6.95) {};
\node[viztick,anchor=west,xshift=0.06cm] (lgl1) at (lgs1.east) {Apple M4 Max};
\node[fill=vizP2,minimum width=0.22cm,minimum height=0.15cm,inner sep=0,anchor=west,xshift=0.3cm] (lgs2) at (lgl1.east) {};
\node[viztick,anchor=west,xshift=0.06cm] (lgl2) at (lgs2.east) {Apple M5 Max};
\node[fill=vizP3,minimum width=0.22cm,minimum height=0.15cm,inner sep=0,anchor=west,xshift=0.3cm] (lgs3) at (lgl2.east) {};
\node[viztick,anchor=west,xshift=0.06cm] (lgl3) at (lgs3.east) {RTX 3090};
\end{tikzpicture}
\caption{Trigger and execution diagnostics, each over its own denominator: C1 episodes with the correction delivered; C2 episodes that reached the ambiguous reply; the original tool-name query proxy among triggered episodes; and committed-world episodes with the reply that violated P. The query proxy checks tool names and ordering, without matching the operation's object or the returned state. Exact counts are in Table~\ref{tab:diagnostics}.}\label{fig:diagnostics}
\end{figure}
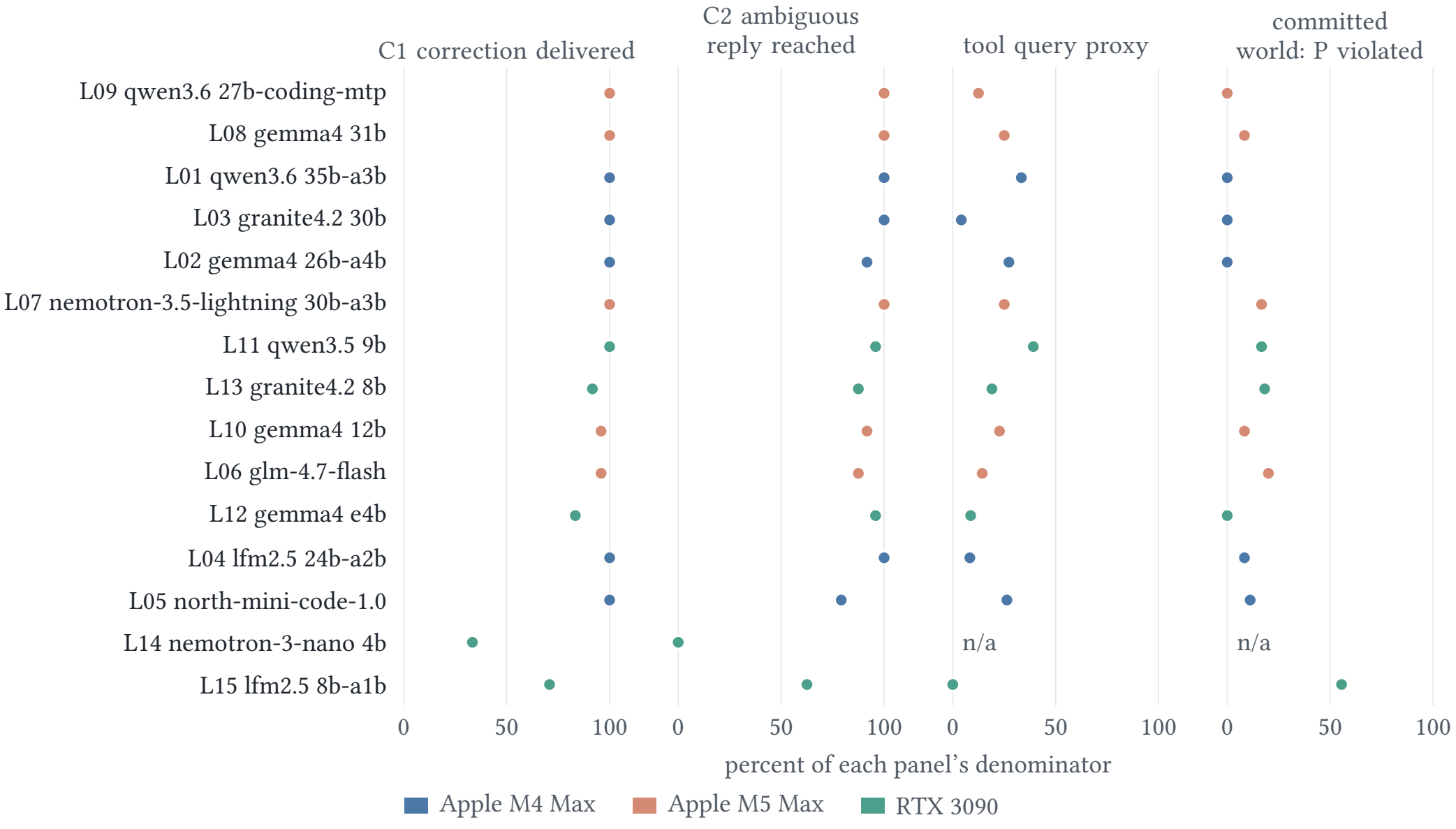

This appendix gives the exact tables behind Section~\ref{sec:results} and the figures not shown there. All of them are generated from the same scored records, after the second-scorer rescoring, and follow the definitions of Section~\ref{sec:analysis} and Appendix~\ref{app:analysis-details}. The original C2 tool-name proxy remains in Table~\ref{tab:diagnostics} and Figure~\ref{fig:diagnostics}; the separate post-hoc operation-matched classification is in Table~\ref{tab:c2-operation-verification}. Time appears only within a platform. Figures that relate completion to resource use are exploratory.

\begin{table}[tbp]
\centering\footnotesize

\begin{tabular}{@{}lrrrrrrrrr@{}}\toprule
& \multicolumn{4}{c}{C0 summary} & \multicolumn{5}{c}{All conditions, records with G, P, E defined}\\
\cmidrule(lr){2-5}\cmidrule(l){6-10}
Config. & $S_G$ & $S_P$ & $S_E$ & $S_J$ & Records & J & G, not P & G, P, not E & not G\\\midrule
L01 & 0.71 & 1.00 & 0.96 & 0.71 & 62 & 37 & 0 & 1 & 24\\
L02 & 0.46 & 0.95 & 0.71 & 0.42 & 62 & 25 & 0 & 2 & 35\\
L03 & 0.71 & 1.00 & 0.88 & 0.71 & 62 & 35 & 0 & 0 & 27\\
L04 & 0.04 & 0.90 & 0.62 & 0.04 & 62 & 5 & 0 & 0 & 57\\
L05 & 0.00 & 0.71 & 0.38 & 0.00 & 62 & 1 & 0 & 0 & 61\\
L06 & 0.17 & 0.81 & 0.67 & 0.12 & 62 & 6 & 1 & 0 & 55\\
L07 & 0.42 & 0.95 & 0.75 & 0.42 & 62 & 22 & 0 & 1 & 39\\
L08 & 0.92 & 0.95 & 0.92 & 0.88 & 62 & 46 & 2 & 0 & 14\\
L09 & 0.92 & 1.00 & 0.96 & 0.92 & 62 & 41 & 0 & 0 & 21\\
L10 & 0.29 & 1.00 & 0.58 & 0.21 & 62 & 14 & 0 & 2 & 46\\
L11 & 0.33 & 0.95 & 0.58 & 0.33 & 62 & 13 & 0 & 1 & 48\\
L12 & 0.12 & 1.00 & 0.62 & 0.08 & 62 & 4 & 0 & 1 & 57\\
L13 & 0.29 & 0.90 & 0.58 & 0.29 & 62 & 16 & 0 & 0 & 46\\
L14 & 0.00 & 1.00 & 0.00 & 0.00 & 62 & 0 & 0 & 0 & 62\\
L15 & 0.00 & 0.76 & 0.12 & 0.00 & 62 & 0 & 0 & 0 & 62\\
\midrule Total & & & & & 930 & 265 & 3 & 8 & 654\\
\bottomrule\end{tabular}

\caption{C0 summaries of the four endpoints and the outcome partition over all conditions. $S_P$ averages over the groups in which P is defined, so the \code{inquiry} group is excluded and the \code{data} group rests on one family. The partition covers the 62 records per configuration on which G, P, and E are all defined: joint success, goal reached with a process violation, goal reached with conforming process but insufficient evidence, and goal not reached.}\label{tab:endpoints}
\end{table}

\begin{table}[tbp]
\centering\footnotesize

\begin{tabular}{@{}lrrrrrr@{}}\toprule
Config. & C1 worse/same/better & $\Delta_1$ & C0 on 12 & C2 on 12 & $\Delta_2$ & C2 $d=-1/-\frac12/0/\frac12/1$\\\midrule
L01 & 8/11/5 & -0.125 & 0.58 & 0.50 & -0.083 & 0/6/3/2/1\\
L02 & 5/18/1 & -0.167 & 0.58 & 0.42 & -0.167 & 1/5/3/3/0\\
L03 & 6/17/1 & -0.208 & 0.67 & 0.42 & -0.250 & 2/6/1/2/1\\
L04 & 0/23/1 & 0.042 & 0.08 & 0.08 & 0.000 & 1/0/9/2/0\\
L05 & 0/23/1 & 0.042 & 0.00 & 0.00 & 0.000 & 0/0/12/0/0\\
L06 & 4/17/3 & -0.042 & 0.17 & 0.04 & -0.125 & 1/1/10/0/0\\
L07 & 5/18/1 & -0.167 & 0.58 & 0.33 & -0.250 & 1/4/7/0/0\\
L08 & 3/20/1 & -0.083 & 0.92 & 0.50 & -0.417 & 1/8/3/0/0\\
L09 & 4/20/0 & -0.167 & 0.92 & 0.46 & -0.458 & 2/7/3/0/0\\
L10 & 4/17/3 & -0.042 & 0.33 & 0.21 & -0.125 & 1/3/6/2/0\\
L11 & 6/16/2 & -0.167 & 0.42 & 0.12 & -0.292 & 3/2/6/1/0\\
L12 & 3/21/0 & -0.125 & 0.17 & 0.08 & -0.083 & 0/2/10/0/0\\
L13 & 4/19/1 & -0.125 & 0.42 & 0.21 & -0.208 & 2/3/5/2/0\\
L14 & 0/24/0 & 0.000 & 0.00 & 0.00 & 0.000 & 0/0/12/0/0\\
L15 & 0/24/0 & 0.000 & 0.00 & 0.00 & 0.000 & 0/0/12/0/0\\
\midrule Pooled (all) & 52/288/20 & -0.089 & & & -0.164 & 15/47/102/14/2\\
\midrule Pooled (without L14) & 52/264/20 & -0.095 & & & -0.176 & 15/47/90/14/2\\
\bottomrule\end{tabular}

\caption{Condition contrasts per configuration and pooled. C1: tasks worsened, unchanged, and improved relative to C0, of 24. C0 on 12 and C2 on 12: goal completion on the twelve state-changing tasks in C0 and averaged over the two C2 worlds. The last column counts the twelve per-task differences $d$ at each value. The pooled rows sum the counts and average the contrasts; the second pooled row excludes L14.}\label{tab:contrasts}
\end{table}

\begin{table}[tbp]
\centering\footnotesize

\begin{tabular}{@{}lrrrrrrr@{}}\toprule
& C1 & \multicolumn{3}{c}{C2 diagnostics} & \multicolumn{3}{c}{G on the 12 C2 tasks (exploratory)}\\
\cmidrule(lr){2-2}\cmidrule(lr){3-5}\cmidrule(l){6-8}
Config. & Delivered & Reply & Tool query & P violated & C0 & Committed & Not committed\\\midrule
L01 & 24/24 & 24/24 & 8/24 & 0/12 & 7/12 & 10/12 & 2/12\\
L02 & 24/24 & 22/24 & 6/22 & 0/11 & 7/12 & 8/12 & 2/12\\
L03 & 24/24 & 24/24 & 1/24 & 0/12 & 8/12 & 9/12 & 1/12\\
L04 & 24/24 & 24/24 & 2/24 & 1/12 & 1/12 & 1/12 & 1/12\\
L05 & 24/24 & 19/24 & 5/19 & 1/9 & 0/12 & 0/12 & 0/12\\
L06 & 23/24 & 21/24 & 3/21 & 2/10 & 2/12 & 1/12 & 0/12\\
L07 & 24/24 & 24/24 & 6/24 & 2/12 & 7/12 & 6/12 & 2/12\\
L08 & 24/24 & 24/24 & 6/24 & 1/12 & 11/12 & 10/12 & 2/12\\
L09 & 24/24 & 24/24 & 3/24 & 0/12 & 11/12 & 9/12 & 2/12\\
L10 & 23/24 & 22/24 & 5/22 & 1/12 & 4/12 & 5/12 & 0/12\\
L11 & 24/24 & 23/24 & 9/23 & 2/12 & 5/12 & 3/12 & 0/12\\
L12 & 20/24 & 23/24 & 2/23 & 0/11 & 2/12 & 2/12 & 0/12\\
L13 & 22/24 & 21/24 & 4/21 & 2/11 & 5/12 & 3/12 & 2/12\\
L14 & 8/24 & 0/24 & 0/0 & 0/0 & 0/12 & 0/12 & 0/12\\
L15 & 17/24 & 15/24 & 0/15 & 5/9 & 0/12 & 0/12 & 0/12\\
\midrule Total & 329/360 & 310/360 & 60/310 & 17/157 & 70/180 & 67/180 & 14/180\\
\bottomrule\end{tabular}

\caption{Trigger diagnostics, each over its own denominator (Appendix~\ref{app:analysis-details}). Delivered: C1 episodes in which the correction was delivered. Reply: C2 episodes in which the designated operation returned the ambiguous status. Tool query: of those, episodes with a successful designated verification-tool call before another successful call of the designated operation's name. This proxy does not check the queried object or returned state. P violated: committed-world episodes with the reply that violated a process constraint, an upper bound on duplicated effects. The last three columns are exploratory: goal completion on the twelve state-changing tasks in C0 and in each hidden world of C2.}\label{tab:diagnostics}
\end{table}

\begin{table}[tbp]
\centering\footnotesize

\begin{tabular}{@{}lrrrrrrr@{}}\toprule
Config. & Submitted & Text only & Stuck & Output limit & Request limit & Truncated & Parse error\\\midrule
L01 & 70 & 1 & 0 & 0 & 0 & 0 & 1\\
L02 & 66 & 0 & 4 & 2 & 0 & 0 & 0\\
L03 & 69 & 0 & 2 & 0 & 0 & 0 & 1\\
L04 & 55 & 15 & 1 & 1 & 0 & 0 & 0\\
L05 & 16 & 53 & 1 & 0 & 1 & 1 & 0\\
L06 & 66 & 0 & 1 & 1 & 2 & 1 & 1\\
L07 & 67 & 1 & 0 & 0 & 0 & 3 & 1\\
L08 & 69 & 0 & 2 & 1 & 0 & 0 & 0\\
L09 & 70 & 1 & 0 & 0 & 0 & 0 & 1\\
L10 & 57 & 1 & 6 & 0 & 2 & 6 & 0\\
L11 & 59 & 0 & 2 & 7 & 4 & 0 & 0\\
L12 & 69 & 0 & 2 & 1 & 0 & 0 & 0\\
L13 & 51 & 0 & 10 & 10 & 0 & 0 & 1\\
L14 & 3 & 64 & 0 & 0 & 0 & 0 & 5\\
L15 & 36 & 29 & 4 & 0 & 2 & 1 & 0\\
\midrule Total & 823 & 165 & 35 & 23 & 11 & 12 & 11\\
\bottomrule\end{tabular}

\caption{Terminal reasons over the 72 episodes of each configuration. The frozen reason codes are \code{submitted}, \code{text\_reply\_no\_tool\_call}, \code{stuck\_detected}, \code{budget\_episode\_output}, \code{budget\_requests}, \code{output\_truncated\_no\_action}, and \code{tool\_call\_parse\_error}. No episode ended on a time limit or for an infrastructure reason.}\label{tab:terminal}
\end{table}

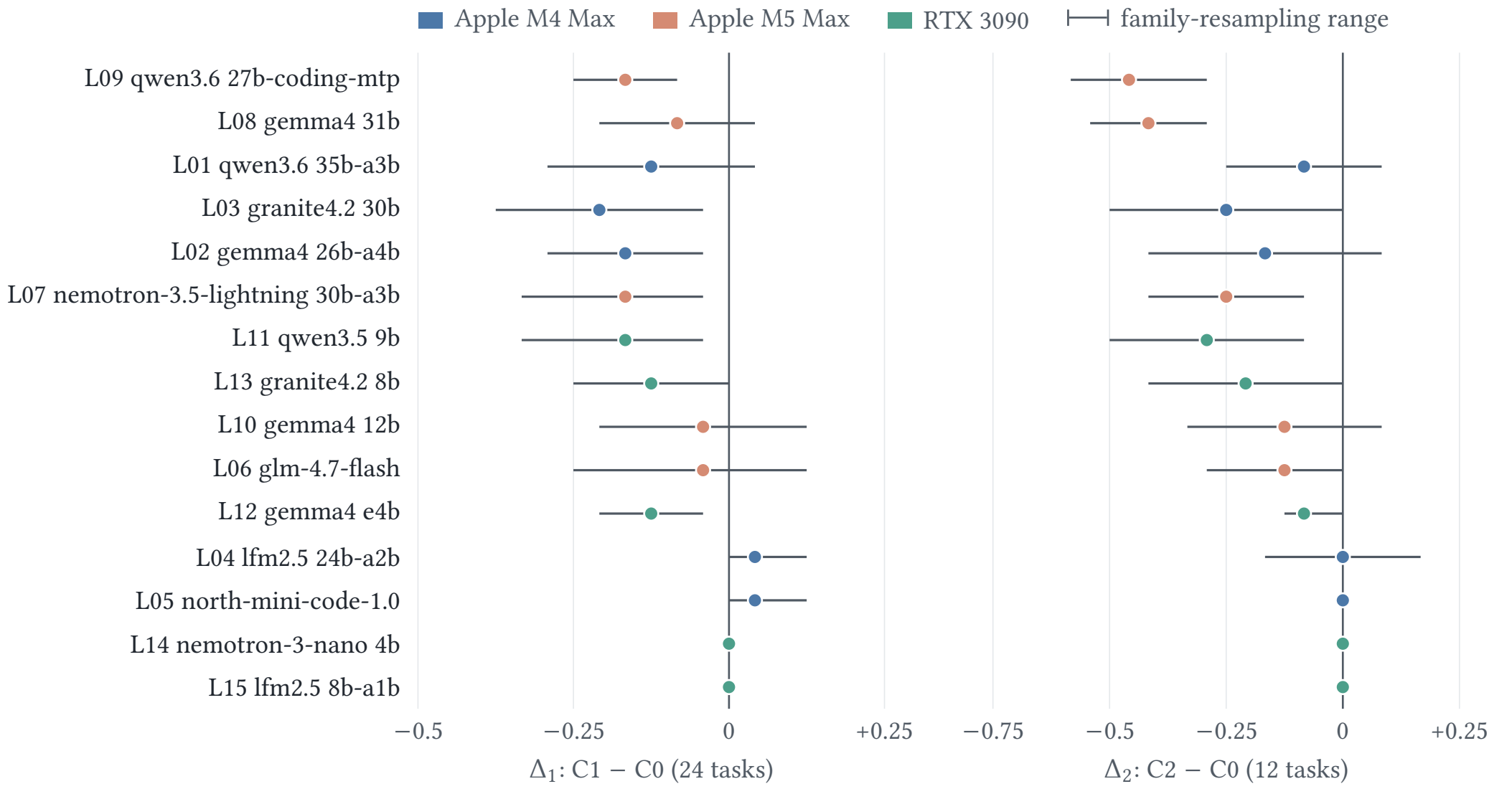
\begin{figure}[tbp]
\centering
\begin{tikzpicture}[x=1cm,y=1cm,line cap=butt]
\draw[vizgrid] (0,-6) -- (0,0.05);
\node[viztick,anchor=north] at (0,-6.08) {$-$0.5};
\draw[vizgrid] (1.433,-6) -- (1.433,0.05);
\node[viztick,anchor=north] at (1.433,-6.08) {$-$0.25};
\draw[vizgrid] (2.867,-6) -- (2.867,0.05);
\node[viztick,anchor=north] at (2.867,-6.08) {0};
\draw[vizgrid] (4.3,-6) -- (4.3,0.05);
\node[viztick,anchor=north] at (4.3,-6.08) {+0.25};
\node[viztick,anchor=north] at (2.15,-6.42) {$\Delta_1$: C1 $-$ C0 (24 tasks)};
\draw[vizInk2,line width=0.5pt] (2.867,-6) -- (2.867,0.05);
\draw[vizInk2,line width=0.6pt] (1.433,-0.2) -- (2.389,-0.2);
\filldraw[fill=vizP2,draw=white,line width=0.5pt] (1.911,-0.2) circle (0.065);
\draw[vizInk2,line width=0.6pt] (1.672,-0.6) -- (3.106,-0.6);
\filldraw[fill=vizP2,draw=white,line width=0.5pt] (2.389,-0.6) circle (0.065);
\draw[vizInk2,line width=0.6pt] (1.194,-1) -- (3.106,-1);
\filldraw[fill=vizP1,draw=white,line width=0.5pt] (2.15,-1) circle (0.065);
\draw[vizInk2,line width=0.6pt] (0.717,-1.4) -- (2.628,-1.4);
\filldraw[fill=vizP1,draw=white,line width=0.5pt] (1.672,-1.4) circle (0.065);
\draw[vizInk2,line width=0.6pt] (1.194,-1.8) -- (2.628,-1.8);
\filldraw[fill=vizP1,draw=white,line width=0.5pt] (1.911,-1.8) circle (0.065);
\draw[vizInk2,line width=0.6pt] (0.956,-2.2) -- (2.628,-2.2);
\filldraw[fill=vizP2,draw=white,line width=0.5pt] (1.911,-2.2) circle (0.065);
\draw[vizInk2,line width=0.6pt] (0.956,-2.6) -- (2.628,-2.6);
\filldraw[fill=vizP3,draw=white,line width=0.5pt] (1.911,-2.6) circle (0.065);
\draw[vizInk2,line width=0.6pt] (1.433,-3) -- (2.867,-3);
\filldraw[fill=vizP3,draw=white,line width=0.5pt] (2.15,-3) circle (0.065);
\draw[vizInk2,line width=0.6pt] (1.672,-3.4) -- (3.583,-3.4);
\filldraw[fill=vizP2,draw=white,line width=0.5pt] (2.628,-3.4) circle (0.065);
\draw[vizInk2,line width=0.6pt] (1.433,-3.8) -- (3.583,-3.8);
\filldraw[fill=vizP2,draw=white,line width=0.5pt] (2.628,-3.8) circle (0.065);
\draw[vizInk2,line width=0.6pt] (1.672,-4.2) -- (2.628,-4.2);
\filldraw[fill=vizP3,draw=white,line width=0.5pt] (2.15,-4.2) circle (0.065);
\draw[vizInk2,line width=0.6pt] (2.867,-4.6) -- (3.583,-4.6);
\filldraw[fill=vizP1,draw=white,line width=0.5pt] (3.106,-4.6) circle (0.065);
\draw[vizInk2,line width=0.6pt] (2.867,-5) -- (3.583,-5);
\filldraw[fill=vizP1,draw=white,line width=0.5pt] (3.106,-5) circle (0.065);
\filldraw[fill=vizP3,draw=white,line width=0.5pt] (2.867,-5.4) circle (0.065);
\filldraw[fill=vizP3,draw=white,line width=0.5pt] (2.867,-5.8) circle (0.065);
\draw[vizgrid] (5.3,-6) -- (5.3,0.05);
\node[viztick,anchor=north] at (5.3,-6.08) {$-$0.75};
\draw[vizgrid] (6.375,-6) -- (6.375,0.05);
\node[viztick,anchor=north] at (6.375,-6.08) {$-$0.5};
\draw[vizgrid] (7.45,-6) -- (7.45,0.05);
\node[viztick,anchor=north] at (7.45,-6.08) {$-$0.25};
\draw[vizgrid] (8.525,-6) -- (8.525,0.05);
\node[viztick,anchor=north] at (8.525,-6.08) {0};
\draw[vizgrid] (9.6,-6) -- (9.6,0.05);
\node[viztick,anchor=north] at (9.6,-6.08) {+0.25};
\node[viztick,anchor=north] at (7.45,-6.42) {$\Delta_2$: C2 $-$ C0 (12 tasks)};
\draw[vizInk2,line width=0.5pt] (8.525,-6) -- (8.525,0.05);
\draw[vizInk2,line width=0.6pt] (6.017,-0.2) -- (7.271,-0.2);
\filldraw[fill=vizP2,draw=white,line width=0.5pt] (6.554,-0.2) circle (0.065);
\draw[vizInk2,line width=0.6pt] (6.196,-0.6) -- (7.271,-0.6);
\filldraw[fill=vizP2,draw=white,line width=0.5pt] (6.733,-0.6) circle (0.065);
\draw[vizInk2,line width=0.6pt] (7.45,-1) -- (8.883,-1);
\filldraw[fill=vizP1,draw=white,line width=0.5pt] (8.167,-1) circle (0.065);
\draw[vizInk2,line width=0.6pt] (6.375,-1.4) -- (8.525,-1.4);
\filldraw[fill=vizP1,draw=white,line width=0.5pt] (7.45,-1.4) circle (0.065);
\draw[vizInk2,line width=0.6pt] (6.733,-1.8) -- (8.883,-1.8);
\filldraw[fill=vizP1,draw=white,line width=0.5pt] (7.808,-1.8) circle (0.065);
\draw[vizInk2,line width=0.6pt] (6.733,-2.2) -- (8.167,-2.2);
\filldraw[fill=vizP2,draw=white,line width=0.5pt] (7.45,-2.2) circle (0.065);
\draw[vizInk2,line width=0.6pt] (6.375,-2.6) -- (8.167,-2.6);
\filldraw[fill=vizP3,draw=white,line width=0.5pt] (7.271,-2.6) circle (0.065);
\draw[vizInk2,line width=0.6pt] (6.733,-3) -- (8.525,-3);
\filldraw[fill=vizP3,draw=white,line width=0.5pt] (7.629,-3) circle (0.065);
\draw[vizInk2,line width=0.6pt] (7.092,-3.4) -- (8.883,-3.4);
\filldraw[fill=vizP2,draw=white,line width=0.5pt] (7.987,-3.4) circle (0.065);
\draw[vizInk2,line width=0.6pt] (7.271,-3.8) -- (8.525,-3.8);
\filldraw[fill=vizP2,draw=white,line width=0.5pt] (7.987,-3.8) circle (0.065);
\draw[vizInk2,line width=0.6pt] (7.987,-4.2) -- (8.525,-4.2);
\filldraw[fill=vizP3,draw=white,line width=0.5pt] (8.167,-4.2) circle (0.065);
\draw[vizInk2,line width=0.6pt] (7.808,-4.6) -- (9.242,-4.6);
\filldraw[fill=vizP1,draw=white,line width=0.5pt] (8.525,-4.6) circle (0.065);
\filldraw[fill=vizP1,draw=white,line width=0.5pt] (8.525,-5) circle (0.065);
\filldraw[fill=vizP3,draw=white,line width=0.5pt] (8.525,-5.4) circle (0.065);
\filldraw[fill=vizP3,draw=white,line width=0.5pt] (8.525,-5.8) circle (0.065);
\node[vizlab,anchor=east] at (-0.12,-0.2) {L09 qwen3.6 27b-coding-mtp};
\node[vizlab,anchor=east] at (-0.12,-0.6) {L08 gemma4 31b};
\node[vizlab,anchor=east] at (-0.12,-1) {L01 qwen3.6 35b-a3b};
\node[vizlab,anchor=east] at (-0.12,-1.4) {L03 granite4.2 30b};
\node[vizlab,anchor=east] at (-0.12,-1.8) {L02 gemma4 26b-a4b};
\node[vizlab,anchor=east] at (-0.12,-2.2) {L07 nemotron-3.5-lightning 30b-a3b};
\node[vizlab,anchor=east] at (-0.12,-2.6) {L11 qwen3.5 9b};
\node[vizlab,anchor=east] at (-0.12,-3) {L13 granite4.2 8b};
\node[vizlab,anchor=east] at (-0.12,-3.4) {L10 gemma4 12b};
\node[vizlab,anchor=east] at (-0.12,-3.8) {L06 glm-4.7-flash};
\node[vizlab,anchor=east] at (-0.12,-4.2) {L12 gemma4 e4b};
\node[vizlab,anchor=east] at (-0.12,-4.6) {L04 lfm2.5 24b-a2b};
\node[vizlab,anchor=east] at (-0.12,-5) {L05 north-mini-code-1.0};
\node[vizlab,anchor=east] at (-0.12,-5.4) {L14 nemotron-3-nano 4b};
\node[vizlab,anchor=east] at (-0.12,-5.8) {L15 lfm2.5 8b-a1b};
\node[fill=vizP1,minimum width=0.22cm,minimum height=0.15cm,inner sep=0,anchor=west] (lgs1) at (0,0.35) {};
\node[viztick,anchor=west,xshift=0.06cm] (lgl1) at (lgs1.east) {Apple M4 Max};
\node[fill=vizP2,minimum width=0.22cm,minimum height=0.15cm,inner sep=0,anchor=west,xshift=0.3cm] (lgs2) at (lgl1.east) {};
\node[viztick,anchor=west,xshift=0.06cm] (lgl2) at (lgs2.east) {Apple M5 Max};
\node[fill=vizP3,minimum width=0.22cm,minimum height=0.15cm,inner sep=0,anchor=west,xshift=0.3cm] (lgs3) at (lgl2.east) {};
\node[viztick,anchor=west,xshift=0.06cm] (lgl3) at (lgs3.east) {RTX 3090};
\node[inner sep=0,anchor=west,xshift=0.3cm] (lgs4) at (lgl3.east) {\textcolor{vizInk2}{\rule[0.3ex]{0.5pt}{1ex}\rule[0.75ex]{0.35cm}{0.5pt}\rule[0.3ex]{0.5pt}{1ex}}};
\node[viztick,anchor=west,xshift=0.06cm] (lgl4) at (lgs4.east) {family-resampling range};
\end{tikzpicture}
\caption{Condition contrasts per configuration. Left: $\Delta_1$, the mean change in G from C0 to C1 over the 24 tasks. Right: $\Delta_2$, the mean change from C0 to the average of the two C2 worlds over the twelve state-changing tasks. Lines are family-resampling ranges.}\label{fig:contrasts}
\end{figure}

\begin{figure}[tbp]
\centering
\input{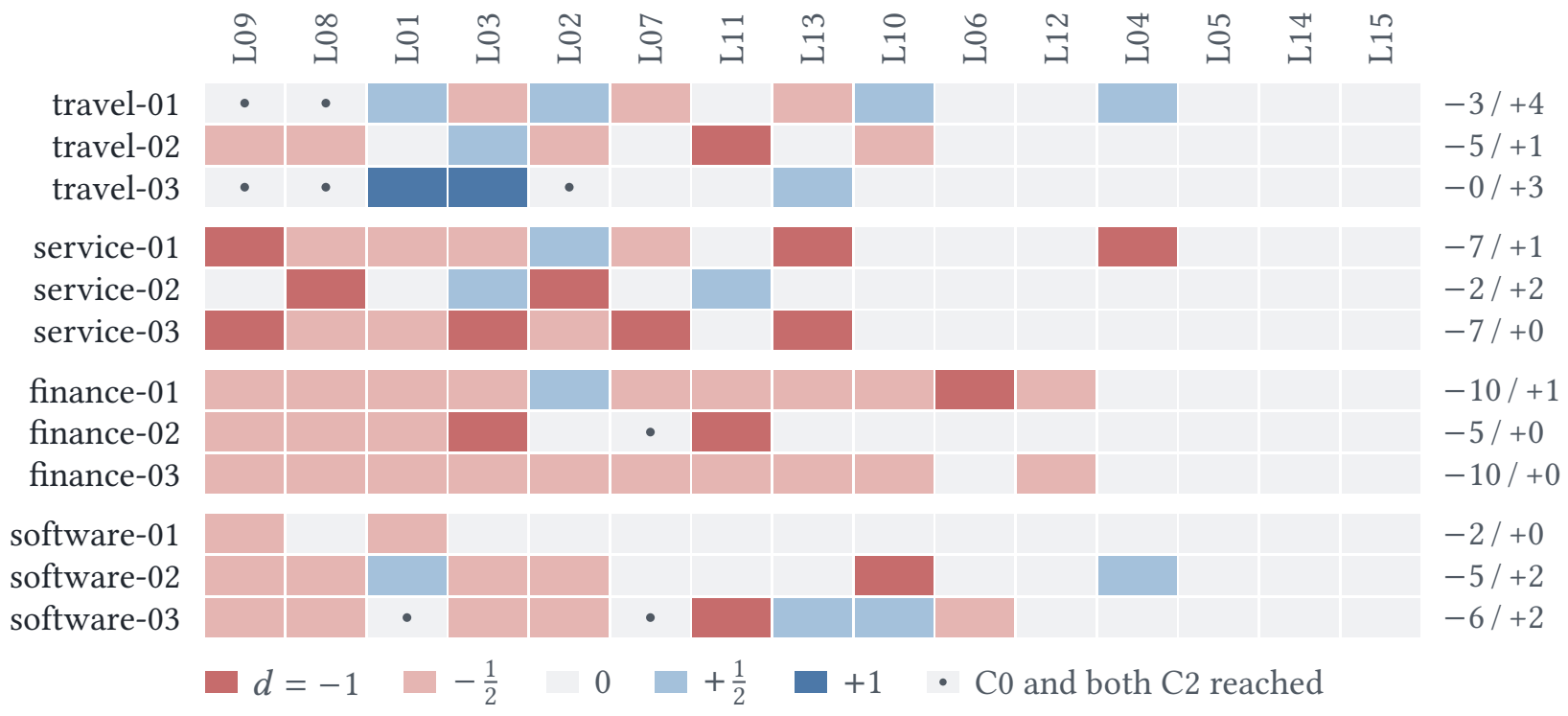}
\caption{Change from C0 to C2 on the twelve state-changing tasks. $d$ is the mean G over the two hidden worlds minus G in C0. Columns are sorted as in Figure~\ref{fig:leaderboard}.}\label{fig:heat-c2}
\end{figure}

\begin{figure}[tbp]
\centering
\input{generated/figures/fig-06-group-profile}
\caption{C0 group profile: tasks reached of the three in each scenario group, per configuration.}\label{fig:group-profile}
\end{figure}

\begin{figure}[tbp]
\centering
\begin{tikzpicture}[x=1cm,y=1cm,line cap=butt]
\draw[vizgrid] (0,-6) -- (0,0.05);
\node[viztick,anchor=north] at (0,-6.08) {0\%};
\draw[vizgrid] (1.9,-6) -- (1.9,0.05);
\node[viztick,anchor=north] at (1.9,-6.08) {25\%};
\draw[vizgrid] (3.8,-6) -- (3.8,0.05);
\node[viztick,anchor=north] at (3.8,-6.08) {50\%};
\draw[vizgrid] (5.7,-6) -- (5.7,0.05);
\node[viztick,anchor=north] at (5.7,-6.08) {75\%};
\draw[vizgrid] (7.6,-6) -- (7.6,0.05);
\node[viztick,anchor=north] at (7.6,-6.08) {100\%};
\node[viztick,anchor=north] at (3.8,-6.42) {share of episodes with G, P and E all defined};
\fill[vizSuccess] (0,-0.32) rectangle (5.026,-0.08);
\draw[vizgap] (0,-0.33) -- (0,-0.07);
\fill[vizNone] (5.026,-0.32) rectangle (7.6,-0.08);
\draw[vizgap] (5.026,-0.33) -- (5.026,-0.07);
\node[viztick,anchor=west] at (7.7,-0.2) {J 41/62};
\fill[vizSuccess] (0,-0.72) rectangle (5.639,-0.48);
\draw[vizgap] (0,-0.73) -- (0,-0.47);
\fill[vizProcess] (5.639,-0.72) rectangle (5.884,-0.48);
\draw[vizgap] (5.639,-0.73) -- (5.639,-0.47);
\fill[vizNone] (5.884,-0.72) rectangle (7.6,-0.48);
\draw[vizgap] (5.884,-0.73) -- (5.884,-0.47);
\node[viztick,anchor=west] at (7.7,-0.6) {J 46/62};
\fill[vizSuccess] (0,-1.12) rectangle (4.535,-0.88);
\draw[vizgap] (0,-1.13) -- (0,-0.87);
\fill[vizEvidence] (4.535,-1.12) rectangle (4.658,-0.88);
\draw[vizgap] (4.535,-1.13) -- (4.535,-0.87);
\fill[vizNone] (4.658,-1.12) rectangle (7.6,-0.88);
\draw[vizgap] (4.658,-1.13) -- (4.658,-0.87);
\node[viztick,anchor=west] at (7.7,-1) {J 37/62};
\fill[vizSuccess] (0,-1.52) rectangle (4.29,-1.28);
\draw[vizgap] (0,-1.53) -- (0,-1.27);
\fill[vizNone] (4.29,-1.52) rectangle (7.6,-1.28);
\draw[vizgap] (4.29,-1.53) -- (4.29,-1.27);
\node[viztick,anchor=west] at (7.7,-1.4) {J 35/62};
\fill[vizSuccess] (0,-1.92) rectangle (3.065,-1.68);
\draw[vizgap] (0,-1.93) -- (0,-1.67);
\fill[vizEvidence] (3.065,-1.92) rectangle (3.31,-1.68);
\draw[vizgap] (3.065,-1.93) -- (3.065,-1.67);
\fill[vizNone] (3.31,-1.92) rectangle (7.6,-1.68);
\draw[vizgap] (3.31,-1.93) -- (3.31,-1.67);
\node[viztick,anchor=west] at (7.7,-1.8) {J 25/62};
\fill[vizSuccess] (0,-2.32) rectangle (2.697,-2.08);
\draw[vizgap] (0,-2.33) -- (0,-2.07);
\fill[vizEvidence] (2.697,-2.32) rectangle (2.819,-2.08);
\draw[vizgap] (2.697,-2.33) -- (2.697,-2.07);
\fill[vizNone] (2.819,-2.32) rectangle (7.6,-2.08);
\draw[vizgap] (2.819,-2.33) -- (2.819,-2.07);
\node[viztick,anchor=west] at (7.7,-2.2) {J 22/62};
\fill[vizSuccess] (0,-2.72) rectangle (1.594,-2.48);
\draw[vizgap] (0,-2.73) -- (0,-2.47);
\fill[vizEvidence] (1.594,-2.72) rectangle (1.716,-2.48);
\draw[vizgap] (1.594,-2.73) -- (1.594,-2.47);
\fill[vizNone] (1.716,-2.72) rectangle (7.6,-2.48);
\draw[vizgap] (1.716,-2.73) -- (1.716,-2.47);
\node[viztick,anchor=west] at (7.7,-2.6) {J 13/62};
\fill[vizSuccess] (0,-3.12) rectangle (1.961,-2.88);
\draw[vizgap] (0,-3.13) -- (0,-2.87);
\fill[vizNone] (1.961,-3.12) rectangle (7.6,-2.88);
\draw[vizgap] (1.961,-3.13) -- (1.961,-2.87);
\node[viztick,anchor=west] at (7.7,-3) {J 16/62};
\fill[vizSuccess] (0,-3.52) rectangle (1.716,-3.28);
\draw[vizgap] (0,-3.53) -- (0,-3.27);
\fill[vizEvidence] (1.716,-3.52) rectangle (1.961,-3.28);
\draw[vizgap] (1.716,-3.53) -- (1.716,-3.27);
\fill[vizNone] (1.961,-3.52) rectangle (7.6,-3.28);
\draw[vizgap] (1.961,-3.53) -- (1.961,-3.27);
\node[viztick,anchor=west] at (7.7,-3.4) {J 14/62};
\fill[vizSuccess] (0,-3.92) rectangle (0.735,-3.68);
\draw[vizgap] (0,-3.93) -- (0,-3.67);
\fill[vizProcess] (0.735,-3.92) rectangle (0.858,-3.68);
\draw[vizgap] (0.735,-3.93) -- (0.735,-3.67);
\fill[vizNone] (0.858,-3.92) rectangle (7.6,-3.68);
\draw[vizgap] (0.858,-3.93) -- (0.858,-3.67);
\node[viztick,anchor=west] at (7.7,-3.8) {J 6/62};
\fill[vizSuccess] (0,-4.32) rectangle (0.49,-4.08);
\draw[vizgap] (0,-4.33) -- (0,-4.07);
\fill[vizEvidence] (0.49,-4.32) rectangle (0.613,-4.08);
\draw[vizgap] (0.49,-4.33) -- (0.49,-4.07);
\fill[vizNone] (0.613,-4.32) rectangle (7.6,-4.08);
\draw[vizgap] (0.613,-4.33) -- (0.613,-4.07);
\node[viztick,anchor=west] at (7.7,-4.2) {J 4/62};
\fill[vizSuccess] (0,-4.72) rectangle (0.613,-4.48);
\draw[vizgap] (0,-4.73) -- (0,-4.47);
\fill[vizNone] (0.613,-4.72) rectangle (7.6,-4.48);
\draw[vizgap] (0.613,-4.73) -- (0.613,-4.47);
\node[viztick,anchor=west] at (7.7,-4.6) {J 5/62};
\fill[vizSuccess] (0,-5.12) rectangle (0.123,-4.88);
\draw[vizgap] (0,-5.13) -- (0,-4.87);
\fill[vizNone] (0.123,-5.12) rectangle (7.6,-4.88);
\draw[vizgap] (0.123,-5.13) -- (0.123,-4.87);
\node[viztick,anchor=west] at (7.7,-5) {J 1/62};
\fill[vizNone] (0,-5.52) rectangle (7.6,-5.28);
\draw[vizgap] (0,-5.53) -- (0,-5.27);
\node[viztick,anchor=west] at (7.7,-5.4) {J 0/62};
\fill[vizNone] (0,-5.92) rectangle (7.6,-5.68);
\draw[vizgap] (0,-5.93) -- (0,-5.67);
\node[viztick,anchor=west] at (7.7,-5.8) {J 0/62};
\node[vizlab,anchor=east] at (-0.12,-0.2) {L09 qwen3.6 27b-coding-mtp};
\node[vizlab,anchor=east] at (-0.12,-0.6) {L08 gemma4 31b};
\node[vizlab,anchor=east] at (-0.12,-1) {L01 qwen3.6 35b-a3b};
\node[vizlab,anchor=east] at (-0.12,-1.4) {L03 granite4.2 30b};
\node[vizlab,anchor=east] at (-0.12,-1.8) {L02 gemma4 26b-a4b};
\node[vizlab,anchor=east] at (-0.12,-2.2) {L07 nemotron-3.5-lightning 30b-a3b};
\node[vizlab,anchor=east] at (-0.12,-2.6) {L11 qwen3.5 9b};
\node[vizlab,anchor=east] at (-0.12,-3) {L13 granite4.2 8b};
\node[vizlab,anchor=east] at (-0.12,-3.4) {L10 gemma4 12b};
\node[vizlab,anchor=east] at (-0.12,-3.8) {L06 glm-4.7-flash};
\node[vizlab,anchor=east] at (-0.12,-4.2) {L12 gemma4 e4b};
\node[vizlab,anchor=east] at (-0.12,-4.6) {L04 lfm2.5 24b-a2b};
\node[vizlab,anchor=east] at (-0.12,-5) {L05 north-mini-code-1.0};
\node[vizlab,anchor=east] at (-0.12,-5.4) {L14 nemotron-3-nano 4b};
\node[vizlab,anchor=east] at (-0.12,-5.8) {L15 lfm2.5 8b-a1b};
\node[fill=vizSuccess,minimum width=0.22cm,minimum height=0.15cm,inner sep=0,anchor=west] (lgs1) at (0,0.62) {};
\node[viztick,anchor=west,xshift=0.06cm] (lgl1) at (lgs1.east) {joint success J};
\node[fill=vizProcess,minimum width=0.22cm,minimum height=0.15cm,inner sep=0,anchor=west,xshift=0.3cm] (lgs2) at (lgl1.east) {};
\node[viztick,anchor=west,xshift=0.06cm] (lgl2) at (lgs2.east) {goal, process violated};
\node[fill=vizEvidence,minimum width=0.22cm,minimum height=0.15cm,inner sep=0,anchor=west] (lgs3) at (0,0.28) {};
\node[viztick,anchor=west,xshift=0.06cm] (lgl3) at (lgs3.east) {goal and process, evidence fails};
\node[fill=vizNone,minimum width=0.22cm,minimum height=0.15cm,inner sep=0,anchor=west,xshift=0.3cm] (lgs4) at (lgl3.east) {};
\node[viztick,anchor=west,xshift=0.06cm] (lgl4) at (lgs4.east) {goal not reached};
\end{tikzpicture}
\caption{Partition of outcomes over all conditions, on the records where G, P, and E are all defined: joint success, goal reached with a process violation, goal and process with insufficient evidence, and goal not reached. Exact counts are in Table~\ref{tab:endpoints}.}\label{fig:partition}
\end{figure}
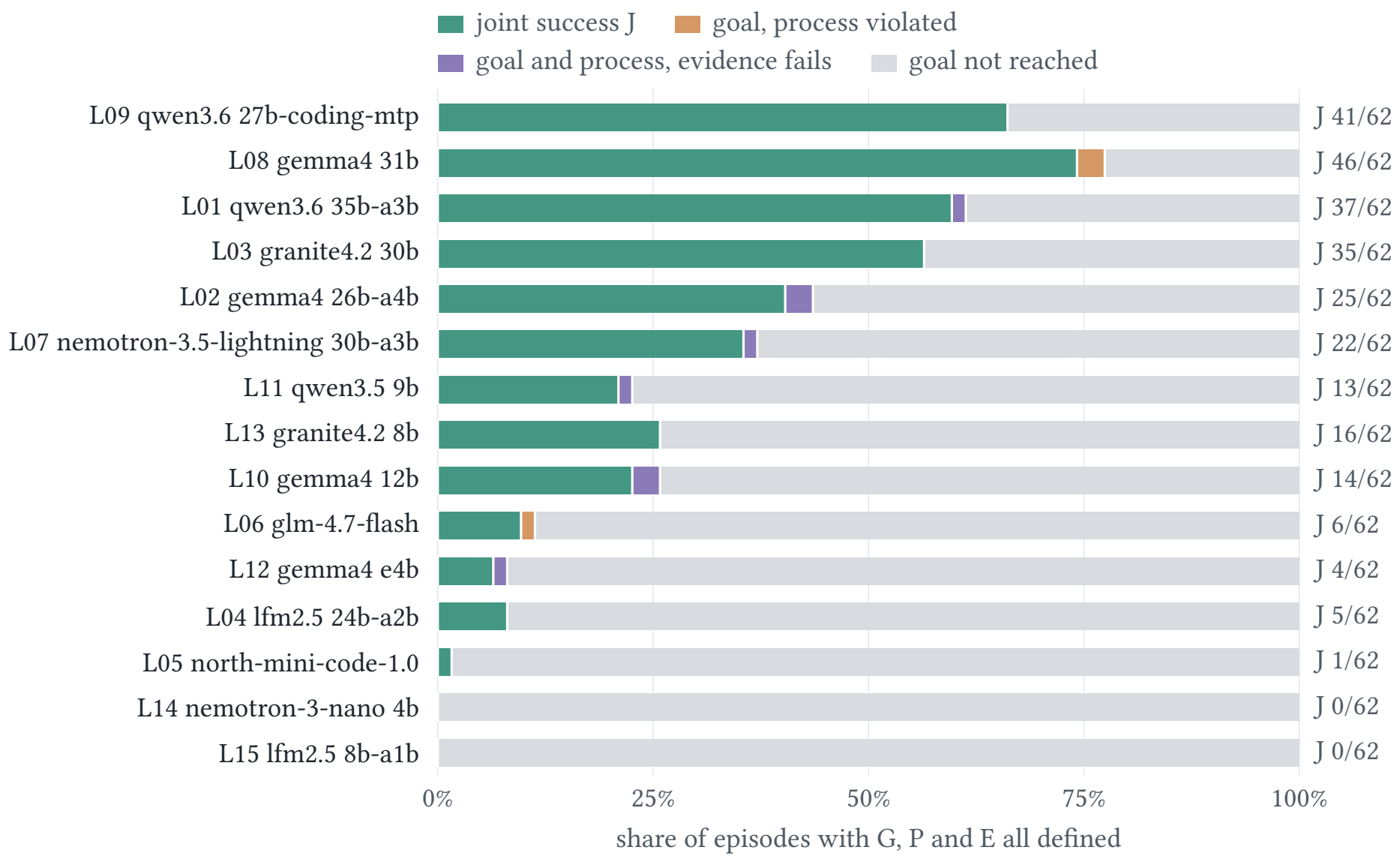

\begin{figure}[tbp]
\centering
\begin{tikzpicture}[x=1cm,y=1cm,line cap=butt]
\draw[vizgrid] (0,-6) -- (0,0.05);
\node[viztick,anchor=north] at (0,-6.08) {0\%};
\draw[vizgrid] (1.9,-6) -- (1.9,0.05);
\node[viztick,anchor=north] at (1.9,-6.08) {25\%};
\draw[vizgrid] (3.8,-6) -- (3.8,0.05);
\node[viztick,anchor=north] at (3.8,-6.08) {50\%};
\draw[vizgrid] (5.7,-6) -- (5.7,0.05);
\node[viztick,anchor=north] at (5.7,-6.08) {75\%};
\draw[vizgrid] (7.6,-6) -- (7.6,0.05);
\node[viztick,anchor=north] at (7.6,-6.08) {100\%};
\node[viztick,anchor=north] at (3.8,-6.42) {share of the 72 episodes};
\fill[vizBlue45] (0,-0.32) rectangle (7.389,-0.08);
\draw[vizgap] (0,-0.33) -- (0,-0.07);
\fill[vizP2] (7.389,-0.32) rectangle (7.494,-0.08);
\draw[vizgap] (7.389,-0.33) -- (7.389,-0.07);
\fill[vizS5] (7.494,-0.32) rectangle (7.6,-0.08);
\draw[vizgap] (7.494,-0.33) -- (7.494,-0.07);
\node[viztick,anchor=west] at (7.7,-0.2) {70};
\fill[vizBlue45] (0,-0.72) rectangle (7.283,-0.48);
\draw[vizgap] (0,-0.73) -- (0,-0.47);
\fill[vizS7] (7.283,-0.72) rectangle (7.494,-0.48);
\draw[vizgap] (7.283,-0.73) -- (7.283,-0.47);
\fill[vizS4] (7.494,-0.72) rectangle (7.6,-0.48);
\draw[vizgap] (7.494,-0.73) -- (7.494,-0.47);
\node[viztick,anchor=west] at (7.7,-0.6) {69};
\fill[vizBlue45] (0,-1.12) rectangle (7.389,-0.88);
\draw[vizgap] (0,-1.13) -- (0,-0.87);
\fill[vizP2] (7.389,-1.12) rectangle (7.494,-0.88);
\draw[vizgap] (7.389,-1.13) -- (7.389,-0.87);
\fill[vizS5] (7.494,-1.12) rectangle (7.6,-0.88);
\draw[vizgap] (7.494,-1.13) -- (7.494,-0.87);
\node[viztick,anchor=west] at (7.7,-1) {70};
\fill[vizBlue45] (0,-1.52) rectangle (7.283,-1.28);
\draw[vizgap] (0,-1.53) -- (0,-1.27);
\fill[vizS7] (7.283,-1.52) rectangle (7.494,-1.28);
\draw[vizgap] (7.283,-1.53) -- (7.283,-1.27);
\fill[vizS5] (7.494,-1.52) rectangle (7.6,-1.28);
\draw[vizgap] (7.494,-1.53) -- (7.494,-1.27);
\node[viztick,anchor=west] at (7.7,-1.4) {69};
\fill[vizBlue45] (0,-1.92) rectangle (6.967,-1.68);
\draw[vizgap] (0,-1.93) -- (0,-1.67);
\fill[vizS7] (6.967,-1.92) rectangle (7.389,-1.68);
\draw[vizgap] (6.967,-1.93) -- (6.967,-1.67);
\fill[vizS4] (7.389,-1.92) rectangle (7.6,-1.68);
\draw[vizgap] (7.389,-1.93) -- (7.389,-1.67);
\node[viztick,anchor=west] at (7.7,-1.8) {66};
\fill[vizBlue45] (0,-2.32) rectangle (7.072,-2.08);
\draw[vizgap] (0,-2.33) -- (0,-2.07);
\fill[vizP2] (7.072,-2.32) rectangle (7.178,-2.08);
\draw[vizgap] (7.072,-2.33) -- (7.072,-2.07);
\fill[vizS5] (7.178,-2.32) rectangle (7.6,-2.08);
\draw[vizgap] (7.178,-2.33) -- (7.178,-2.07);
\node[viztick,anchor=west] at (7.7,-2.2) {67};
\fill[vizBlue45] (0,-2.72) rectangle (6.228,-2.48);
\draw[vizgap] (0,-2.73) -- (0,-2.47);
\fill[vizS7] (6.228,-2.72) rectangle (6.439,-2.48);
\draw[vizgap] (6.228,-2.73) -- (6.228,-2.47);
\fill[vizS4] (6.439,-2.72) rectangle (7.6,-2.48);
\draw[vizgap] (6.439,-2.73) -- (6.439,-2.47);
\node[viztick,anchor=west] at (7.7,-2.6) {59};
\fill[vizBlue45] (0,-3.12) rectangle (5.383,-2.88);
\draw[vizgap] (0,-3.13) -- (0,-2.87);
\fill[vizS7] (5.383,-3.12) rectangle (6.439,-2.88);
\draw[vizgap] (5.383,-3.13) -- (5.383,-2.87);
\fill[vizS4] (6.439,-3.12) rectangle (7.494,-2.88);
\draw[vizgap] (6.439,-3.13) -- (6.439,-2.87);
\fill[vizS5] (7.494,-3.12) rectangle (7.6,-2.88);
\draw[vizgap] (7.494,-3.13) -- (7.494,-2.87);
\node[viztick,anchor=west] at (7.7,-3) {51};
\fill[vizBlue45] (0,-3.52) rectangle (6.017,-3.28);
\draw[vizgap] (0,-3.53) -- (0,-3.27);
\fill[vizP2] (6.017,-3.52) rectangle (6.122,-3.28);
\draw[vizgap] (6.017,-3.53) -- (6.017,-3.27);
\fill[vizS7] (6.122,-3.52) rectangle (6.756,-3.28);
\draw[vizgap] (6.122,-3.53) -- (6.122,-3.27);
\fill[vizS4] (6.756,-3.52) rectangle (6.967,-3.28);
\draw[vizgap] (6.756,-3.53) -- (6.756,-3.27);
\fill[vizS5] (6.967,-3.52) rectangle (7.6,-3.28);
\draw[vizgap] (6.967,-3.53) -- (6.967,-3.27);
\node[viztick,anchor=west] at (7.7,-3.4) {57};
\fill[vizBlue45] (0,-3.92) rectangle (6.967,-3.68);
\draw[vizgap] (0,-3.93) -- (0,-3.67);
\fill[vizS7] (6.967,-3.92) rectangle (7.072,-3.68);
\draw[vizgap] (6.967,-3.93) -- (6.967,-3.67);
\fill[vizS4] (7.072,-3.92) rectangle (7.389,-3.68);
\draw[vizgap] (7.072,-3.93) -- (7.072,-3.67);
\fill[vizS5] (7.389,-3.92) rectangle (7.6,-3.68);
\draw[vizgap] (7.389,-3.93) -- (7.389,-3.67);
\node[viztick,anchor=west] at (7.7,-3.8) {66};
\fill[vizBlue45] (0,-4.32) rectangle (7.283,-4.08);
\draw[vizgap] (0,-4.33) -- (0,-4.07);
\fill[vizS7] (7.283,-4.32) rectangle (7.494,-4.08);
\draw[vizgap] (7.283,-4.33) -- (7.283,-4.07);
\fill[vizS4] (7.494,-4.32) rectangle (7.6,-4.08);
\draw[vizgap] (7.494,-4.33) -- (7.494,-4.07);
\node[viztick,anchor=west] at (7.7,-4.2) {69};
\fill[vizBlue45] (0,-4.72) rectangle (5.806,-4.48);
\draw[vizgap] (0,-4.73) -- (0,-4.47);
\fill[vizP2] (5.806,-4.72) rectangle (7.389,-4.48);
\draw[vizgap] (5.806,-4.73) -- (5.806,-4.47);
\fill[vizS7] (7.389,-4.72) rectangle (7.494,-4.48);
\draw[vizgap] (7.389,-4.73) -- (7.389,-4.47);
\fill[vizS4] (7.494,-4.72) rectangle (7.6,-4.48);
\draw[vizgap] (7.494,-4.73) -- (7.494,-4.47);
\node[viztick,anchor=west] at (7.7,-4.6) {55};
\fill[vizBlue45] (0,-5.12) rectangle (1.689,-4.88);
\draw[vizgap] (0,-5.13) -- (0,-4.87);
\fill[vizP2] (1.689,-5.12) rectangle (7.283,-4.88);
\draw[vizgap] (1.689,-5.13) -- (1.689,-4.87);
\fill[vizS7] (7.283,-5.12) rectangle (7.389,-4.88);
\draw[vizgap] (7.283,-5.13) -- (7.283,-4.87);
\fill[vizS4] (7.389,-5.12) rectangle (7.494,-4.88);
\draw[vizgap] (7.389,-5.13) -- (7.389,-4.87);
\fill[vizS5] (7.494,-5.12) rectangle (7.6,-4.88);
\draw[vizgap] (7.494,-5.13) -- (7.494,-4.87);
\node[viztick,anchor=west] at (7.7,-5) {16};
\fill[vizBlue45] (0,-5.52) rectangle (0.317,-5.28);
\draw[vizgap] (0,-5.53) -- (0,-5.27);
\fill[vizP2] (0.317,-5.52) rectangle (7.072,-5.28);
\draw[vizgap] (0.317,-5.53) -- (0.317,-5.27);
\fill[vizS5] (7.072,-5.52) rectangle (7.6,-5.28);
\draw[vizgap] (7.072,-5.53) -- (7.072,-5.27);
\node[viztick,anchor=west] at (7.7,-5.4) {3};
\fill[vizBlue45] (0,-5.92) rectangle (3.8,-5.68);
\draw[vizgap] (0,-5.93) -- (0,-5.67);
\fill[vizP2] (3.8,-5.92) rectangle (6.861,-5.68);
\draw[vizgap] (3.8,-5.93) -- (3.8,-5.67);
\fill[vizS7] (6.861,-5.92) rectangle (7.283,-5.68);
\draw[vizgap] (6.861,-5.93) -- (6.861,-5.67);
\fill[vizS4] (7.283,-5.92) rectangle (7.494,-5.68);
\draw[vizgap] (7.283,-5.93) -- (7.283,-5.67);
\fill[vizS5] (7.494,-5.92) rectangle (7.6,-5.68);
\draw[vizgap] (7.494,-5.93) -- (7.494,-5.67);
\node[viztick,anchor=west] at (7.7,-5.8) {36};
\node[vizlab,anchor=east] at (-0.12,-0.2) {L09 qwen3.6 27b-coding-mtp};
\node[vizlab,anchor=east] at (-0.12,-0.6) {L08 gemma4 31b};
\node[vizlab,anchor=east] at (-0.12,-1) {L01 qwen3.6 35b-a3b};
\node[vizlab,anchor=east] at (-0.12,-1.4) {L03 granite4.2 30b};
\node[vizlab,anchor=east] at (-0.12,-1.8) {L02 gemma4 26b-a4b};
\node[vizlab,anchor=east] at (-0.12,-2.2) {L07 nemotron-3.5-lightning 30b-a3b};
\node[vizlab,anchor=east] at (-0.12,-2.6) {L11 qwen3.5 9b};
\node[vizlab,anchor=east] at (-0.12,-3) {L13 granite4.2 8b};
\node[vizlab,anchor=east] at (-0.12,-3.4) {L10 gemma4 12b};
\node[vizlab,anchor=east] at (-0.12,-3.8) {L06 glm-4.7-flash};
\node[vizlab,anchor=east] at (-0.12,-4.2) {L12 gemma4 e4b};
\node[vizlab,anchor=east] at (-0.12,-4.6) {L04 lfm2.5 24b-a2b};
\node[vizlab,anchor=east] at (-0.12,-5) {L05 north-mini-code-1.0};
\node[vizlab,anchor=east] at (-0.12,-5.4) {L14 nemotron-3-nano 4b};
\node[vizlab,anchor=east] at (-0.12,-5.8) {L15 lfm2.5 8b-a1b};
\node[fill=vizBlue45,minimum width=0.22cm,minimum height=0.15cm,inner sep=0,anchor=west] (lgs1) at (0,0.62) {};
\node[viztick,anchor=west,xshift=0.06cm] (lgl1) at (lgs1.east) {submitted};
\node[fill=vizP2,minimum width=0.22cm,minimum height=0.15cm,inner sep=0,anchor=west,xshift=0.3cm] (lgs2) at (lgl1.east) {};
\node[viztick,anchor=west,xshift=0.06cm] (lgl2) at (lgs2.east) {text-only reply};
\node[fill=vizS7,minimum width=0.22cm,minimum height=0.15cm,inner sep=0,anchor=west,xshift=0.3cm] (lgs3) at (lgl2.east) {};
\node[viztick,anchor=west,xshift=0.06cm] (lgl3) at (lgs3.east) {stuck (repetition)};
\node[fill=vizS4,minimum width=0.22cm,minimum height=0.15cm,inner sep=0,anchor=west] (lgs4) at (0,0.28) {};
\node[viztick,anchor=west,xshift=0.06cm] (lgl4) at (lgs4.east) {budget exhausted};
\node[fill=vizS5,minimum width=0.22cm,minimum height=0.15cm,inner sep=0,anchor=west,xshift=0.3cm] (lgs5) at (lgl4.east) {};
\node[viztick,anchor=west,xshift=0.06cm] (lgl5) at (lgs5.east) {malformed output};
\end{tikzpicture}
\caption{Terminal reasons over the 72 episodes of each configuration. Right: submitted episodes. Exact counts are in Table~\ref{tab:terminal}.}\label{fig:terminal}
\end{figure}
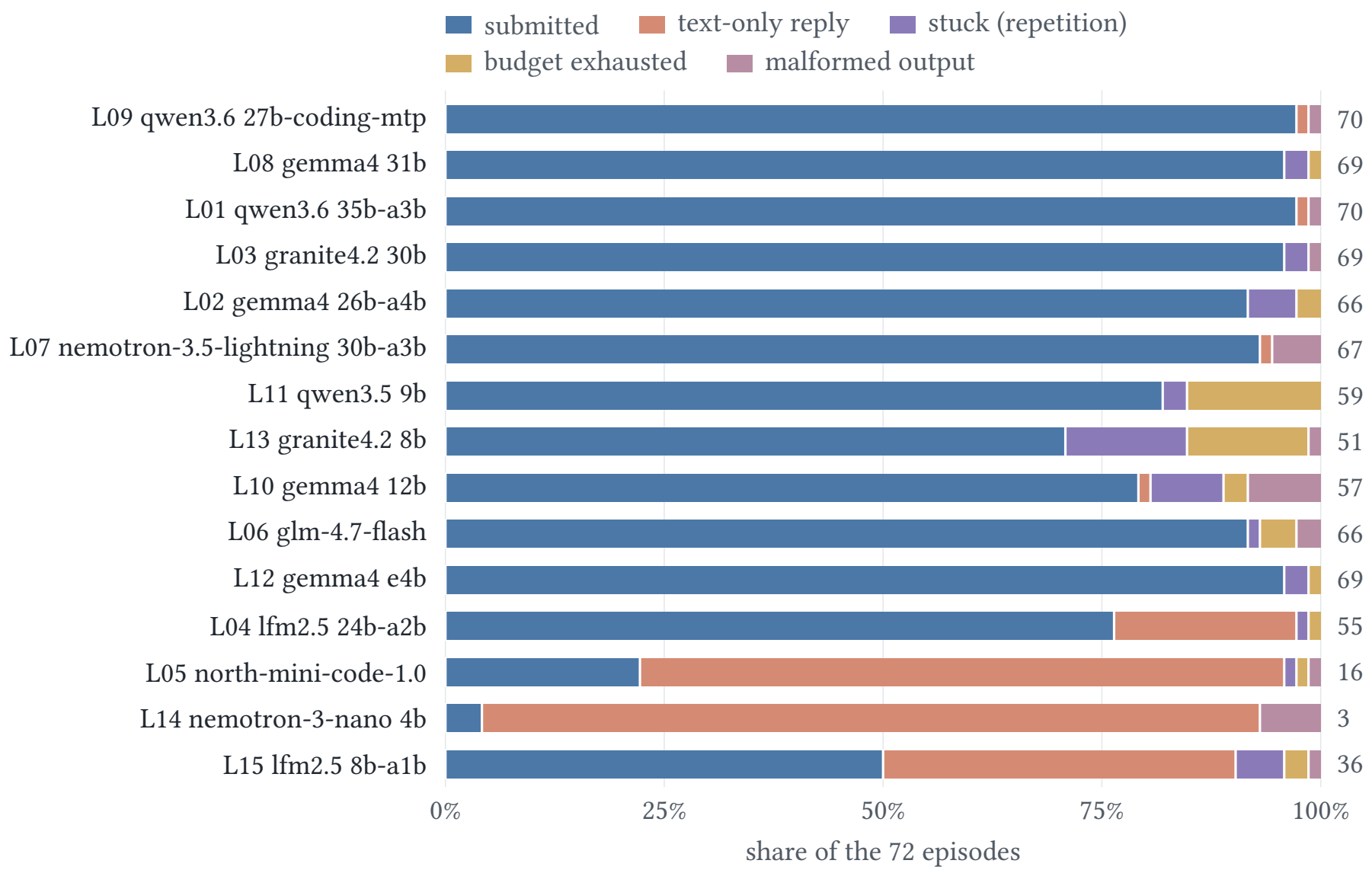

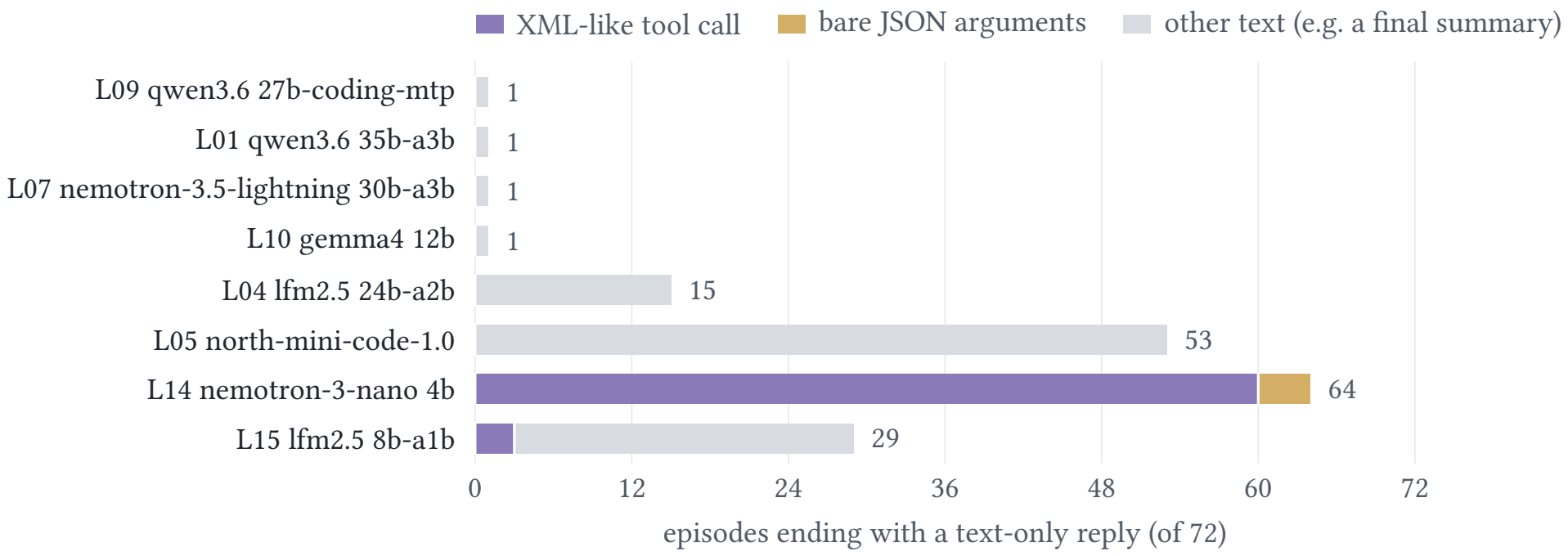
\begin{figure}[tbp]
\centering
\begin{tikzpicture}[x=1cm,y=1cm,line cap=butt]
\draw[vizgrid] (0,-3.2) -- (0,0.05);
\node[viztick,anchor=north] at (0,-3.28) {0};
\draw[vizgrid] (1.267,-3.2) -- (1.267,0.05);
\node[viztick,anchor=north] at (1.267,-3.28) {12};
\draw[vizgrid] (2.533,-3.2) -- (2.533,0.05);
\node[viztick,anchor=north] at (2.533,-3.28) {24};
\draw[vizgrid] (3.8,-3.2) -- (3.8,0.05);
\node[viztick,anchor=north] at (3.8,-3.28) {36};
\draw[vizgrid] (5.067,-3.2) -- (5.067,0.05);
\node[viztick,anchor=north] at (5.067,-3.28) {48};
\draw[vizgrid] (6.333,-3.2) -- (6.333,0.05);
\node[viztick,anchor=north] at (6.333,-3.28) {60};
\draw[vizgrid] (7.6,-3.2) -- (7.6,0.05);
\node[viztick,anchor=north] at (7.6,-3.28) {72};
\node[viztick,anchor=north] at (3.8,-3.62) {episodes ending with a text-only reply (of 72)};
\fill[vizNone] (0,-0.32) rectangle (0.106,-0.08);
\draw[vizgap] (0,-0.33) -- (0,-0.07);
\fill[vizNone] (0,-0.72) rectangle (0.106,-0.48);
\draw[vizgap] (0,-0.73) -- (0,-0.47);
\fill[vizNone] (0,-1.12) rectangle (0.106,-0.88);
\draw[vizgap] (0,-1.13) -- (0,-0.87);
\fill[vizNone] (0,-1.52) rectangle (0.106,-1.28);
\draw[vizgap] (0,-1.53) -- (0,-1.27);
\fill[vizNone] (0,-1.92) rectangle (1.583,-1.68);
\draw[vizgap] (0,-1.93) -- (0,-1.67);
\fill[vizNone] (0,-2.32) rectangle (5.594,-2.08);
\draw[vizgap] (0,-2.33) -- (0,-2.07);
\fill[vizS7] (0,-2.72) rectangle (6.333,-2.48);
\draw[vizgap] (0,-2.73) -- (0,-2.47);
\fill[vizS4] (6.333,-2.72) rectangle (6.756,-2.48);
\draw[vizgap] (6.333,-2.73) -- (6.333,-2.47);
\fill[vizS7] (0,-3.12) rectangle (0.317,-2.88);
\draw[vizgap] (0,-3.13) -- (0,-2.87);
\fill[vizNone] (0.317,-3.12) rectangle (3.061,-2.88);
\draw[vizgap] (0.317,-3.13) -- (0.317,-2.87);
\node[viztick,anchor=west] at (0.206,-0.2) {1};
\node[viztick,anchor=west] at (0.206,-0.6) {1};
\node[viztick,anchor=west] at (0.206,-1) {1};
\node[viztick,anchor=west] at (0.206,-1.4) {1};
\node[viztick,anchor=west] at (1.683,-1.8) {15};
\node[viztick,anchor=west] at (5.694,-2.2) {53};
\node[viztick,anchor=west] at (6.856,-2.6) {64};
\node[viztick,anchor=west] at (3.161,-3) {29};
\node[vizlab,anchor=east] at (-0.12,-0.2) {L09 qwen3.6 27b-coding-mtp};
\node[vizlab,anchor=east] at (-0.12,-0.6) {L01 qwen3.6 35b-a3b};
\node[vizlab,anchor=east] at (-0.12,-1) {L07 nemotron-3.5-lightning 30b-a3b};
\node[vizlab,anchor=east] at (-0.12,-1.4) {L10 gemma4 12b};
\node[vizlab,anchor=east] at (-0.12,-1.8) {L04 lfm2.5 24b-a2b};
\node[vizlab,anchor=east] at (-0.12,-2.2) {L05 north-mini-code-1.0};
\node[vizlab,anchor=east] at (-0.12,-2.6) {L14 nemotron-3-nano 4b};
\node[vizlab,anchor=east] at (-0.12,-3) {L15 lfm2.5 8b-a1b};
\node[fill=vizS7,minimum width=0.22cm,minimum height=0.15cm,inner sep=0,anchor=west] (lgs1) at (0,0.35) {};
\node[viztick,anchor=west,xshift=0.06cm] (lgl1) at (lgs1.east) {XML-like tool call};
\node[fill=vizS4,minimum width=0.22cm,minimum height=0.15cm,inner sep=0,anchor=west,xshift=0.25cm] (lgs2) at (lgl1.east) {};
\node[viztick,anchor=west,xshift=0.06cm] (lgl2) at (lgs2.east) {bare JSON arguments};
\node[fill=vizNone,minimum width=0.22cm,minimum height=0.15cm,inner sep=0,anchor=west,xshift=0.25cm] (lgs3) at (lgl2.east) {};
\node[viztick,anchor=west,xshift=0.06cm] (lgl3) at (lgs3.east) {other text (e.g.\ a final summary)};
\end{tikzpicture}
\caption{Episodes that ended with a text-only reply, by the content of that reply. The XML-like and bare-JSON replies are tool calls that the inference server did not parse.}\label{fig:text-replies}
\end{figure}

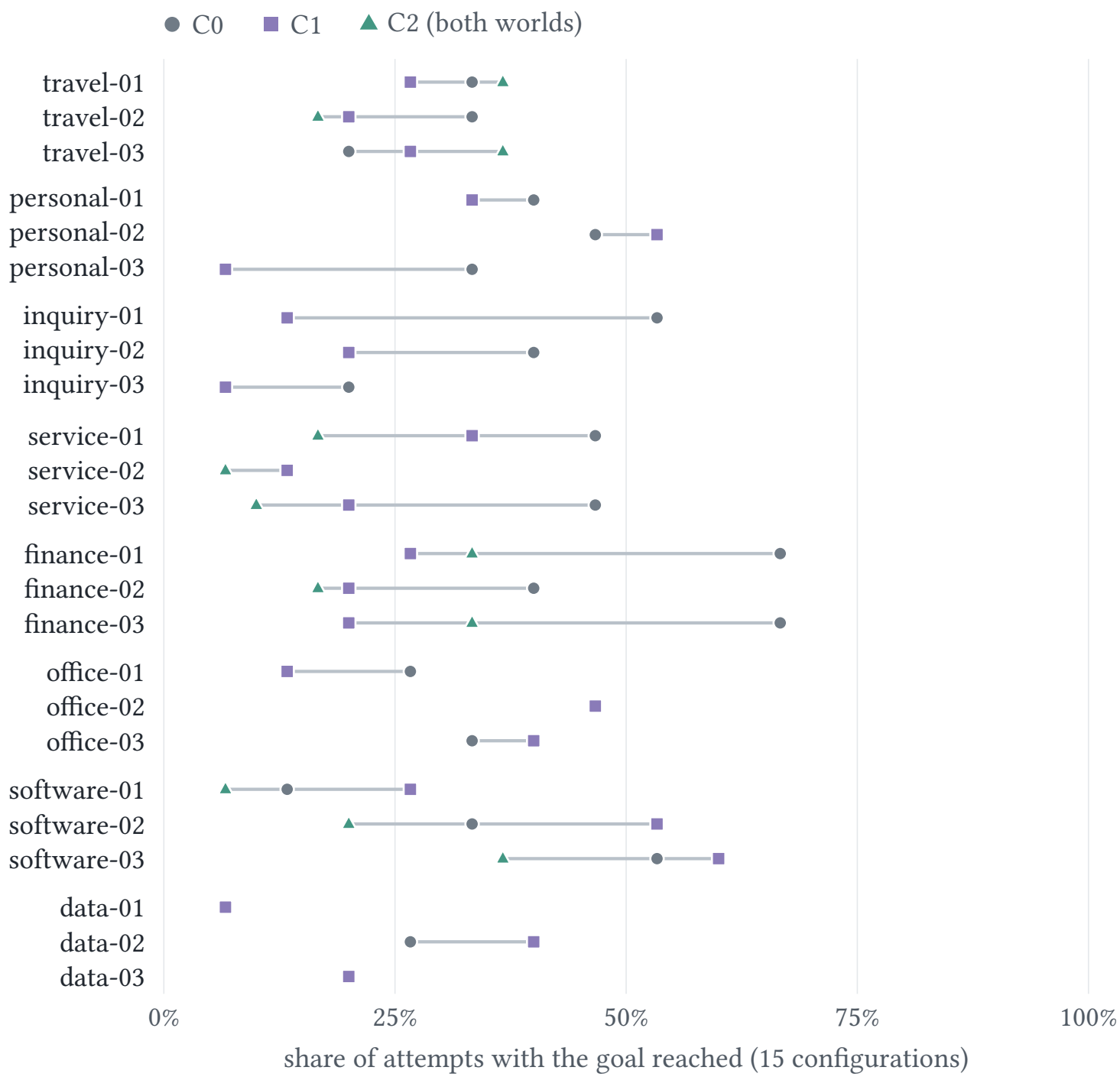
\begin{figure}[tbp]
\centering
\begin{tikzpicture}[x=1cm,y=1cm,line cap=butt]
\draw[vizgrid] (0,-8.04) -- (0,0.05);
\node[viztick,anchor=north] at (0,-8.12) {0\%};
\draw[vizgrid] (2,-8.04) -- (2,0.05);
\node[viztick,anchor=north] at (2,-8.12) {25\%};
\draw[vizgrid] (4,-8.04) -- (4,0.05);
\node[viztick,anchor=north] at (4,-8.12) {50\%};
\draw[vizgrid] (6,-8.04) -- (6,0.05);
\node[viztick,anchor=north] at (6,-8.12) {75\%};
\draw[vizgrid] (8,-8.04) -- (8,0.05);
\node[viztick,anchor=north] at (8,-8.12) {100\%};
\node[viztick,anchor=north] at (4,-8.46) {share of attempts with the goal reached (15 configurations)};
\draw[vizAxis,line width=0.8pt] (2.133,-0.15) -- (2.933,-0.15);
\filldraw[fill=vizC0,draw=white,line width=0.5pt] (2.667,-0.15) circle (0.058);
\filldraw[fill=vizC1,draw=white,line width=0.5pt] (2.075,-0.208) rectangle (2.191,-0.092);
\filldraw[fill=vizC2,draw=white,line width=0.5pt] (2.933,-0.077) -- (2.87,-0.194) -- (2.997,-0.194) -- cycle;
\node[vizlab,anchor=east] at (-0.12,-0.15) {travel-01};
\draw[vizAxis,line width=0.8pt] (1.333,-0.45) -- (2.667,-0.45);
\filldraw[fill=vizC0,draw=white,line width=0.5pt] (2.667,-0.45) circle (0.058);
\filldraw[fill=vizC1,draw=white,line width=0.5pt] (1.542,-0.508) rectangle (1.658,-0.392);
\filldraw[fill=vizC2,draw=white,line width=0.5pt] (1.333,-0.377) -- (1.27,-0.493) -- (1.397,-0.493) -- cycle;
\node[vizlab,anchor=east] at (-0.12,-0.45) {travel-02};
\draw[vizAxis,line width=0.8pt] (1.6,-0.75) -- (2.933,-0.75);
\filldraw[fill=vizC0,draw=white,line width=0.5pt] (1.6,-0.75) circle (0.058);
\filldraw[fill=vizC1,draw=white,line width=0.5pt] (2.075,-0.808) rectangle (2.191,-0.692);
\filldraw[fill=vizC2,draw=white,line width=0.5pt] (2.933,-0.677) -- (2.87,-0.793) -- (2.997,-0.793) -- cycle;
\node[vizlab,anchor=east] at (-0.12,-0.75) {travel-03};
\draw[vizAxis,line width=0.8pt] (2.667,-1.17) -- (3.2,-1.17);
\filldraw[fill=vizC0,draw=white,line width=0.5pt] (3.2,-1.17) circle (0.058);
\filldraw[fill=vizC1,draw=white,line width=0.5pt] (2.609,-1.228) rectangle (2.725,-1.112);
\node[vizlab,anchor=east] at (-0.12,-1.17) {personal-01};
\draw[vizAxis,line width=0.8pt] (3.733,-1.47) -- (4.267,-1.47);
\filldraw[fill=vizC0,draw=white,line width=0.5pt] (3.733,-1.47) circle (0.058);
\filldraw[fill=vizC1,draw=white,line width=0.5pt] (4.209,-1.528) rectangle (4.325,-1.412);
\node[vizlab,anchor=east] at (-0.12,-1.47) {personal-02};
\draw[vizAxis,line width=0.8pt] (0.533,-1.77) -- (2.667,-1.77);
\filldraw[fill=vizC0,draw=white,line width=0.5pt] (2.667,-1.77) circle (0.058);
\filldraw[fill=vizC1,draw=white,line width=0.5pt] (0.475,-1.828) rectangle (0.591,-1.712);
\node[vizlab,anchor=east] at (-0.12,-1.77) {personal-03};
\draw[vizAxis,line width=0.8pt] (1.067,-2.19) -- (4.267,-2.19);
\filldraw[fill=vizC0,draw=white,line width=0.5pt] (4.267,-2.19) circle (0.058);
\filldraw[fill=vizC1,draw=white,line width=0.5pt] (1.009,-2.248) rectangle (1.125,-2.132);
\node[vizlab,anchor=east] at (-0.12,-2.19) {inquiry-01};
\draw[vizAxis,line width=0.8pt] (1.6,-2.49) -- (3.2,-2.49);
\filldraw[fill=vizC0,draw=white,line width=0.5pt] (3.2,-2.49) circle (0.058);
\filldraw[fill=vizC1,draw=white,line width=0.5pt] (1.542,-2.548) rectangle (1.658,-2.432);
\node[vizlab,anchor=east] at (-0.12,-2.49) {inquiry-02};
\draw[vizAxis,line width=0.8pt] (0.533,-2.79) -- (1.6,-2.79);
\filldraw[fill=vizC0,draw=white,line width=0.5pt] (1.6,-2.79) circle (0.058);
\filldraw[fill=vizC1,draw=white,line width=0.5pt] (0.475,-2.848) rectangle (0.591,-2.732);
\node[vizlab,anchor=east] at (-0.12,-2.79) {inquiry-03};
\draw[vizAxis,line width=0.8pt] (1.333,-3.21) -- (3.733,-3.21);
\filldraw[fill=vizC0,draw=white,line width=0.5pt] (3.733,-3.21) circle (0.058);
\filldraw[fill=vizC1,draw=white,line width=0.5pt] (2.609,-3.268) rectangle (2.725,-3.152);
\filldraw[fill=vizC2,draw=white,line width=0.5pt] (1.333,-3.137) -- (1.27,-3.253) -- (1.397,-3.253) -- cycle;
\node[vizlab,anchor=east] at (-0.12,-3.21) {service-01};
\draw[vizAxis,line width=0.8pt] (0.533,-3.51) -- (1.067,-3.51);
\filldraw[fill=vizC0,draw=white,line width=0.5pt] (1.067,-3.51) circle (0.058);
\filldraw[fill=vizC1,draw=white,line width=0.5pt] (1.009,-3.568) rectangle (1.125,-3.452);
\filldraw[fill=vizC2,draw=white,line width=0.5pt] (0.533,-3.437) -- (0.47,-3.553) -- (0.597,-3.553) -- cycle;
\node[vizlab,anchor=east] at (-0.12,-3.51) {service-02};
\draw[vizAxis,line width=0.8pt] (0.8,-3.81) -- (3.733,-3.81);
\filldraw[fill=vizC0,draw=white,line width=0.5pt] (3.733,-3.81) circle (0.058);
\filldraw[fill=vizC1,draw=white,line width=0.5pt] (1.542,-3.868) rectangle (1.658,-3.752);
\filldraw[fill=vizC2,draw=white,line width=0.5pt] (0.8,-3.737) -- (0.736,-3.853) -- (0.864,-3.853) -- cycle;
\node[vizlab,anchor=east] at (-0.12,-3.81) {service-03};
\draw[vizAxis,line width=0.8pt] (2.133,-4.23) -- (5.333,-4.23);
\filldraw[fill=vizC0,draw=white,line width=0.5pt] (5.333,-4.23) circle (0.058);
\filldraw[fill=vizC1,draw=white,line width=0.5pt] (2.075,-4.288) rectangle (2.191,-4.172);
\filldraw[fill=vizC2,draw=white,line width=0.5pt] (2.667,-4.157) -- (2.603,-4.273) -- (2.73,-4.273) -- cycle;
\node[vizlab,anchor=east] at (-0.12,-4.23) {finance-01};
\draw[vizAxis,line width=0.8pt] (1.333,-4.53) -- (3.2,-4.53);
\filldraw[fill=vizC0,draw=white,line width=0.5pt] (3.2,-4.53) circle (0.058);
\filldraw[fill=vizC1,draw=white,line width=0.5pt] (1.542,-4.588) rectangle (1.658,-4.472);
\filldraw[fill=vizC2,draw=white,line width=0.5pt] (1.333,-4.457) -- (1.27,-4.573) -- (1.397,-4.573) -- cycle;
\node[vizlab,anchor=east] at (-0.12,-4.53) {finance-02};
\draw[vizAxis,line width=0.8pt] (1.6,-4.83) -- (5.333,-4.83);
\filldraw[fill=vizC0,draw=white,line width=0.5pt] (5.333,-4.83) circle (0.058);
\filldraw[fill=vizC1,draw=white,line width=0.5pt] (1.542,-4.888) rectangle (1.658,-4.772);
\filldraw[fill=vizC2,draw=white,line width=0.5pt] (2.667,-4.757) -- (2.603,-4.873) -- (2.73,-4.873) -- cycle;
\node[vizlab,anchor=east] at (-0.12,-4.83) {finance-03};
\draw[vizAxis,line width=0.8pt] (1.067,-5.25) -- (2.133,-5.25);
\filldraw[fill=vizC0,draw=white,line width=0.5pt] (2.133,-5.25) circle (0.058);
\filldraw[fill=vizC1,draw=white,line width=0.5pt] (1.009,-5.308) rectangle (1.125,-5.192);
\node[vizlab,anchor=east] at (-0.12,-5.25) {office-01};
\draw[vizAxis,line width=0.8pt] (3.733,-5.55) -- (3.733,-5.55);
\filldraw[fill=vizC0,draw=white,line width=0.5pt] (3.733,-5.55) circle (0.058);
\filldraw[fill=vizC1,draw=white,line width=0.5pt] (3.675,-5.608) rectangle (3.791,-5.492);
\node[vizlab,anchor=east] at (-0.12,-5.55) {office-02};
\draw[vizAxis,line width=0.8pt] (2.667,-5.85) -- (3.2,-5.85);
\filldraw[fill=vizC0,draw=white,line width=0.5pt] (2.667,-5.85) circle (0.058);
\filldraw[fill=vizC1,draw=white,line width=0.5pt] (3.142,-5.908) rectangle (3.258,-5.792);
\node[vizlab,anchor=east] at (-0.12,-5.85) {office-03};
\draw[vizAxis,line width=0.8pt] (0.533,-6.27) -- (2.133,-6.27);
\filldraw[fill=vizC0,draw=white,line width=0.5pt] (1.067,-6.27) circle (0.058);
\filldraw[fill=vizC1,draw=white,line width=0.5pt] (2.075,-6.328) rectangle (2.191,-6.212);
\filldraw[fill=vizC2,draw=white,line width=0.5pt] (0.533,-6.197) -- (0.47,-6.313) -- (0.597,-6.313) -- cycle;
\node[vizlab,anchor=east] at (-0.12,-6.27) {software-01};
\draw[vizAxis,line width=0.8pt] (1.6,-6.57) -- (4.267,-6.57);
\filldraw[fill=vizC0,draw=white,line width=0.5pt] (2.667,-6.57) circle (0.058);
\filldraw[fill=vizC1,draw=white,line width=0.5pt] (4.209,-6.628) rectangle (4.325,-6.512);
\filldraw[fill=vizC2,draw=white,line width=0.5pt] (1.6,-6.497) -- (1.536,-6.613) -- (1.664,-6.613) -- cycle;
\node[vizlab,anchor=east] at (-0.12,-6.57) {software-02};
\draw[vizAxis,line width=0.8pt] (2.933,-6.87) -- (4.8,-6.87);
\filldraw[fill=vizC0,draw=white,line width=0.5pt] (4.267,-6.87) circle (0.058);
\filldraw[fill=vizC1,draw=white,line width=0.5pt] (4.742,-6.928) rectangle (4.858,-6.812);
\filldraw[fill=vizC2,draw=white,line width=0.5pt] (2.933,-6.797) -- (2.87,-6.913) -- (2.997,-6.913) -- cycle;
\node[vizlab,anchor=east] at (-0.12,-6.87) {software-03};
\draw[vizAxis,line width=0.8pt] (0.533,-7.29) -- (0.533,-7.29);
\filldraw[fill=vizC0,draw=white,line width=0.5pt] (0.533,-7.29) circle (0.058);
\filldraw[fill=vizC1,draw=white,line width=0.5pt] (0.475,-7.348) rectangle (0.591,-7.232);
\node[vizlab,anchor=east] at (-0.12,-7.29) {data-01};
\draw[vizAxis,line width=0.8pt] (2.133,-7.59) -- (3.2,-7.59);
\filldraw[fill=vizC0,draw=white,line width=0.5pt] (2.133,-7.59) circle (0.058);
\filldraw[fill=vizC1,draw=white,line width=0.5pt] (3.142,-7.648) rectangle (3.258,-7.532);
\node[vizlab,anchor=east] at (-0.12,-7.59) {data-02};
\draw[vizAxis,line width=0.8pt] (1.6,-7.89) -- (1.6,-7.89);
\filldraw[fill=vizC0,draw=white,line width=0.5pt] (1.6,-7.89) circle (0.058);
\filldraw[fill=vizC1,draw=white,line width=0.5pt] (1.542,-7.948) rectangle (1.658,-7.832);
\node[vizlab,anchor=east] at (-0.12,-7.89) {data-03};
\node[circle,fill=vizC0,minimum size=0.13cm,inner sep=0,anchor=west] (lgs1) at (0,0.35) {};
\node[viztick,anchor=west,xshift=0.06cm] (lgl1) at (lgs1.east) {C0};
\node[fill=vizC1,minimum size=0.12cm,inner sep=0,anchor=west,xshift=0.3cm] (lgs2) at (lgl1.east) {};
\node[viztick,anchor=west,xshift=0.06cm] (lgl2) at (lgs2.east) {C1};
\node[regular polygon,regular polygon sides=3,fill=vizC2,minimum size=0.18cm,inner sep=0,anchor=west,xshift=0.3cm] (lgs3) at (lgl2.east) {};
\node[viztick,anchor=west,xshift=0.06cm] (lgl3) at (lgs3.east) {C2 (both worlds)};
\end{tikzpicture}
\caption{Task difficulty: share of the fifteen configurations reaching the goal of each task in C0 and in C1, and share of the 30 C2 episodes of the task (two worlds each).}\label{fig:families}
\end{figure}

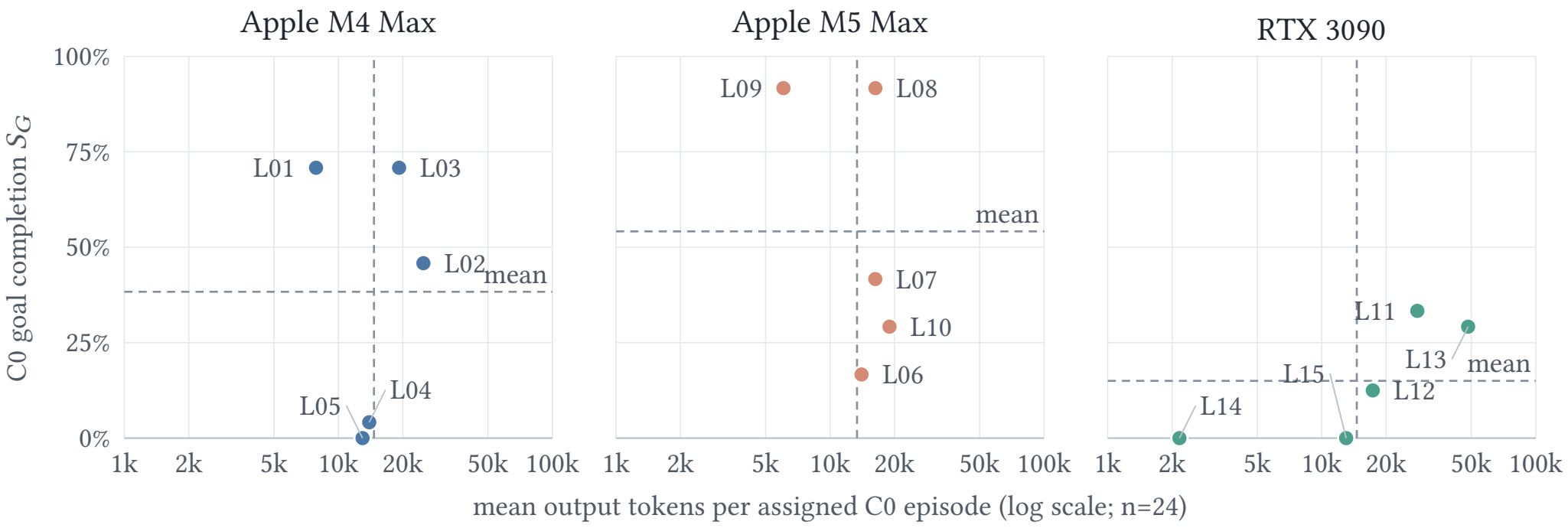
\begin{figure}[tbp]
\centering
\begin{tikzpicture}[x=1cm,y=1cm,line cap=butt]
\draw[vizgrid] (0,0) -- (3.7,0);
\node[viztick,anchor=east] at (-0.08,0) {0\%};
\draw[vizgrid] (0,0.825) -- (3.7,0.825);
\node[viztick,anchor=east] at (-0.08,0.825) {25\%};
\draw[vizgrid] (0,1.65) -- (3.7,1.65);
\node[viztick,anchor=east] at (-0.08,1.65) {50\%};
\draw[vizgrid] (0,2.475) -- (3.7,2.475);
\node[viztick,anchor=east] at (-0.08,2.475) {75\%};
\draw[vizgrid] (0,3.3) -- (3.7,3.3);
\node[viztick,anchor=east] at (-0.08,3.3) {100\%};
\draw[vizgrid] (0,0) -- (0,3.3);
\node[viztick,anchor=north] at (0,-0.08) {1k};
\draw[vizgrid] (0.557,0) -- (0.557,3.3);
\node[viztick,anchor=north] at (0.557,-0.08) {2k};
\draw[vizgrid] (1.293,0) -- (1.293,3.3);
\node[viztick,anchor=north] at (1.293,-0.08) {5k};
\draw[vizgrid] (1.85,0) -- (1.85,3.3);
\node[viztick,anchor=north] at (1.85,-0.08) {10k};
\draw[vizgrid] (2.407,0) -- (2.407,3.3);
\node[viztick,anchor=north] at (2.407,-0.08) {20k};
\draw[vizgrid] (3.143,0) -- (3.143,3.3);
\node[viztick,anchor=north] at (3.143,-0.08) {50k};
\draw[vizgrid] (3.7,0) -- (3.7,3.3);
\node[viztick,anchor=north] at (3.7,-0.08) {100k};
\draw[vizaxis] (0,0) -- (3.7,0);
\draw[vizMuted,line width=0.5pt,dash pattern=on 2pt off 1.5pt] (0,1.265) -- (3.7,1.265) (2.158,0) -- (2.158,3.3);
\node[viztick,anchor=south east] at (3.7,1.295) {mean};
\node[viztitle,anchor=south] at (1.85,3.4) {Apple M4 Max};
\filldraw[fill=vizP1,draw=white,line width=0.5pt] (1.657,2.337) circle (0.07);
\node[viztick,anchor=east] at (1.517,2.337) {L01};
\filldraw[fill=vizP1,draw=white,line width=0.5pt] (2.375,2.337) circle (0.07);
\node[viztick,anchor=west] at (2.515,2.337) {L03};
\filldraw[fill=vizP1,draw=white,line width=0.5pt] (2.585,1.512) circle (0.07);
\node[viztick,anchor=west] at (2.725,1.512) {L02};
\filldraw[fill=vizP1,draw=white,line width=0.5pt] (2.117,0.137) circle (0.07);
\draw[vizAxis,line width=0.35pt] (2.117,0.137) -- (2.257,0.417);
\node[viztick,anchor=west] at (2.257,0.417) {L04};
\filldraw[fill=vizP1,draw=white,line width=0.5pt] (2.059,0) circle (0.07);
\draw[vizAxis,line width=0.35pt] (2.059,0) -- (1.919,0.28);
\node[viztick,anchor=east] at (1.919,0.28) {L05};
\draw[vizgrid] (4.25,0) -- (7.95,0);
\draw[vizgrid] (4.25,0.825) -- (7.95,0.825);
\draw[vizgrid] (4.25,1.65) -- (7.95,1.65);
\draw[vizgrid] (4.25,2.475) -- (7.95,2.475);
\draw[vizgrid] (4.25,3.3) -- (7.95,3.3);
\draw[vizgrid] (4.25,0) -- (4.25,3.3);
\node[viztick,anchor=north] at (4.25,-0.08) {1k};
\draw[vizgrid] (4.807,0) -- (4.807,3.3);
\node[viztick,anchor=north] at (4.807,-0.08) {2k};
\draw[vizgrid] (5.543,0) -- (5.543,3.3);
\node[viztick,anchor=north] at (5.543,-0.08) {5k};
\draw[vizgrid] (6.1,0) -- (6.1,3.3);
\node[viztick,anchor=north] at (6.1,-0.08) {10k};
\draw[vizgrid] (6.657,0) -- (6.657,3.3);
\node[viztick,anchor=north] at (6.657,-0.08) {20k};
\draw[vizgrid] (7.393,0) -- (7.393,3.3);
\node[viztick,anchor=north] at (7.393,-0.08) {50k};
\draw[vizgrid] (7.95,0) -- (7.95,3.3);
\node[viztick,anchor=north] at (7.95,-0.08) {100k};
\draw[vizaxis] (4.25,0) -- (7.95,0);
\draw[vizMuted,line width=0.5pt,dash pattern=on 2pt off 1.5pt] (4.25,1.787) -- (7.95,1.787) (6.333,0) -- (6.333,3.3);
\node[viztick,anchor=south east] at (7.95,1.817) {mean};
\node[viztitle,anchor=south] at (6.1,3.4) {Apple M5 Max};
\filldraw[fill=vizP2,draw=white,line width=0.5pt] (5.697,3.025) circle (0.07);
\node[viztick,anchor=east] at (5.557,3.025) {L09};
\filldraw[fill=vizP2,draw=white,line width=0.5pt] (6.492,3.025) circle (0.07);
\node[viztick,anchor=west] at (6.632,3.025) {L08};
\filldraw[fill=vizP2,draw=white,line width=0.5pt] (6.491,1.375) circle (0.07);
\node[viztick,anchor=west] at (6.631,1.375) {L07};
\filldraw[fill=vizP2,draw=white,line width=0.5pt] (6.614,0.963) circle (0.07);
\node[viztick,anchor=west] at (6.754,0.963) {L10};
\filldraw[fill=vizP2,draw=white,line width=0.5pt] (6.372,0.55) circle (0.07);
\node[viztick,anchor=west] at (6.512,0.55) {L06};
\draw[vizgrid] (8.5,0) -- (12.2,0);
\draw[vizgrid] (8.5,0.825) -- (12.2,0.825);
\draw[vizgrid] (8.5,1.65) -- (12.2,1.65);
\draw[vizgrid] (8.5,2.475) -- (12.2,2.475);
\draw[vizgrid] (8.5,3.3) -- (12.2,3.3);
\draw[vizgrid] (8.5,0) -- (8.5,3.3);
\node[viztick,anchor=north] at (8.5,-0.08) {1k};
\draw[vizgrid] (9.057,0) -- (9.057,3.3);
\node[viztick,anchor=north] at (9.057,-0.08) {2k};
\draw[vizgrid] (9.793,0) -- (9.793,3.3);
\node[viztick,anchor=north] at (9.793,-0.08) {5k};
\draw[vizgrid] (10.35,0) -- (10.35,3.3);
\node[viztick,anchor=north] at (10.35,-0.08) {10k};
\draw[vizgrid] (10.907,0) -- (10.907,3.3);
\node[viztick,anchor=north] at (10.907,-0.08) {20k};
\draw[vizgrid] (11.643,0) -- (11.643,3.3);
\node[viztick,anchor=north] at (11.643,-0.08) {50k};
\draw[vizgrid] (12.2,0) -- (12.2,3.3);
\node[viztick,anchor=north] at (12.2,-0.08) {100k};
\draw[vizaxis] (8.5,0) -- (12.2,0);
\draw[vizMuted,line width=0.5pt,dash pattern=on 2pt off 1.5pt] (8.5,0.495) -- (12.2,0.495) (10.653,0) -- (10.653,3.3);
\node[viztick,anchor=south east] at (12.2,0.525) {mean};
\node[viztitle,anchor=south] at (10.35,3.4) {RTX 3090};
\filldraw[fill=vizP3,draw=white,line width=0.5pt] (11.176,1.1) circle (0.07);
\node[viztick,anchor=east] at (11.036,1.1) {L11};
\filldraw[fill=vizP3,draw=white,line width=0.5pt] (11.615,0.963) circle (0.07);
\draw[vizAxis,line width=0.35pt] (11.615,0.963) -- (11.475,0.682);
\node[viztick,anchor=east] at (11.475,0.682) {L13};
\filldraw[fill=vizP3,draw=white,line width=0.5pt] (10.791,0.412) circle (0.07);
\node[viztick,anchor=west] at (10.931,0.412) {L12};
\filldraw[fill=vizP3,draw=white,line width=0.5pt] (9.12,0) circle (0.07);
\draw[vizAxis,line width=0.35pt] (9.12,0) -- (9.26,0.28);
\node[viztick,anchor=west] at (9.26,0.28) {L14};
\filldraw[fill=vizP3,draw=white,line width=0.5pt] (10.561,0) circle (0.07);
\draw[vizAxis,line width=0.35pt] (10.561,0) -- (10.421,0.56);
\node[viztick,anchor=east] at (10.421,0.56) {L15};
\node[viztick,anchor=south,rotate=90] at (-0.75,1.65) {C0 goal completion $S_G$};
\node[viztick,anchor=north] at (6.1,-0.45) {mean output tokens per assigned C0 episode (log scale; n=24)};
\end{tikzpicture}
\caption{Exploratory C0 quality--output view. Both axes use the same 24 assigned baseline episodes, with failures included in arithmetic mean output tokens. Panels separate actual platforms; compare within a panel. Runtime-reported tokens are model-specific units, not equal compute or total cost. Dashed lines mark within-platform means (geometric on the logarithmic output axis).}\label{fig:tokens-score}
\end{figure}

\begin{figure}[tbp]
\centering
\begin{tikzpicture}[x=1cm,y=1cm,line cap=butt]
\draw[vizgrid] (0,-7.14) -- (0,0.05);
\node[viztick,anchor=north] at (0,-7.22) {0k};
\draw[vizgrid] (1.8,-7.14) -- (1.8,0.05);
\node[viztick,anchor=north] at (1.8,-7.22) {100k};
\draw[vizgrid] (3.6,-7.14) -- (3.6,0.05);
\node[viztick,anchor=north] at (3.6,-7.22) {200k};
\draw[vizgrid] (5.4,-7.14) -- (5.4,0.05);
\node[viztick,anchor=north] at (5.4,-7.22) {300k};
\draw[vizgrid] (7.2,-7.14) -- (7.2,0.05);
\node[viztick,anchor=north] at (7.2,-7.22) {400k};
\node[viztick,anchor=north] at (3.6,-7.58) {C0 output tokens per goal reached (all 24 assigned episodes)};
\node[viztitle,anchor=east] at (-0.12,-0.19) {\textit{Apple M4 Max}};
\node[vizlab,anchor=east] at (-0.12,-0.54) {L01 qwen3.6 35b-a3b};
\node[vizlab,anchor=east] at (-0.12,-0.94) {L03 granite4.2 30b};
\node[vizlab,anchor=east] at (-0.12,-1.34) {L02 gemma4 26b-a4b};
\node[vizlab,anchor=east] at (-0.12,-1.74) {L04 lfm2.5 24b-a2b};
\node[vizlab,anchor=east] at (-0.12,-2.14) {L05 north-mini-code-1.0};
\node[viztitle,anchor=east] at (-0.12,-2.57) {\textit{Apple M5 Max}};
\node[vizlab,anchor=east] at (-0.12,-2.92) {L09 qwen3.6 27b-coding-mtp};
\node[vizlab,anchor=east] at (-0.12,-3.32) {L08 gemma4 31b};
\node[vizlab,anchor=east] at (-0.12,-3.72) {L07 nemotron-3.5-lightning 30b-a3b};
\node[vizlab,anchor=east] at (-0.12,-4.12) {L10 gemma4 12b};
\node[vizlab,anchor=east] at (-0.12,-4.52) {L06 glm-4.7-flash};
\node[viztitle,anchor=east] at (-0.12,-4.95) {\textit{RTX 3090}};
\node[vizlab,anchor=east] at (-0.12,-5.3) {L11 qwen3.5 9b};
\node[vizlab,anchor=east] at (-0.12,-5.7) {L13 granite4.2 8b};
\node[vizlab,anchor=east] at (-0.12,-6.1) {L12 gemma4 e4b};
\node[vizlab,anchor=east] at (-0.12,-6.5) {L14 nemotron-3-nano 4b};
\node[vizlab,anchor=east] at (-0.12,-6.9) {L15 lfm2.5 8b-a1b};
\node[viztick,anchor=south west] at (8.15,0.2) {Goals reached};
\fill[vizP1] (0,-0.66) [rounded corners=0.05cm] -- (0.2,-0.66) -- (0.2,-0.42) [sharp corners] -- (0,-0.42) -- cycle;
\node[viztick,anchor=west] at (0.3,-0.54) {11.1k};
\node[viztick,anchor=west] at (8.15,-0.54) {17/24};
\fill[vizP1] (0,-1.06) [rounded corners=0.05cm] -- (0.488,-1.06) -- (0.488,-0.82) [sharp corners] -- (0,-0.82) -- cycle;
\node[viztick,anchor=west] at (0.588,-0.94) {27.1k};
\node[viztick,anchor=west] at (8.15,-0.94) {17/24};
\fill[vizP1] (0,-1.46) [rounded corners=0.05cm] -- (0.98,-1.46) -- (0.98,-1.22) [sharp corners] -- (0,-1.22) -- cycle;
\node[viztick,anchor=west] at (1.08,-1.34) {54.5k};
\node[viztick,anchor=west] at (8.15,-1.34) {11/24};
\fill[vizP1] (0,-1.86) [rounded corners=0.05cm] -- (6.021,-1.86) -- (6.021,-1.62) [sharp corners] -- (0,-1.62) -- cycle;
\node[viztick,anchor=west] at (6.121,-1.74) {334.5k};
\node[viztick,anchor=west] at (8.15,-1.74) {1/24};
\node[viztick,anchor=west] at (0.05,-2.14) {-- (no goal reached)};
\node[viztick,anchor=west] at (8.15,-2.14) {0/24};
\fill[vizP2] (0,-3.04) [rounded corners=0.05cm] -- (0.119,-3.04) -- (0.119,-2.8) [sharp corners] -- (0,-2.8) -- cycle;
\node[viztick,anchor=west] at (0.219,-2.92) {6.6k};
\node[viztick,anchor=west] at (8.15,-2.92) {22/24};
\fill[vizP2] (0,-3.44) [rounded corners=0.05cm] -- (0.32,-3.44) -- (0.32,-3.2) [sharp corners] -- (0,-3.2) -- cycle;
\node[viztick,anchor=west] at (0.42,-3.32) {17.8k};
\node[viztick,anchor=west] at (8.15,-3.32) {22/24};
\fill[vizP2] (0,-3.84) [rounded corners=0.05cm] -- (0.702,-3.84) -- (0.702,-3.6) [sharp corners] -- (0,-3.6) -- cycle;
\node[viztick,anchor=west] at (0.802,-3.72) {39.0k};
\node[viztick,anchor=west] at (8.15,-3.72) {10/24};
\fill[vizP2] (0,-4.24) [rounded corners=0.05cm] -- (1.17,-4.24) -- (1.17,-4) [sharp corners] -- (0,-4) -- cycle;
\node[viztick,anchor=west] at (1.27,-4.12) {65.0k};
\node[viztick,anchor=west] at (8.15,-4.12) {7/24};
\fill[vizP2] (0,-4.64) [rounded corners=0.05cm] -- (1.515,-4.64) -- (1.515,-4.4) [sharp corners] -- (0,-4.4) -- cycle;
\node[viztick,anchor=west] at (1.615,-4.52) {84.2k};
\node[viztick,anchor=west] at (8.15,-4.52) {4/24};
\fill[vizP3] (0,-5.42) [rounded corners=0.05cm] -- (1.51,-5.42) -- (1.51,-5.18) [sharp corners] -- (0,-5.18) -- cycle;
\node[viztick,anchor=west] at (1.61,-5.3) {83.9k};
\node[viztick,anchor=west] at (8.15,-5.3) {8/24};
\fill[vizP3] (0,-5.82) [rounded corners=0.05cm] -- (2.981,-5.82) -- (2.981,-5.58) [sharp corners] -- (0,-5.58) -- cycle;
\node[viztick,anchor=west] at (3.081,-5.7) {165.6k};
\node[viztick,anchor=west] at (8.15,-5.7) {7/24};
\fill[vizP3] (0,-6.22) [rounded corners=0.05cm] -- (2.494,-6.22) -- (2.494,-5.98) [sharp corners] -- (0,-5.98) -- cycle;
\node[viztick,anchor=west] at (2.594,-6.1) {138.6k};
\node[viztick,anchor=west] at (8.15,-6.1) {3/24};
\node[viztick,anchor=west] at (0.05,-6.5) {-- (no goal reached)};
\node[viztick,anchor=west] at (8.15,-6.5) {0/24};
\node[viztick,anchor=west] at (0.05,-6.9) {-- (no goal reached)};
\node[viztick,anchor=west] at (8.15,-6.9) {0/24};
\end{tikzpicture}
\caption{Total C0 output tokens divided by goals reached, with all 24 assigned episodes contributing output usage. Zero-goal configurations have an undefined ratio, marked by a dash rather than a zero bar. Exact goal counts are shown at right. Grouping separates actual platforms; values are output-only accounting, not equal compute, energy, or monetary cost.}\label{fig:c0-output-per-goal}
\end{figure}
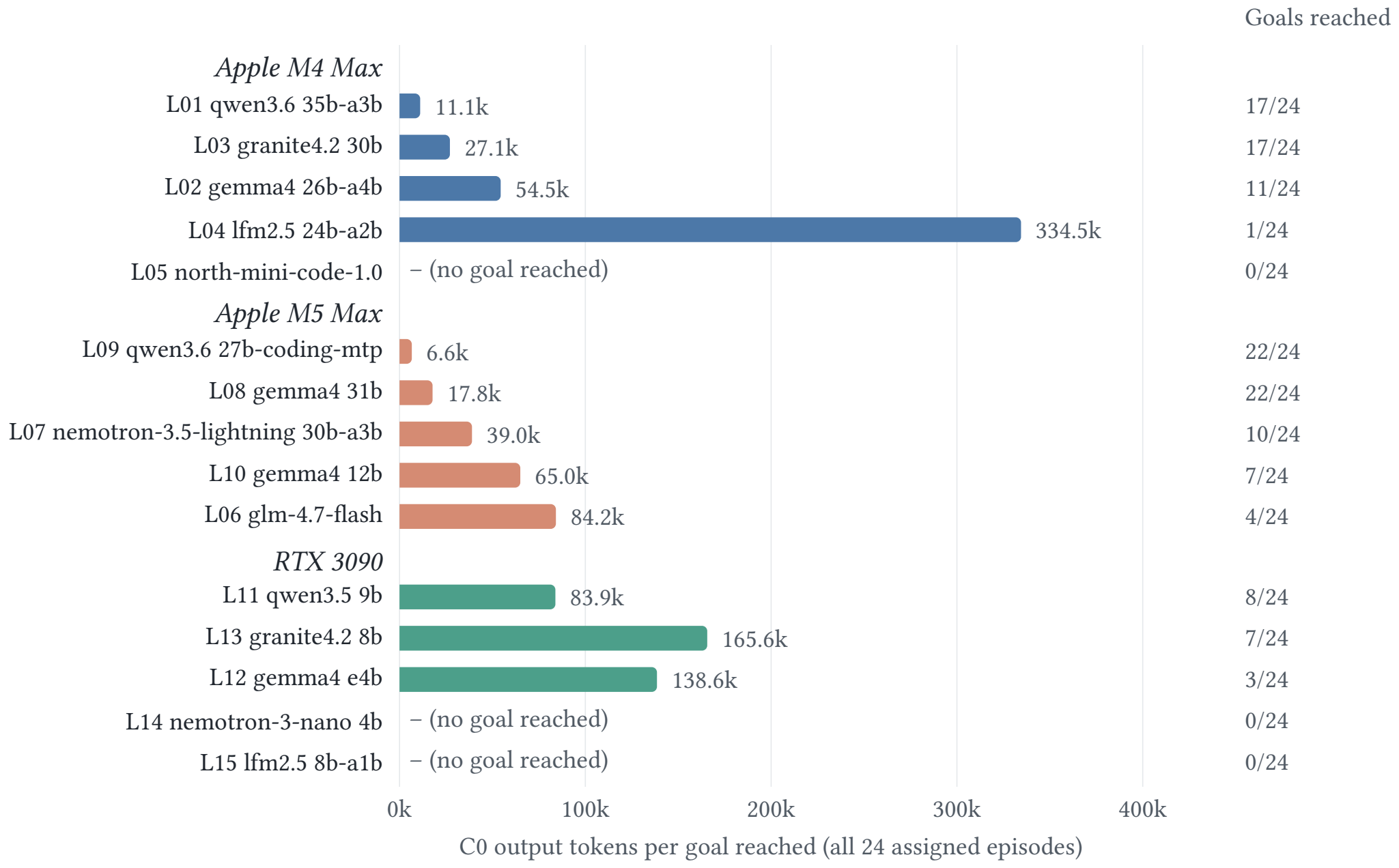

\begin{figure}[tbp]
\centering
\input{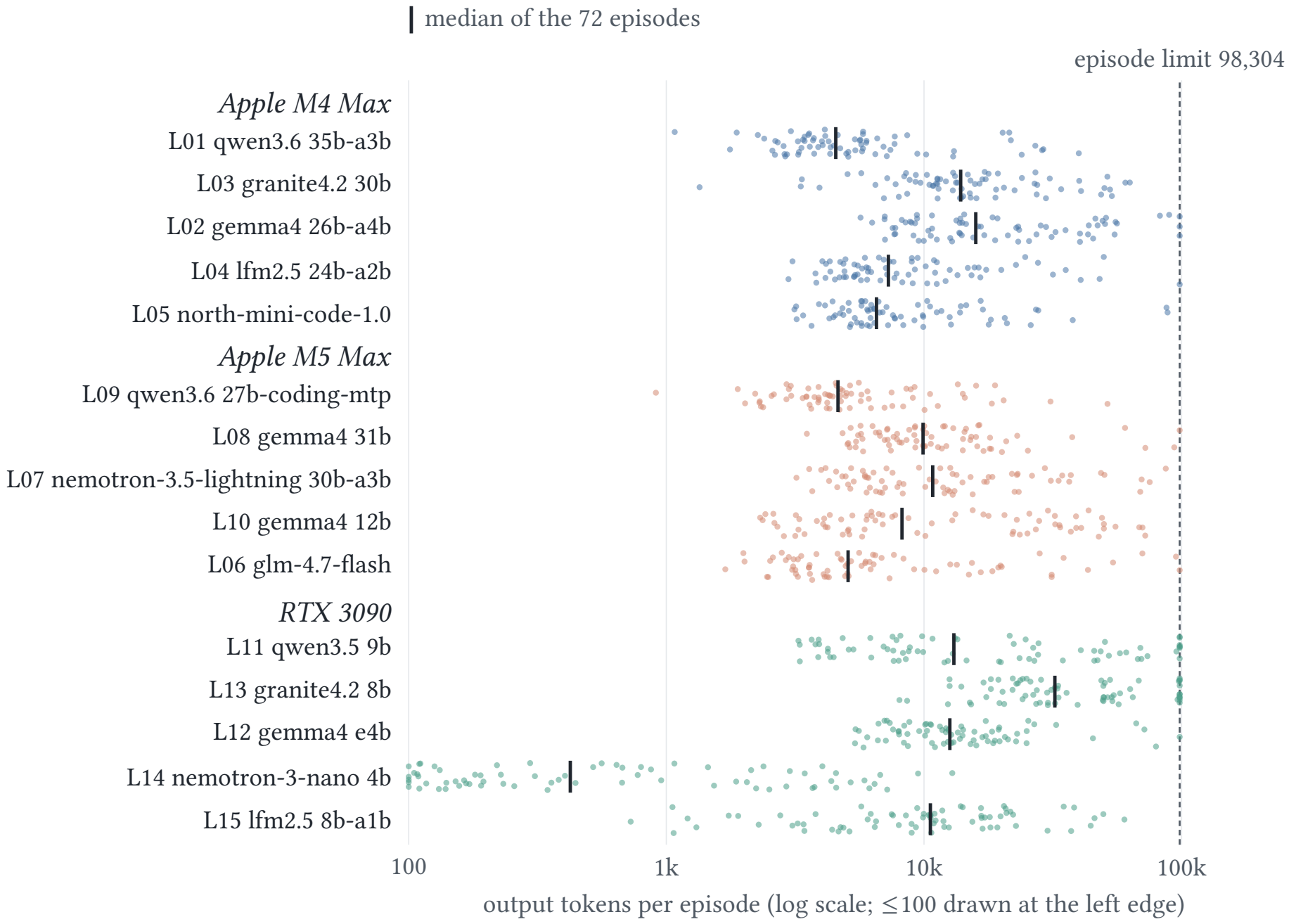}
\caption{Output tokens of every episode (dots, all conditions) with the median, grouped by platform. The dashed line is the frozen per-episode output limit.}\label{fig:tokens}
\end{figure}

\begin{figure}[tbp]
\centering
\begin{tikzpicture}[x=1cm,y=1cm,line cap=butt]
\draw[vizgrid] (0,-7.14) -- (0,0.05);
\node[viztick,anchor=north] at (0,-7.22) {0};
\draw[vizgrid] (0.88,-7.14) -- (0.88,0.05);
\node[viztick,anchor=north] at (0.88,-7.22) {10};
\draw[vizgrid] (1.76,-7.14) -- (1.76,0.05);
\node[viztick,anchor=north] at (1.76,-7.22) {20};
\draw[vizgrid] (2.64,-7.14) -- (2.64,0.05);
\node[viztick,anchor=north] at (2.64,-7.22) {30};
\draw[vizgrid] (3.52,-7.14) -- (3.52,0.05);
\node[viztick,anchor=north] at (3.52,-7.22) {40};
\draw[vizgrid] (4.4,-7.14) -- (4.4,0.05);
\node[viztick,anchor=north] at (4.4,-7.22) {50};
\node[viztick,anchor=north] at (2.2,-7.56) {model requests per episode};
\draw[vizInk2,line width=0.5pt,dash pattern=on 1.5pt off 1pt] (4.224,-7.14) -- (4.224,0.05);
\node[viztick,anchor=south] at (4.224,0.08) {limit 48};
\draw[vizInk2,line width=0.4pt] (0.352,-0.54) -- (3.432,-0.54);
\fill[vizP1] (0.616,-0.63) rectangle (0.88,-0.45);
\draw[vizInk,line width=0.9pt] (0.704,-0.67) -- (0.704,-0.41);
\draw[vizInk2,line width=0.4pt] (0.44,-0.94) -- (2.112,-0.94);
\fill[vizP1] (0.792,-1.03) rectangle (1.144,-0.85);
\draw[vizInk,line width=0.9pt] (0.968,-1.07) -- (0.968,-0.81);
\draw[vizInk2,line width=0.4pt] (0.528,-1.34) -- (2.288,-1.34);
\fill[vizP1] (1.034,-1.43) rectangle (1.496,-1.25);
\draw[vizInk,line width=0.9pt] (1.232,-1.47) -- (1.232,-1.21);
\draw[vizInk2,line width=0.4pt] (0.44,-1.74) -- (2.728,-1.74);
\fill[vizP1] (0.594,-1.83) rectangle (1.056,-1.65);
\draw[vizInk,line width=0.9pt] (0.704,-1.87) -- (0.704,-1.61);
\draw[vizInk2,line width=0.4pt] (0.528,-2.14) -- (4.312,-2.14);
\fill[vizP1] (1.056,-2.23) rectangle (1.672,-2.05);
\draw[vizInk,line width=0.9pt] (1.32,-2.27) -- (1.32,-2.01);
\draw[vizInk2,line width=0.4pt] (0.44,-2.92) -- (1.672,-2.92);
\fill[vizP2] (0.616,-3.01) rectangle (0.792,-2.83);
\draw[vizInk,line width=0.9pt] (0.704,-3.05) -- (0.704,-2.79);
\draw[vizInk2,line width=0.4pt] (0.44,-3.32) -- (1.76,-3.32);
\fill[vizP2] (0.704,-3.41) rectangle (1.232,-3.23);
\draw[vizInk,line width=0.9pt] (0.88,-3.45) -- (0.88,-3.19);
\draw[vizInk2,line width=0.4pt] (0.44,-3.72) -- (4.048,-3.72);
\fill[vizP2] (0.858,-3.81) rectangle (1.32,-3.63);
\draw[vizInk,line width=0.9pt] (1.144,-3.85) -- (1.144,-3.59);
\draw[vizInk2,line width=0.4pt] (0.528,-4.12) -- (4.312,-4.12);
\fill[vizP2] (1.21,-4.21) rectangle (1.518,-4.03);
\draw[vizInk,line width=0.9pt] (1.32,-4.25) -- (1.32,-3.99);
\draw[vizInk2,line width=0.4pt] (0.44,-4.52) -- (4.312,-4.52);
\fill[vizP2] (0.704,-4.61) rectangle (1.43,-4.43);
\draw[vizInk,line width=0.9pt] (0.924,-4.65) -- (0.924,-4.39);
\draw[vizInk2,line width=0.4pt] (0.528,-5.3) -- (4.312,-5.3);
\fill[vizP3] (1.056,-5.39) rectangle (1.804,-5.21);
\draw[vizInk,line width=0.9pt] (1.364,-5.43) -- (1.364,-5.17);
\draw[vizInk2,line width=0.4pt] (0.528,-5.7) -- (2.728,-5.7);
\fill[vizP3] (1.056,-5.79) rectangle (1.408,-5.61);
\draw[vizInk,line width=0.9pt] (1.232,-5.83) -- (1.232,-5.57);
\draw[vizInk2,line width=0.4pt] (0.352,-6.1) -- (3.168,-6.1);
\fill[vizP3] (0.858,-6.19) rectangle (1.32,-6.01);
\draw[vizInk,line width=0.9pt] (1.056,-6.23) -- (1.056,-5.97);
\draw[vizInk2,line width=0.4pt] (0.088,-6.5) -- (1.232,-6.5);
\fill[vizP3] (0.176,-6.59) rectangle (0.462,-6.41);
\draw[vizInk,line width=0.9pt] (0.264,-6.63) -- (0.264,-6.37);
\draw[vizInk2,line width=0.4pt] (0.176,-6.9) -- (4.312,-6.9);
\fill[vizP3] (0.704,-6.99) rectangle (1.584,-6.81);
\draw[vizInk,line width=0.9pt] (1.012,-7.03) -- (1.012,-6.77);
\draw[vizgrid] (5.1,-7.14) -- (5.1,0.05);
\node[viztick,anchor=north] at (5.1,-7.22) {0};
\draw[vizgrid] (5.729,-7.14) -- (5.729,0.05);
\node[viztick,anchor=north] at (5.729,-7.22) {10};
\draw[vizgrid] (6.357,-7.14) -- (6.357,0.05);
\node[viztick,anchor=north] at (6.357,-7.22) {20};
\draw[vizgrid] (6.986,-7.14) -- (6.986,0.05);
\node[viztick,anchor=north] at (6.986,-7.22) {30};
\draw[vizgrid] (7.614,-7.14) -- (7.614,0.05);
\node[viztick,anchor=north] at (7.614,-7.22) {40};
\draw[vizgrid] (8.243,-7.14) -- (8.243,0.05);
\node[viztick,anchor=north] at (8.243,-7.22) {50};
\draw[vizgrid] (8.871,-7.14) -- (8.871,0.05);
\node[viztick,anchor=north] at (8.871,-7.22) {60};
\draw[vizgrid] (9.5,-7.14) -- (9.5,0.05);
\node[viztick,anchor=north] at (9.5,-7.22) {70};
\node[viztick,anchor=north] at (7.3,-7.56) {executed tool calls per episode};
\draw[vizInk2,line width=0.5pt,dash pattern=on 1.5pt off 1pt] (9.123,-7.14) -- (9.123,0.05);
\node[viztick,anchor=south] at (9.123,0.08) {limit 64};
\draw[vizInk2,line width=0.4pt] (5.477,-0.54) -- (8.557,-0.54);
\fill[vizP1] (5.964,-0.63) rectangle (6.169,-0.45);
\draw[vizInk,line width=0.9pt] (6.043,-0.67) -- (6.043,-0.41);
\draw[vizInk2,line width=0.4pt] (5.351,-0.94) -- (6.609,-0.94);
\fill[vizP1] (5.854,-1.03) rectangle (6.106,-0.85);
\draw[vizInk,line width=0.9pt] (5.98,-1.07) -- (5.98,-0.81);
\draw[vizInk2,line width=0.4pt] (5.289,-1.34) -- (6.609,-1.34);
\fill[vizP1] (5.791,-1.43) rectangle (6.121,-1.25);
\draw[vizInk,line width=0.9pt] (5.98,-1.47) -- (5.98,-1.21);
\draw[vizInk2,line width=0.4pt] (5.289,-1.74) -- (6.923,-1.74);
\fill[vizP1] (5.791,-1.83) rectangle (6.043,-1.65);
\draw[vizInk,line width=0.9pt] (5.98,-1.87) -- (5.98,-1.61);
\draw[vizInk2,line width=0.4pt] (5.414,-2.14) -- (7.866,-2.14);
\fill[vizP1] (5.854,-2.23) rectangle (6.247,-2.05);
\draw[vizInk,line width=0.9pt] (5.98,-2.27) -- (5.98,-2.01);
\draw[vizInk2,line width=0.4pt] (5.666,-2.92) -- (6.923,-2.92);
\fill[vizP2] (5.917,-3.01) rectangle (6.169,-2.83);
\draw[vizInk,line width=0.9pt] (6.043,-3.05) -- (6.043,-2.79);
\draw[vizInk2,line width=0.4pt] (5.477,-3.32) -- (6.357,-3.32);
\fill[vizP2] (5.854,-3.41) rectangle (6.106,-3.23);
\draw[vizInk,line width=0.9pt] (6.011,-3.45) -- (6.011,-3.19);
\draw[vizInk2,line width=0.4pt] (5.289,-3.72) -- (7.991,-3.72);
\fill[vizP2] (5.917,-3.81) rectangle (6.121,-3.63);
\draw[vizInk,line width=0.9pt] (6.043,-3.85) -- (6.043,-3.59);
\draw[vizInk2,line width=0.4pt] (5.289,-4.12) -- (8.054,-4.12);
\fill[vizP2] (5.854,-4.21) rectangle (6.169,-4.03);
\draw[vizInk,line width=0.9pt] (6.043,-4.25) -- (6.043,-3.99);
\draw[vizInk2,line width=0.4pt] (5.603,-4.52) -- (8.18,-4.52);
\fill[vizP2] (5.917,-4.61) rectangle (6.294,-4.43);
\draw[vizInk,line width=0.9pt] (6.043,-4.65) -- (6.043,-4.39);
\draw[vizInk2,line width=0.4pt] (5.666,-5.3) -- (8.494,-5.3);
\fill[vizP3] (5.98,-5.39) rectangle (6.373,-5.21);
\draw[vizInk,line width=0.9pt] (6.106,-5.43) -- (6.106,-5.17);
\draw[vizInk2,line width=0.4pt] (5.289,-5.7) -- (6.923,-5.7);
\fill[vizP3] (5.839,-5.79) rectangle (6.043,-5.61);
\draw[vizInk,line width=0.9pt] (5.917,-5.83) -- (5.917,-5.57);
\draw[vizInk2,line width=0.4pt] (5.414,-6.1) -- (7.363,-6.1);
\fill[vizP3] (5.729,-6.19) rectangle (6.043,-6.01);
\draw[vizInk,line width=0.9pt] (5.917,-6.23) -- (5.917,-5.97);
\draw[vizInk2,line width=0.4pt] (5.1,-6.5) -- (5.603,-6.5);
\fill[vizP3] (5.163,-6.59) rectangle (5.351,-6.41);
\draw[vizInk,line width=0.9pt] (5.226,-6.63) -- (5.226,-6.37);
\draw[vizInk2,line width=0.4pt] (5.163,-6.9) -- (8.117,-6.9);
\fill[vizP3] (5.603,-6.99) rectangle (6.106,-6.81);
\draw[vizInk,line width=0.9pt] (5.791,-7.03) -- (5.791,-6.77);
\node[viztitle,anchor=east] at (-0.12,-0.19) {\textit{Apple M4 Max}};
\node[vizlab,anchor=east] at (-0.12,-0.54) {L01 qwen3.6 35b-a3b};
\node[vizlab,anchor=east] at (-0.12,-0.94) {L03 granite4.2 30b};
\node[vizlab,anchor=east] at (-0.12,-1.34) {L02 gemma4 26b-a4b};
\node[vizlab,anchor=east] at (-0.12,-1.74) {L04 lfm2.5 24b-a2b};
\node[vizlab,anchor=east] at (-0.12,-2.14) {L05 north-mini-code-1.0};
\node[viztitle,anchor=east] at (-0.12,-2.57) {\textit{Apple M5 Max}};
\node[vizlab,anchor=east] at (-0.12,-2.92) {L09 qwen3.6 27b-coding-mtp};
\node[vizlab,anchor=east] at (-0.12,-3.32) {L08 gemma4 31b};
\node[vizlab,anchor=east] at (-0.12,-3.72) {L07 nemotron-3.5-lightning 30b-a3b};
\node[vizlab,anchor=east] at (-0.12,-4.12) {L10 gemma4 12b};
\node[vizlab,anchor=east] at (-0.12,-4.52) {L06 glm-4.7-flash};
\node[viztitle,anchor=east] at (-0.12,-4.95) {\textit{RTX 3090}};
\node[vizlab,anchor=east] at (-0.12,-5.3) {L11 qwen3.5 9b};
\node[vizlab,anchor=east] at (-0.12,-5.7) {L13 granite4.2 8b};
\node[vizlab,anchor=east] at (-0.12,-6.1) {L12 gemma4 e4b};
\node[vizlab,anchor=east] at (-0.12,-6.5) {L14 nemotron-3-nano 4b};
\node[vizlab,anchor=east] at (-0.12,-6.9) {L15 lfm2.5 8b-a1b};
\node[inner sep=0,anchor=west] (lgs1) at (0,-8.09) {\textcolor{vizInk}{\rule[-0.3ex]{0.9pt}{1.7ex}}};
\node[viztick,anchor=west,xshift=0.06cm] (lgl1) at (lgs1.east) {median; box = middle half; line = range};
\end{tikzpicture}
\caption{Model requests and executed tool calls per episode, grouped by platform. Dashed lines are the frozen limits.}\label{fig:effort}
\end{figure}
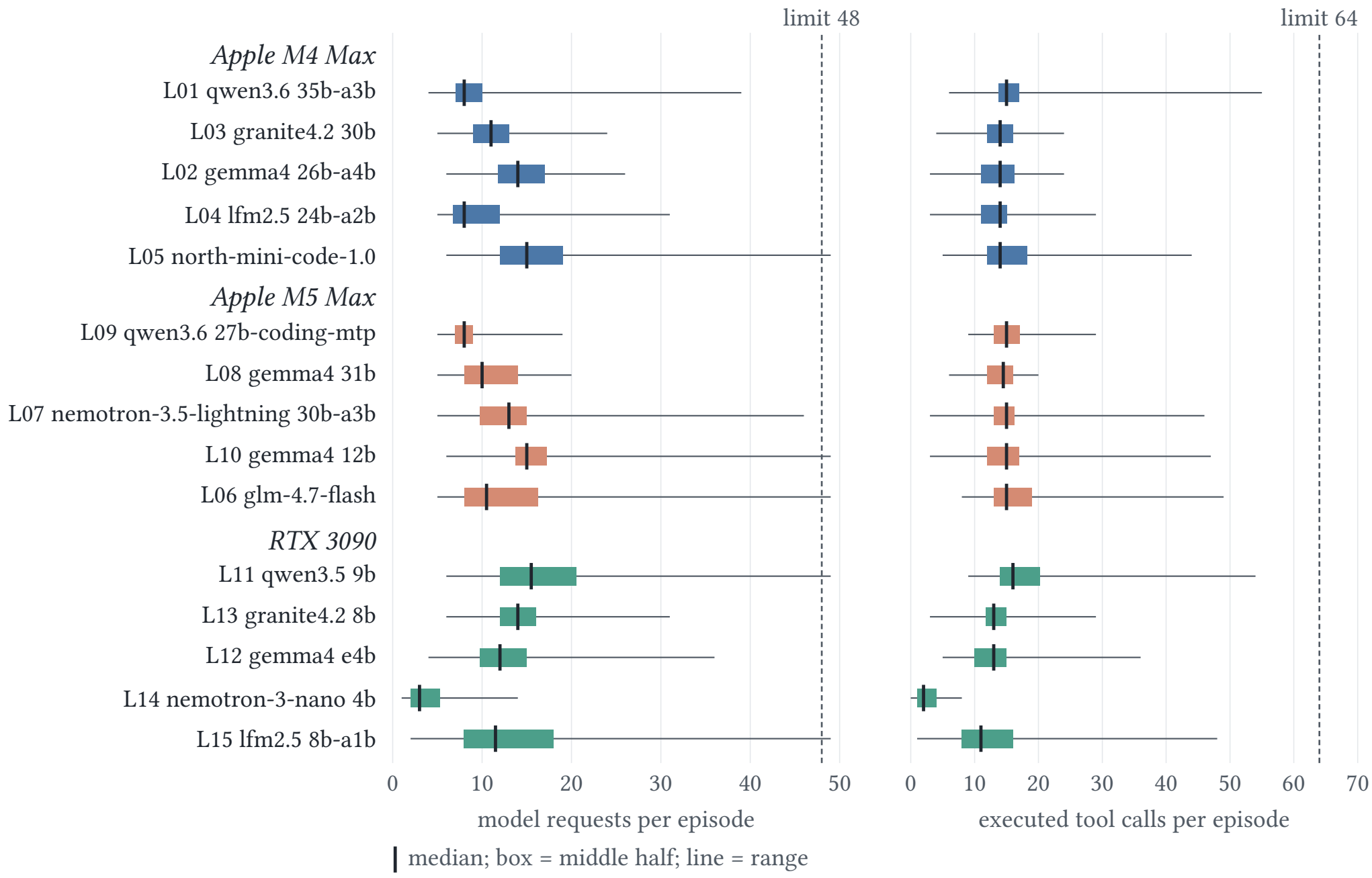

\begin{figure}[tbp]
\centering
\begin{tikzpicture}[x=1cm,y=1cm,line cap=butt]
\draw[vizgrid] (0,-2.4) -- (0,0.05);
\node[viztick,anchor=north] at (0,-2.48) {0\%};
\draw[vizgrid] (1.1,-2.4) -- (1.1,0.05);
\node[viztick,anchor=north] at (1.1,-2.48) {25\%};
\draw[vizgrid] (2.2,-2.4) -- (2.2,0.05);
\node[viztick,anchor=north] at (2.2,-2.48) {50\%};
\draw[vizgrid] (3.3,-2.4) -- (3.3,0.05);
\node[viztick,anchor=north] at (3.3,-2.48) {75\%};
\draw[vizgrid] (4.4,-2.4) -- (4.4,0.05);
\node[viztick,anchor=north] at (4.4,-2.48) {100\%};
\node[viztick,anchor=north] at (2.2,-2.82) {C0 goal reached on this driver};
\fill[vizS7] (0,-0.32) [rounded corners=0.05cm] -- (1.76,-0.32) -- (1.76,-0.08) [sharp corners] -- (0,-0.08) -- cycle;
\node[viztick,anchor=west] at (1.84,-0.2) {6/15};
\fill[vizS4] (0,-0.72) [rounded corners=0.05cm] -- (0.978,-0.72) -- (0.978,-0.48) [sharp corners] -- (0,-0.48) -- cycle;
\node[viztick,anchor=west] at (1.058,-0.6) {2/9};
\fill[vizS7] (0,-1.12) [rounded corners=0.05cm] -- (1.173,-1.12) -- (1.173,-0.88) [sharp corners] -- (0,-0.88) -- cycle;
\node[viztick,anchor=west] at (1.253,-1) {4/15};
\fill[vizS4] (0,-1.52) [rounded corners=0.05cm] -- (1.467,-1.52) -- (1.467,-1.28) [sharp corners] -- (0,-1.28) -- cycle;
\node[viztick,anchor=west] at (1.547,-1.4) {3/9};
\node[viztick,anchor=west] at (0.08,-1.8) {0/7};
\node[viztick,anchor=west] at (0.08,-2.2) {0/17};
\draw[vizgrid] (5.1,-2.4) -- (5.1,0.05);
\node[viztick,anchor=north] at (5.1,-2.48) {0\%};
\draw[vizgrid] (6.2,-2.4) -- (6.2,0.05);
\node[viztick,anchor=north] at (6.2,-2.48) {25\%};
\draw[vizgrid] (7.3,-2.4) -- (7.3,0.05);
\node[viztick,anchor=north] at (7.3,-2.48) {50\%};
\draw[vizgrid] (8.4,-2.4) -- (8.4,0.05);
\node[viztick,anchor=north] at (8.4,-2.48) {75\%};
\draw[vizgrid] (9.5,-2.4) -- (9.5,0.05);
\node[viztick,anchor=north] at (9.5,-2.48) {100\%};
\node[viztick,anchor=north] at (7.3,-2.82) {goal reached, all conditions};
\fill[vizS7] (5.1,-0.32) [rounded corners=0.05cm] -- (6.078,-0.32) -- (6.078,-0.08) [sharp corners] -- (5.1,-0.08) -- cycle;
\node[viztick,anchor=west] at (6.158,-0.2) {8/36};
\fill[vizS4] (5.1,-0.72) [rounded corners=0.05cm] -- (5.956,-0.72) -- (5.956,-0.48) [sharp corners] -- (5.1,-0.48) -- cycle;
\node[viztick,anchor=west] at (6.036,-0.6) {7/36};
\fill[vizS7] (5.1,-1.12) [rounded corners=0.05cm] -- (5.854,-1.12) -- (5.854,-0.88) [sharp corners] -- (5.1,-0.88) -- cycle;
\node[viztick,anchor=west] at (5.934,-1) {6/35};
\fill[vizS4] (5.1,-1.52) [rounded corners=0.05cm] -- (6.289,-1.52) -- (6.289,-1.28) [sharp corners] -- (5.1,-1.28) -- cycle;
\node[viztick,anchor=west] at (6.369,-1.4) {10/37};
\node[viztick,anchor=west] at (5.18,-1.8) {0/18};
\node[viztick,anchor=west] at (5.18,-2.2) {0/54};
\node[vizlab,anchor=east] at (-0.12,-0.2) {L11 \textcolor{vizInk2}{driver 580.173.02}};
\node[vizlab,anchor=east] at (-0.12,-0.6) {L11 \textcolor{vizInk2}{driver 615.71.09}};
\node[vizlab,anchor=east] at (-0.12,-1) {L13 \textcolor{vizInk2}{driver 580.173.02}};
\node[vizlab,anchor=east] at (-0.12,-1.4) {L13 \textcolor{vizInk2}{driver 615.71.09}};
\node[vizlab,anchor=east] at (-0.12,-1.8) {L15 \textcolor{vizInk2}{driver 580.173.02}};
\node[vizlab,anchor=east] at (-0.12,-2.2) {L15 \textcolor{vizInk2}{driver 615.71.09}};
\node[fill=vizS7,minimum width=0.22cm,minimum height=0.15cm,inner sep=0,anchor=west] (lgs1) at (0,0.35) {};
\node[viztick,anchor=west,xshift=0.06cm] (lgl1) at (lgs1.east) {driver 580.173.02};
\node[fill=vizS4,minimum width=0.22cm,minimum height=0.15cm,inner sep=0,anchor=west,xshift=0.3cm] (lgs2) at (lgl1.east) {};
\node[viztick,anchor=west,xshift=0.06cm] (lgl2) at (lgs2.east) {driver 615.71.09};
\end{tikzpicture}
\caption{Descriptive driver split for the three configurations whose episodes ran on two GPU driver versions. Episodes were assigned to machines by availability, not at random, so the split does not estimate a driver effect.}\label{fig:drivers}
\end{figure}
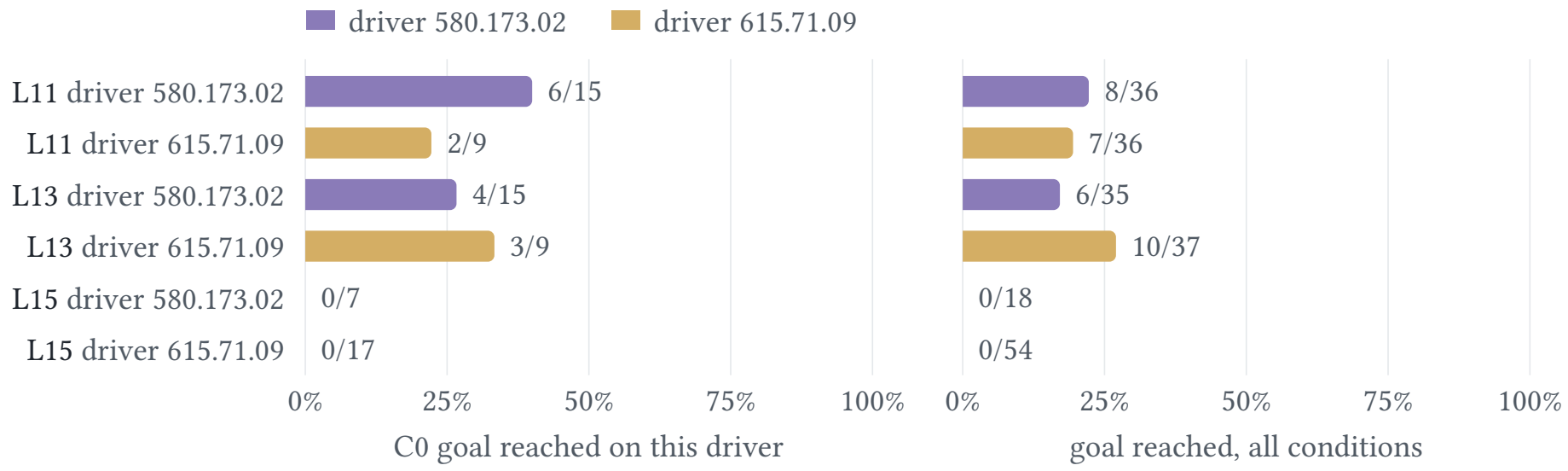

\end{document}